\documentclass[10pt]{article} 
\usepackage[preprint]{tmlr}

\usepackage{amsmath,amsfonts,bm}

\def\eqref#1{equation~\ref{#1}}

\def\1{\bm{1}}

\DeclareMathAlphabet{\mathsfit}{\encodingdefault}{\sfdefault}{m}{sl}
\SetMathAlphabet{\mathsfit}{bold}{\encodingdefault}{\sfdefault}{bx}{n}

\usepackage[utf8]{inputenc} 
\usepackage[T1]{fontenc}    
\usepackage{CJKutf8}        
\usepackage{hyperref}       
\usepackage{url}            
\usepackage{graphicx}       
\usepackage{subcaption}     
\usepackage{booktabs}       
\usepackage{adjustbox}      
\usepackage{rotating}       
\usepackage{wrapfig}        
\usepackage{tabularx}       
\usepackage{longtable}      
\usepackage{amsfonts}       
\usepackage{amssymb}        
\usepackage{nicefrac}       
\usepackage{microtype}      
\usepackage[table]{xcolor} 
\usepackage{xspace}         

\newcommand{\ourmethodname}{Mr.LHDR\xspace}

\newif\ifshowzh
\showzhfalse

\ifdefined\zhon\showzhtrue\fi
\ifdefined\zhoff\showzhfalse\fi

\ifshowzh
  \DeclareRobustCommand{\zh}[1]{\begin{CJK}{UTF8}{gbsn}#1\end{CJK}}
\else
  \DeclareRobustCommand{\zh}[1]{}
\fi

\title{\ourmethodname: A Benchmark for Multimodal Real-World \\ Long-Horizon Deep Research Agents}

\author{\name Minghao Guo$^{*}$ \email minghaoguo20@gmail.com     \\ \addr Mohamed bin Zayed University of Artificial Intelligence
      \AND \name Meng Cao$^{*}$ \email mengcaopku@gmail.com       \\ \addr Mohamed bin Zayed University of Artificial Intelligence
      \AND \name Sui Zhao       \email sui@mail.ustc.edu.cn       \\ \addr University of Science and Technology of China
      \AND \name Siyu Ning      \email ningsiyu@mail.ustc.edu.cn  \\ \addr University of Science and Technology of China
      \AND \name Xin Wang       \email wangxin2252@gmail.com      \\ \addr University of Science and Technology of China
      \AND \name Haoze Zhao     \email Haoze.Zhao@mbzuai.ac.ae    \\ \addr Mohamed bin Zayed University of Artificial Intelligence
      \AND \name Jiaxuan Yang   \email yangjiaxuan@zju.edu.cn     \\ \addr Zhejiang University
      \AND \name Haihong Hao    \email haohaihong@mail.ustc.edu.cn\\ \addr University of Science and Technology of China
      \AND \name Mingfei Han    \email Mingfei.Han@mbzuai.ac.ae   \\ \addr Mohamed bin Zayed University of Artificial Intelligence
      \AND \name Shunlin Rong   \email                            \\ \addr Tencent
      \AND \name Haijun Wu      \email                            \\ \addr Tencent
      \AND \name Xiaodan Liang  \email Xiaodan.Liang@mbzuai.ac.ae \\ \addr Mohamed bin Zayed University of Artificial Intelligence
      \AND \name Xiaojun Chang  \email Xiaojun.Chang@mbzuai.ac.ae \\ \addr Mohamed bin Zayed University of Artificial Intelligence
}

\newcommand{\BREWmainNsystems}{25}
\newcommand{\BREWtopModel}{GPT-5.5}
\newcommand{\BREWtopOA}{43.1}
\newcommand{\BREWtopSA}{34.3}

\newcommand{\BREWtopDRModel}{o3 Deep Research}
\newcommand{\BREWtopDROA}{32.4}
\newcommand{\BREWtopDRSA}{19.6}

\newcommand{\BREWjudgeBackendModel}{DeerFlow (qwen3-vl-235b)}
\newcommand{\BREWjudgeBackendOA}{15.7}
\newcommand{\BREWjudgeBackendSA}{9.8}

\newcommand{\BREWbootB}{10000}
\newcommand{\BREWbootSeed}{20260731}
\newcommand{\BREWgptfiveOA}{43.1}
\newcommand{\BREWgptfiveOAlo}{33.3}
\newcommand{\BREWgptfiveOAhi}{52.9}
\newcommand{\BREWgptVsOfourSep}{96.4}
\newcommand{\BREWclusterMaxSep}{67.1}
\newcommand{\BREWcsdacsMin}{3.0}
\newcommand{\BREWcsdacsMax}{8.9}
\newcommand{\BREWoasaMaxCond}{72.7}
\newcommand{\BREWjudgeN}{12}
\newcommand{\BREWjudgeFamilies}{6}
\newcommand{\BREWjudgeResponses}{2065}
\newcommand{\BREWjudgeSteps}{24782}

\newcommand{\BREWitemsubsetItems}{102}
\newcommand{\BREWitemsubsetCats}{8}
\newcommand{\BREWitemsubsetJudges}{12}

\newcommand{\BREWitemsubsetCells}{702}
\newcommand{\BREWitemsubsetSteps}{8417}
\newcommand{\BREWitemsubsetOAagreeMin}{95.9}
\newcommand{\BREWitemsubsetOAagreeMax}{99.7}
\newcommand{\BREWitemsubsetOAkappaMin}{0.845}
\newcommand{\BREWitemsubsetOAkappaMax}{0.989}

\newcommand{\BREWitemsubsetXfamPairs}{58}
\newcommand{\BREWitemsubsetXfamAgreeMin}{95.9}
\newcommand{\BREWitemsubsetXfamKappaMin}{0.845}
\newcommand{\BREWitemsubsetFleissOA}{0.940}
\newcommand{\BREWitemsubsetFleissOACIlo}{0.918}
\newcommand{\BREWitemsubsetFleissOACIhi}{0.958}
\newcommand{\BREWitemsubsetFleissStep}{0.803}
\newcommand{\BREWitemsubsetUnanimous}{94.3}

\newcommand{\BREWitemsubsetPaperKappaMax}{0.870}
\newcommand{\BREWitemsubsetOtherKappaMin}{0.877}
\newcommand{\BREWitemsubsetPaperOArate}{16.5}
\newcommand{\BREWitemsubsetPaperStepRate}{50.9}
\newcommand{\BREWitemsubsetOtherOArateMax}{16.1}
\newcommand{\BREWitemsubsetOtherStepRateMax}{50.0}

\newcommand{\BREWitemsubsetSensWeakestFleissOA}{0.909}
\newcommand{\BREWitemsubsetSensWeakestFleissOACIlo}{0.877}
\newcommand{\BREWitemsubsetSensWeakestFleissOACIhi}{0.937}

\newcommand{\BREWitemsubsetBootstrapB}{4000}
\newcommand{\BREWjuryModels}{10}
\newcommand{\BREWjuryThreshold}{50\%}
\newcommand{\BREWjuryResponses}{2157}
\newcommand{\BREWjuryUnanimous}{97.5}
\newcommand{\BREWjuryPaperOAagree}{97.1}
\newcommand{\BREWjuryPaperOAkappa}{0.890}
\newcommand{\BREWjuryPaperStepAgree}{86.6}
\newcommand{\BREWjuryPaperStepKappa}{0.728}

\newcommand{\BREWtoImageOA}{7.8}
\newcommand{\BREWtoTextOA}{3.9}
\newcommand{\BREWtoImageCS}{39.3}
\newcommand{\BREWtoTextCS}{26.9}
\newcommand{\BREWtoImageDACS}{34.2}
\newcommand{\BREWtoTextDACS}{21.6}
\newcommand{\BREWtoCSgain}{12.4}
\newcommand{\BREWtoCSgainCIlo}{6.9}
\newcommand{\BREWtoCSgainCIhi}{18.3}
\newcommand{\BREWtoCSgainProb}{100.0}
\newcommand{\BREWtoDACSgain}{12.6}
\newcommand{\BREWtoDACSgainCIlo}{6.9}
\newcommand{\BREWtoDACSgainCIhi}{18.9}
\newcommand{\BREWtoDACSgainProb}{100.0}

\newcommand{\BREWtoXjudgeN}{12}
\newcommand{\BREWtoXfamilyN}{6}
\newcommand{\BREWtoXcrossN}{10}

\newcommand{\BREWtoXposOA}{12}

\newcommand{\BREWtoXposSA}{7}

\newcommand{\BREWtoXposCS}{12}
\newcommand{\BREWtoXselfDACS}{12.6}
\newcommand{\BREWtoXmedDACS}{15.0}
\newcommand{\BREWtoXposDACS}{12}

\newcommand{\BREWtfGemTopDACS}{70.1}

\newcommand{\BREWtfCtrlBestDACS}{37.8}
\newcommand{\BREWtfCtrlWorstDACS}{29.0}
\newcommand{\BREWtfGemZero}{12.7}
\newcommand{\BREWtfGemFull}{33.3}

\newcommand{\BREWtfJudgeN}{12}

\newcommand{\BREWtfJudgeWorstRank}{2}
\newcommand{\BREWfailSystems}{25}

\newcommand{\BREWfailRuns}{2,550}
\newcommand{\BREWfailSteps}{30,775}
\newcommand{\BREWfailFailedSteps}{17,409}
\newcommand{\BREWfailStepRate}{56.6}
\newcommand{\BREWfailRunsWithFailure}{2,247}
\newcommand{\BREWfailRunRate}{88.1}

\newcommand{\BREWfailDagEarlyRate}{48.8}
\newcommand{\BREWfailDagMiddleRate}{55.3}
\newcommand{\BREWfailDagLateRate}{69.3}

\def\month{MM}  
\def\year{YYYY} 
\def\openreview{\url{https://openreview.net/forum?id=XXXX}} 

\begin{document}
\maketitle
{\renewcommand{\thefootnote}{}\footnotetext{$^{*}$Equal contribution.}}

\begin{abstract}
Deep research agents are increasingly capable of searching the web, invoking tools, examining multimodal evidence, and synthesizing information from multiple sources. Yet most existing benchmarks are confined to relatively medium-horizon evidence exploration. MM-BrowseComp, for example, contains an average of only 3.0 checklist items per question, leaving the ability of agents to sustain long-horizon deep research underexplored. To this end, we introduce \ourmethodname (\textbf{M}ultimodal \textbf{r}eal-world \textbf{L}ong-\textbf{H}orizon \textbf{D}eep \textbf{R}esearch), a benchmark designed to evaluate real-world deep research over long, irreducible chains of interdependent evidence across eight categories. Each question is constructed from a hidden Node-Relation graph and requires, on average, 12.1 necessary intermediate conclusions with a mean dependency depth of 10.4 before reaching a short, unique, and verifiable answer. We design each question so that multiple key steps involve images, maps, PDFs, logos, charts, tables, or video frames. As a hard inclusion criterion, every question contains at least one piece of non-text evidence that changes the reasoning state. \ourmethodname therefore evaluates both whether an agent returns the final answer and whether its stated intermediate conclusions are correct given the annotated dependencies. We evaluate regular models, dedicated deep research systems, and framework-based agents using Overall Accuracy (OA), Strict Accuracy (SA), Checklist Score (CS), and Dependency-Aware Checklist Score (DACS). Empirically, we find that: (1) the run with the highest OA, \BREWtopModel, reaches only \BREWtopOA\% OA and \BREWtopSA\% SA, while the dedicated deep research system with the highest OA, \BREWtopDRModel, reaches \BREWtopDROA\% OA and \BREWtopDRSA\% SA, showing that final-answer accuracy substantially overstates complete, dependency-consistent task success; (2) withholding images lowers DACS by \BREWtoDACSgain{} percentage points in a controlled ablation, confirming that non-text evidence materially contributes to the reasoning chain; and (3) across checklist-length ranges with sufficient observations, SA decreases monotonically with chain length for every representative model. Together, these findings identify sustained, dependency-consistent evidence integration, instead of isolated fact retrieval, as a central bottleneck for current deep research agents.
\makeatletter
\if@accepted
Code and data are available at \url{https://github.com/minghaoguo20/Mr-LHDR}.
\fi
\makeatother

\zh{
深度研究智能体在网页搜索、工具调用、多模态证据分析以及多来源信息综合方面的能力日益增强。然而，现有基准大多局限于短期或中期的证据探索。例如，MM-BrowseComp 每道题平均仅包含 3.0 个 checklist item，因此智能体持续开展 long-horizon deep research 的能力仍缺乏充分评估。为此，我们提出 \ourmethodname（\textbf{M}ultimodal \textbf{r}eal-world \textbf{L}ong-\textbf{H}orizon \textbf{D}eep \textbf{R}esearch），一个面向八个类别、专门评估真实世界深度研究中长而不可约的相互依赖证据链的基准。每道题均基于隐藏的 Node-Relation graph 构建，在得到简短、唯一且可验证的答案之前，平均需要完成 12.1 个必要中间结论，平均依赖深度为 10.4。我们在设计每道题时，使多个关键步骤涉及图片、地图、PDF、logo、图表、表格或视频帧。作为硬性纳入标准，每道题都至少包含一处能够改变推理状态的非文本证据。因此，\ourmethodname 既评估智能体能否返回最终答案，也评估其陈述的中间结论在给定标注依赖关系时是否正确。我们使用 Overall Accuracy (OA)、Strict Accuracy (SA)、Checklist Score (CS) 和 Dependency-Aware Checklist Score (DACS) 评估常规模型、专用 deep research 系统以及基于框架的 agent。实证结果表明：（1）OA 最高的运行 \BREWtopModel{} 仅达到 \BREWtopOA\% OA 和 \BREWtopSA\% SA，而 OA 最高的专用 deep research 系统 \BREWtopDRModel{} 仅达到 \BREWtopDROA\% OA 和 \BREWtopDRSA\% SA，说明最终答案正确率会明显高估完整且依赖一致的任务完成能力；（2）在受控消融中，移除图像使 DACS 降低 \BREWtoDACSgain{} 个百分点，证实非文本证据对推理链具有实质贡献；（3）在样本量充足的 checklist 长度区间内，所有代表性模型的 SA 均随链条增长而单调下降。综合来看，持续保持依赖一致的证据整合能力，而非孤立事实检索，是当前 deep research agent 的核心瓶颈。
}
\end{abstract}

\section{Introduction}

AI agents are increasingly expected to conduct deep research rather than answer isolated questions~\citep{Zheng2025DeepResearcherSD,Zhang2025FromWS,Du2025DeepResearchBA}. A useful research assistant should decompose an underspecified goal, search across heterogeneous sources, inspect visual and document evidence, maintain intermediate conclusions, and verify a final answer against the accumulated evidence. Many real-world tasks follow exactly this pattern: an agent may need to identify an object in an image, trace an entity chain across the web, read a map, chart, or PDF, and then combine the evidence into a short factual answer. The difficulty stems not only from obscure individual facts, but also from the risk that an early error redirects the entire investigation.

\zh{
AI 智能体越来越需要开展深度研究，而不只是回答孤立的问题。~\citep{Zheng2025DeepResearcherSD,Zhang2025FromWS,Du2025DeepResearchBA}一个有效的研究助手应当能够分解描述不充分的目标，在异质来源中搜索，检查视觉和文档证据，维护中间结论，并依据累积的证据验证最终答案。许多真实任务都遵循这一模式：智能体可能需要识别图片中的对象，在网页中追踪实体链，阅读地图、图表或 PDF，再将证据整合为简短的事实性答案。难点不仅在于单个事实可能较为冷门，还在于早期错误可能使整个调查偏离方向。
}

\begin{figure}[t]
\centering
\includegraphics[width=0.8\linewidth]{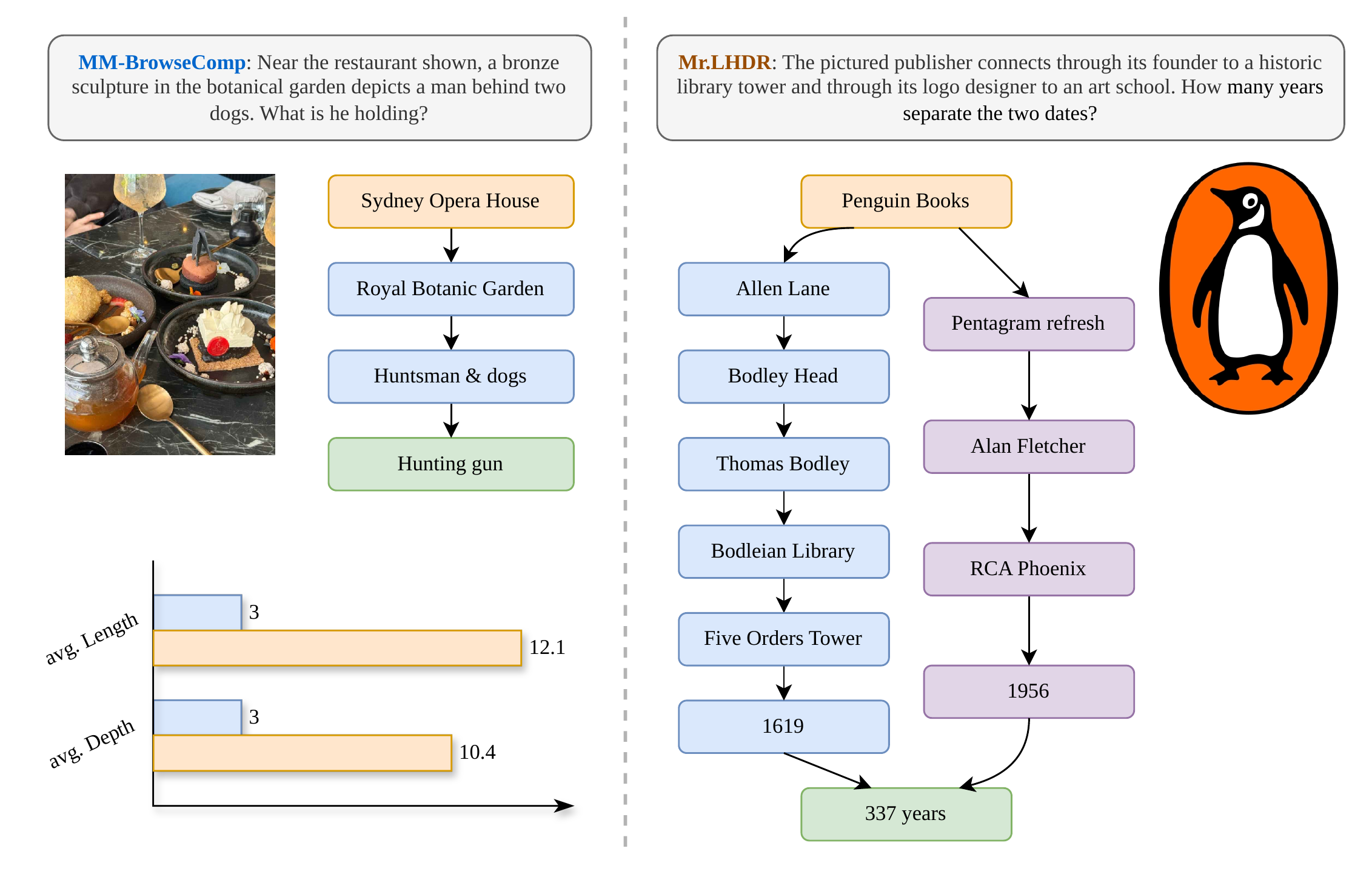}
\caption{Comparison of annotated structural horizon. The public MM-BrowseComp image-input example uses four sequential checklist items; its original 224-item release averages 3.0 items per question~\citep{li2025mmbrowsecompcomprehensivebenchmarkmultimodal}. The \ourmethodname example uses 12 necessary conclusions in two prerequisite branches, while the full benchmark averages 12.1 conclusions and a dependency depth of 10.4. Both settings require multimodal web research and short verifiable answers. Nodes denote conclusions rather than browsing actions; arrows denote annotated prerequisites.\zh{ 标注结构性 horizon 的对比。MM-BrowseComp 公开的 image-input 示例包含四个顺序 checklist item，其最初的 224 题版本平均每题包含 3.0 项。\ourmethodname 示例包含两条前置依赖分支上的 12 个必要结论，而完整 benchmark 的平均结论数为 12.1，平均 dependency depth 为 10.4。两种设定都要求进行多模态网页研究并返回简短且可验证的答案。节点表示结论而非浏览动作，箭头表示已标注的前置依赖。}}
\label{fig:structural_horizon_comparison}
\end{figure}

Existing benchmarks have made important progress in evaluating web-enabled agents~\citep{wei2025browsecomp,Zhou2025BrowseCompZHBW,li2025mmbrowsecompcomprehensivebenchmarkmultimodal,Du2025DeepResearchBA}, but they do not fully capture this setting. BrowseComp-style tasks emphasize hard-to-find information on the open web and are useful for measuring search and browsing competence, but they remain primarily text-centered. MM-BrowseComp-style tasks add multimodal browsing, but remain concentrated on short- to medium-horizon research: MM-BrowseComp contains only 3.0 checklist items per question on average~\citep{li2025mmbrowsecompcomprehensivebenchmarkmultimodal}, leaving long chains of dependent evidence steps largely untested. \ourmethodname targets the long, tightly coupled, multi-source, cross-modal research processes found in real-world investigations. Strong agents may solve individual browsing or multimodal steps, yet still fail when a task requires many dependent steps, cross-modal bridges, and repeated verification. Figure~\ref{fig:structural_horizon_comparison} illustrates this distinction through one public MM-BrowseComp example and one \ourmethodname example.

\zh{
现有基准在评估具备网页能力的智能体方面已取得重要进展~\citep{wei2025browsecomp,Zhou2025BrowseCompZHBW,li2025mmbrowsecompcomprehensivebenchmarkmultimodal,Du2025DeepResearchBA}，但仍未完全覆盖这一场景。BrowseComp 风格的任务强调在开放网页中寻找难以发现的信息，有助于衡量搜索和浏览能力，但仍主要以文本为中心。MM-BrowseComp 风格的任务加入了多模态浏览，但仍主要集中于短至中等 horizon 的研究：MM-BrowseComp 每道题平均仅包含 3.0 个 checklist item，因此长链依赖证据步骤在很大程度上仍未得到检验。\ourmethodname 面向真实研究中的长链路、高耦合、多来源、跨模态研究过程。强智能体可能能够完成单个浏览或多模态步骤，但当任务需要许多相互依赖的步骤、跨模态桥接和反复验证时，仍可能失败。Figure~\ref{fig:structural_horizon_comparison} 通过一个公开的 MM-BrowseComp 示例和一个 \ourmethodname 示例直观展示了这一区别。
}

We introduce \ourmethodname (Multimodal real-world Long-Horizon Deep Research), a benchmark for real-world open-web research tasks. \ourmethodname contains 102 questions across eight categories. Each item includes a natural-language question, a short verifiable answer, source evidence, multimodal evidence, an irreducible checklist of necessary intermediate conclusions, a category label, a coarse authored-template label, and operation tags. The final answers are intentionally short, unique, and stable, such as a person name, place name, organization name, year, number, color, or short phrase. \ourmethodname frames real-world research tasks as verifiable short-answer questions whose difficulty arises from real evidence dependencies.

\zh{
我们提出 \ourmethodname（Multimodal real-world Long-Horizon Deep Research），一个面向真实开放网页研究任务的基准。\ourmethodname 包含 102 个问题，覆盖八个类别。每个条目包含自然语言问题、简短且可验证的答案、来源证据、多模态证据、由必要中间结论组成的不可约 checklist，以及 category 标签、粗粒度 authored-template 标签和 operation tags。最终答案被有意设计为简短、唯一且稳定的形式，例如人名、地名、机构名、年份、数字、颜色或短语。\ourmethodname 将真实世界研究任务表述为可验证的短答案问题，其难度来自真实的证据依赖关系。
}

The design of \ourmethodname follows three principles. \textbf{(1) Structural horizon.} Long-horizon denotes the depth of prerequisite-linked evidence structures rather than input or output length: an agent must preserve an evolving research state across dependent steps, not merely process a long prompt or produce a long response. \textbf{(2) Open-world multimodal investigation.} Solving an item may require decomposing the goal, locating evidence across webpages, PDFs, images, maps, videos, audio, and tables, bridging information across modalities, preserving intermediate conclusions, repeatedly verifying constraints, and returning a final short answer. \textbf{(3) Dependency-aware evaluation.} \ourmethodname reports Overall Accuracy (OA), Strict Accuracy (SA), Checklist Score (CS), and Dependency-Aware Checklist Score (DACS) to distinguish final-answer correctness, complete response correctness, intermediate-conclusion coverage, and dependency-consistent coverage. SA, CS, and DACS are computed from conclusions stated in the submitted response, with DACS additionally requiring their annotated prerequisites to be satisfied. Because these metrics neither observe the agent's internal reasoning nor establish that a stated conclusion came from actual retrieval, they evaluate dependency-consistent stated conclusions rather than actual evidence-chain completion.

\zh{
\ourmethodname 的设计遵循三项原则。\textbf{（1）结构性 horizon。} Long-horizon 表示具有前置依赖的证据结构深度，而非输入或输出的文本长度：智能体必须在相互依赖的步骤中维护不断演化的研究状态，而不只是处理长 prompt 或生成长回复。\textbf{（2）开放世界多模态调查。} 解决一个条目可能需要分解目标，在网页、PDF、图片、地图、视频、音频和表格中定位证据，桥接不同模态的信息，维护中间结论，反复验证约束，并给出最终短答案。\textbf{（3）依赖感知评测。} \ourmethodname 报告 Overall Accuracy (OA)、Strict Accuracy (SA)、Checklist Score (CS) 和 Dependency-Aware Checklist Score (DACS)，以区分最终答案正确性、完整回复正确性、中间结论覆盖度和依赖一致覆盖度。SA、CS 和 DACS 根据提交回复中陈述的结论计算，其中 DACS 还要求其已标注的前置依赖得到满足。由于这些指标既不观测智能体的内部推理，也不能证明某项陈述确实来自真实检索，因此它们衡量的是符合依赖关系的已陈述结论，而不是实际证据链的完成情况。
}

We evaluate \BREWmainNsystems{} regular models, dedicated deep research systems, and framework-based agents on \ourmethodname, yielding four main findings. \textbf{First, current systems remain far from saturation.} The regular-model run with the highest OA, \BREWtopModel{}, achieves only \BREWtopOA\% OA and \BREWtopSA\% SA; the strongest dedicated system by OA, \BREWtopDRModel{}, reaches \BREWtopDROA\% OA and \BREWtopDRSA\% SA; and \BREWjudgeBackendModel{} reaches \BREWjudgeBackendOA\% OA and \BREWjudgeBackendSA\% SA. \textbf{Second, final-answer success overstates complete task execution.} The substantial OA--SA gaps show that a correct final answer often coexists with missing required conclusions, while the lower DACS than CS reveals conclusions whose annotated prerequisites are absent. \textbf{Third, visual evidence materially improves evidence-chain recovery.} In a controlled same-model ablation, providing images raises DACS from \BREWtoTextDACS\% to \BREWtoImageDACS\%, a gain of \BREWtoDACSgain{} points (95\% CI [\BREWtoDACSgainCIlo, \BREWtoDACSgainCIhi]). \textbf{Fourth, longer evidence chains increase strict-completion pressure.} Across the three checklist-length ranges with at least 20 items, SA decreases monotonically with checklist length for every representative model. Because confidence intervals for many neighboring systems overlap, these results support broad capability patterns rather than a statistically separable ranking.

\zh{
我们在 \ourmethodname 上评估了 \BREWmainNsystems{} 个常规模型、专用 deep research 系统以及基于框架的 agent，并得到四项主要发现。\textbf{第一，当前系统距离饱和仍然很远。} OA 最高的常规模型运行 \BREWtopModel{} 仅达到 \BREWtopOA\% OA 和 \BREWtopSA\% SA；OA 最高的专用系统 \BREWtopDRModel{} 达到 \BREWtopDROA\% OA 和 \BREWtopDRSA\% SA；\BREWjudgeBackendModel{} 则仅达到 \BREWjudgeBackendOA\% OA 和 \BREWjudgeBackendSA\% SA。\textbf{第二，最终答案正确率会高估任务的完整执行程度。} 明显的 OA--SA 差距表明，最终答案正确时，回复仍可能缺失必要结论；DACS 低于 CS 则揭示了缺少已标注前置依赖的结论。\textbf{第三，视觉证据能够实质性提升证据链恢复能力。} 在使用同一模型的受控消融中，提供图像使 DACS 从 \BREWtoTextDACS\% 提升至 \BREWtoImageDACS\%，增益为 \BREWtoDACSgain{} 个百分点（95\% CI [\BREWtoDACSgainCIlo, \BREWtoDACSgainCIhi]）。\textbf{第四，更长的证据链会增加严格完成压力。} 在样本量不少于 20 的三个 checklist-length 区间中，所有代表性模型的 SA 均随 checklist 长度单调下降。由于许多相邻系统的置信区间相互重叠，这些结果支持的是总体能力规律，而非统计上可区分的系统排名。
}

\paragraph{Contributions.} We make the following contributions:

\zh{
我们的主要贡献如下：
}

\begin{itemize}
    \item \textbf{A benchmark for long-horizon multimodal deep research.} We introduce \ourmethodname{}, an open-web benchmark comprising 102 questions across eight real-world categories. Each question is constructed using a Node-Relation design and paired with multimodal source evidence, a short, unique, and verifiable answer, and an irreducible checklist structured as an author-verified dependency DAG. The benchmark contains 1,231 annotated intermediate conclusions with a mean dependency depth of 10.4, operationalizing long-horizon difficulty through prerequisite-linked evidence structures.

    \zh{
    \textbf{一个面向 long-horizon 多模态深度研究的基准。} 我们提出 \ourmethodname{}，一个包含 102 个问题、覆盖八类真实世界主题的开放网页基准。每个问题均采用 Node-Relation 设计构建，并配有多模态来源证据、简短、唯一且可验证的答案，以及组织为作者核定依赖 DAG 的不可约 checklist。该基准共包含 1,231 个已标注的中间结论，平均依赖深度为 10.4，通过具有前置依赖的证据结构刻画 long-horizon 难度。
    }
    \item \textbf{A dependency-aware evaluation protocol.} We introduce Dependency-Aware Checklist Score (DACS), which recursively conditions the credit for each stated conclusion on the correctness of its annotated prerequisites. Together with Overall Accuracy (OA), Strict Accuracy (SA), and Checklist Score (CS), the protocol distinguishes among final-answer correctness, complete response correctness, intermediate-conclusion coverage, and dependency-consistent coverage, enabling uniform evaluation across heterogeneous research systems.

    \zh{
    \textbf{一套依赖感知的评测协议。} 我们提出 Dependency-Aware Checklist Score (DACS)，根据每个已陈述结论的已标注前置依赖是否正确，递归决定该结论的得分。该协议结合 Overall Accuracy (OA)、Strict Accuracy (SA) 和 Checklist Score (CS)，区分最终答案正确性、完整回复正确性、中间结论覆盖度和依赖一致覆盖度，从而统一评估异构研究系统。
    }
    \item \textbf{A systematic empirical study of current research systems.} We evaluate \BREWmainNsystems{} systems spanning tool-augmented VLMs, tool-free VLMs, dedicated deep research systems, and framework-based agents. The results reveal systematic gaps between final-answer correctness and complete stated-conclusion correctness, as well as between checklist coverage and dependency-consistent coverage. A controlled image ablation using the same model further demonstrates the contribution of visual evidence to recovering dependency-constrained conclusions. Bootstrap uncertainty estimation and robustness analyses of the judge and dependency graph support these findings.

    \zh{
    \textbf{一项针对当前研究系统的系统性实证研究。} 我们评估 \BREWmainNsystems{} 个系统，涵盖 tool-augmented VLM、tool-free VLM、专用 deep research system 和 framework-based agent。结果揭示了最终答案正确性与完整已陈述结论正确性之间，以及 checklist 覆盖度与依赖一致覆盖度之间的系统性差距。使用同一模型开展的受控图像消融进一步表明，视觉证据有助于恢复受依赖关系约束的结论。Item-bootstrap 不确定性估计以及对 judge 和依赖图的稳健性分析为这些发现提供了支持。
    }
\end{itemize}

\section{Related Work}

\paragraph{Multimodal Understanding and Evidence Benchmarks.}
Recent multimodal large language models have moved beyond image--text perception to support documents, charts, videos, screens, long contexts, and agent-centric visual reasoning. 
\citet{li2025llavaonevision,guo2025seed1,vteam2025glm45vglm41vthinkingversatilemultimodal,Qwen3-VL} exemplify this shift toward general-purpose multimodal understanding, long-context grounding, and reasoning over interleaved visual and textual inputs. Multimodal evaluation has evolved accordingly. \citet{liu2024mmbench} and \citet{Yue2023MMMUAM} assess broad and expert-level reasoning; \citet{masry-etal-2022-chartqa} and \citet{Mathew2020DocVQAAD} focus on charts and document images; and newer evidence-oriented benchmarks examine long multimodal documents, retrieval-augmented visual culture understanding, multimodal document RAG, and conflicts between parametric knowledge and external multimodal evidence.~\citep{duan2024vlmevalkit,Huybrechts2025DocumentHA,Li2025RAVENEAAB,Dong2025BenchmarkingRM,Jia2025BenchmarkingMK} Collectively, these studies establish the importance of evidence selection, cross-modal integration, domain knowledge, and robustness to conflicting priors. In these benchmarks, however, evidence is typically supplied with the query, contained in a fixed corpus, or retrieved within a predefined document or RAG setting. \ourmethodname considers a complementary open-research setting in which an agent must discover necessary non-textual evidence on the open web, connect it to evolving textual and symbolic constraints, and retain the resulting intermediate conclusions throughout a long-horizon Node-Relation research process.

\zh{
近年来，多模态大语言模型已超越图文感知，能够处理文档、图表、视频、屏幕和长上下文，并支持以智能体为中心的视觉推理。
\citet{li2025llavaonevision,guo2025seed1,vteam2025glm45vglm41vthinkingversatilemultimodal,Qwen3-VL} 等系统体现了这一转变，即向通用多模态理解、长上下文信息对齐以及交错视觉与文本输入的推理发展。多模态评测也随之演进。\citet{liu2024mmbench} 和 \citet{Yue2023MMMUAM} 评估广泛领域与专家级推理； \citet{masry-etal-2022-chartqa} 和 \citet{Mathew2020DocVQAAD} 聚焦图表与文档图像；较新的证据导向基准则研究长篇多模态文档、检索增强的视觉文化理解、多模态文档 RAG，以及参数化知识与外部多模态证据之间的冲突~\citep{duan2024vlmevalkit,Huybrechts2025DocumentHA,Li2025RAVENEAAB,Dong2025BenchmarkingRM,Jia2025BenchmarkingMK}。总体而言，这些研究确立了证据选择、跨模态整合、领域知识运用以及对冲突先验保持鲁棒性的重要性。然而，在这些基准中，证据通常随问题一同提供、包含在固定语料库中，或从预定义的文档或 RAG 场景中检索。\ourmethodname 研究一个与之互补的开放式研究场景：智能体必须在开放网页上发现必要的非文本证据，将其与不断演化的文本和符号约束相联系，并在 long-horizon Node-Relation 研究过程中持续保留由此得到的中间结论。
}

\paragraph{Tool-Enhanced Browsing and Deep Research Agents.}
Tool-enhanced agents extend language and multimodal models with search engines, browsers, APIs, code execution, and other external tools. \citet{Nakano2021WebGPTBQ}, \citet{yao2023react} and \citet{Schick2023ToolformerLM} established early paradigms for browser-assisted question answering, interleaved reasoning and acting, and learned tool invocation. More recent work on search and deep-research training has shifted from retrieval-augmented answering toward agents that plan searches, decompose questions, inspect sources, and revise intermediate states over multiple turns. Examples include \citet{Li2025Searcho1AS,Song2025R1SearcherIT,Jin2025SearchR1TL,Zheng2025DeepResearcherSD,Zhang2025FromWS,Li2025WebSailorNS} User-facing systems, including \citet{oaidr,geminidr,perplexitydr,grokdeepsearch,msftdr}, and open-source frameworks such as \citet{bytedance2026deerflow} reflect the same shift toward multi-step research workflows and report synthesis. Although these developments show that agents can search, browse, invoke tools, collect sources, and synthesize long-form outputs, they leave a more specific evaluation question open: can an agent maintain the research state itself---including the objective, entities, constraints, source support, and multimodal bridge conclusions---across many dependent steps? This question is the focus of \ourmethodname.

\zh{
工具增强型智能体利用搜索引擎、浏览器、API、代码执行及其他外部工具扩展语言模型和多模态模型的能力。\citet{Nakano2021WebGPTBQ}、\citet{yao2023react} 和 \citet{Schick2023ToolformerLM} 奠定了浏览器辅助问答、交错推理与行动，以及通过学习调用工具等早期范式。近期的搜索与深度研究训练正从检索增强回答转向能够规划搜索、分解问题、检查来源并在多轮交互中修订中间状态的智能体。代表性工作包括 \citet{Li2025Searcho1AS,Song2025R1SearcherIT,Jin2025SearchR1TL,Zheng2025DeepResearcherSD,Zhang2025FromWS,Li2025WebSailorNS}。\citet{oaidr,geminidr,perplexitydr,grokdeepsearch,msftdr} 等面向用户的系统，以及 \citet{bytedance2026deerflow} 等开源框架，也体现了向多步研究工作流和报告综合发展的相同趋势。尽管这些进展表明智能体已能够进行搜索与浏览、调用工具、收集来源并综合生成长篇内容，但一个更具体的评测问题仍有待回答：智能体能否在众多相互依赖的步骤中持续维护研究状态本身，包括目标、实体、约束、来源支撑和多模态桥接结论？这正是 \ourmethodname 所关注的问题。
}

\paragraph{Browsing, Search, and Deep Research Benchmarks.}
Existing benchmarks evaluate related agent capabilities from several perspectives. \citet{yao2022webshop,deng2023mind2web,zhou2024webarena,he2024webvoyager,xie2024osworld,mialon2024gaia,Jiang2024MMSearchBT} span e-commerce interaction, real-website navigation, realistic web environments, multimodal web operation, desktop control, general assistant tasks, and multimodal search. 
Information-seeking evaluations more directly related to deep research include \citet{wei2025browsecomp} and \citet{Zhou2025BrowseCompZHBW}, which emphasize hard-to-locate short answers on the open web; \citet{li2025mmbrowsecompcomprehensivebenchmarkmultimodal}, which requires multimodal browsing evidence and provides verified checklists; and \citet{Du2025DeepResearchBA}, which evaluates long-form research reports. 
Complementary agentic-search benchmarks examine dynamic information seeking, source attribution, reproducibility over fixed corpora, process-aware search, and the breadth or depth of evidence collection.~\citep{xi2025infodeepseekbenchmarkingagenticinformation,gou2025mind2web2evaluatingagentic,chen2025browsecompplusfairtransparentevaluation,xu2025ravinerealityalignedevaluationagentic,wong2025widesearchbenchmarkingagenticbroad,lan2025deepwidesearchbenchmarkingdepthwidth} 
Recent work narrows the gap further. \citet{tao2026mmsearchplusbenchmarkingprovenanceawaresearch} emphasizes provenance-aware browsing from fine-grained visual cues, while \citet{zhang2026browsecompv3visualverticalverifiable} introduces expert-validated subgoals and a subgoal-coverage process score for open-web multimodal browsing. \citet{gupta2026deepsearchqabridgingcomprehensivenessgap} evaluates causal-chain open-web research through objectively verifiable single or set-valued answers, but scores outcomes rather than intermediate chains. \citet{ning2026mcsearchevaluatingenhancingmultimodal} provides step-wise multimodal reasoning graphs and process metrics, albeit over a fixed local knowledge base. Report-oriented benchmarks address a different output regime. \citet{huang2026mmdeepresearchbenchbenchmarkmultimodaldeep} assesses citation-grounded multimodal research reports, while \citet{ye2026miroevalbenchmarkingmultimodaldeep} jointly evaluates report quality, factuality, and exposed research trajectories on text-only and multimodal tasks. \ourmethodname complements these settings by combining open-web search, required real-world multimodal evidence, concise verifiable answers, long-horizon Node-Relation structures, and response-level dependency-aware checklist evaluation. Here, horizon refers to maintaining research state across prerequisite-linked conclusions, rather than producing long-form output, processing long-context input, or executing a prescribed number of browsing actions. Unlike unordered checklist or subgoal coverage, DACS uses an item-specific Directed Acyclic Graph (DAG). It withholds downstream credit when required predecessors are not satisfied. Unlike trace-based or fixed-corpus process evaluation, DACS applies to heterogeneous systems using only their submitted responses, without claiming to observe latent reasoning or tool-use traces. Table~\ref{tab:dataset_comparison} compares \ourmethodname with representative short-answer browsing and structured-search benchmarks in terms of retrieval setting, multimodal requirements, output form, gold intermediate structure, and prerequisite-gated scoring.

\zh{
现有基准从多个角度评估相关的智能体能力。\citet{yao2022webshop,deng2023mind2web,zhou2024webarena,he2024webvoyager,xie2024osworld,mialon2024gaia,Jiang2024MMSearchBT} 涵盖电子商务交互、真实网站导航、真实感较强的网页环境、多模态网页操作、桌面控制、通用助手任务和多模态搜索。
与深度研究更直接相关的信息搜寻评测包括 \citet{wei2025browsecomp} and \citet{Zhou2025BrowseCompZHBW}，它们强调在开放网页中寻找难以定位的短答案；\citet{li2025mmbrowsecompcomprehensivebenchmarkmultimodal} 要求提供多模态浏览证据，并包含经验证的 checklist；\citet{Du2025DeepResearchBA} 则评估长篇研究报告。
其他互补的智能体搜索基准研究动态信息搜寻、来源归因、固定语料库上的可复现性、过程感知搜索，以及证据收集的广度或深度~\citep{xi2025infodeepseekbenchmarkingagenticinformation,gou2025mind2web2evaluatingagentic,chen2025browsecompplusfairtransparentevaluation,xu2025ravinerealityalignedevaluationagentic,wong2025widesearchbenchmarkingagenticbroad,lan2025deepwidesearchbenchmarkingdepthwidth}。
近期工作进一步缩小了这一差距。\citet{tao2026mmsearchplusbenchmarkingprovenanceawaresearch} 强调根据细粒度视觉线索进行来源感知浏览， \citet{zhang2026browsecompv3visualverticalverifiable} 则为开放网页多模态浏览引入经专家验证的子目标和子目标覆盖率过程分数。\citet{gupta2026deepsearchqabridgingcomprehensivenessgap} 通过客观可验证的单一答案或集合答案，评估基于因果链的开放网页研究，但仍对结果而非中间链路进行评分。\citet{ning2026mcsearchevaluatingenhancingmultimodal} 提供逐步标注的多模态推理图和过程指标，但其检索范围是固定的本地知识库。面向报告的基准研究的是另一种输出形态。 \citet{huang2026mmdeepresearchbenchbenchmarkmultimodaldeep} 评估带引用支撑的多模态研究报告，\citet{ye2026miroevalbenchmarkingmultimodaldeep} 则在纯文本和多模态任务上联合评估报告质量、事实性和系统所暴露的研究轨迹。\ourmethodname 对这些设定形成补充，将开放网页搜索、必需的真实世界多模态证据、简短可验证答案、long-horizon Node-Relation 结构和 response-level dependency-aware checklist evaluation 结合起来。这里的 horizon 指在具有前置依赖的结论之间持续维护研究状态，而非生成长篇输出、处理长上下文输入或执行预先规定数量的浏览操作。与无序 checklist 或子目标覆盖率不同，DACS 使用逐题 Directed Acyclic Graph (DAG)。必要的前置结论不满足时，它不会给予下游结论分数。与依赖轨迹或固定语料库的过程评估不同，DACS 仅根据异构系统提交的回复进行计算，但不声称观察其隐式推理或工具调用轨迹。表~\ref{tab:dataset_comparison} 从检索环境、多模态要求、输出形式、标准中间结构和前置依赖门控评分五个维度，对比 \ourmethodname 与代表性的短答案浏览和结构化搜索基准。
}

\begin{table}[t]
\centering
\scriptsize
\setlength{\tabcolsep}{2.5pt}
\caption{Comparison with representative browsing and structured-search benchmarks. ``Gold structure'' denotes task-level supervision of intermediate conclusions or retrieval steps; ``prerequisite-gated'' indicates that a downstream step receives credit only when its required predecessors are also satisfied. ``Open'' denotes the open web and ``local KB'' a fixed retrieval corpus.\zh{ 与代表性 browsing 和 structured-search benchmark 的对比。``Gold structure'' 表示对中间结论或检索步骤的 task-level supervision；``prerequisite-gated'' 表示下游步骤只有在其必要前置步骤也满足时才获得分数。``Open'' 表示开放网页，``local KB'' 表示固定检索语料库。}}
\label{tab:dataset_comparison}
\resizebox{\linewidth}{!}{%
\begin{tabular}{lccccc}
\toprule
Benchmark & Retrieval & MM required & Output & Gold structure & Prerequisite-gated \\
\midrule
BrowseComp~\citep{wei2025browsecomp} & Open & $\times$ & Short & -- & $\times$ \\
MM-BrowseComp~\citep{li2025mmbrowsecompcomprehensivebenchmarkmultimodal} & Open & \checkmark & Short & Checklist & $\times$ \\
MMSearch-Plus~\citep{tao2026mmsearchplusbenchmarkingprovenanceawaresearch} & Open & \checkmark & Short & -- & $\times$ \\
BrowseComp-$V^3$~\citep{zhang2026browsecompv3visualverticalverifiable} & Open & \checkmark & Short & Subgoals & $\times$ \\
DeepSearchQA~\citep{gupta2026deepsearchqabridgingcomprehensivenessgap} & Open & $\times$ & Short/set & -- & $\times$ \\
MC-Search~\citep{ning2026mcsearchevaluatingenhancingmultimodal} & Local KB & Mixed & QA & Step-wise graph & $\times$ \\
\ourmethodname & Open & \checkmark & Short & Checklist DAG & \checkmark \\
\bottomrule
\end{tabular}
}%
\end{table}

\section{\ourmethodname Benchmark}

\subsection{Task Definition and Annotation Schema}

During evaluation, the model sees only the natural-language question and its associated image evidence; the Node-Relation graph, checklist, sources, and reference answer remain hidden.

\zh{
评测时，模型只能看到自然语言问题及其关联图像证据；Node-Relation graph、checklist、sources 和 reference answer 均保持隐藏。
}

\ourmethodname is a real-world multimodal deep research benchmark constructed by a six-person annotation team of master's- and PhD-level AI researchers. Its data span eight topical categories: media, people, organizations, geography, society, academia, technology, and sports. Each item also retains one coarse authored-template label, Linear or Multi-branch, and one or more operation tags from Constraint, Symbolic, Temporal, and Numerical. The template label records the intended construction pattern; it does not define the exact topology of the subsequently verified dependency DAG. Beyond these labels, each item contains a natural-language question, a short verifiable answer, source evidence, image or other non-text evidence, and an irreducible checklist of necessary intermediate conclusions. This schema supports both final-answer evaluation and post-hoc response analysis: the answer field supports answer-level scoring, the checklist supports assessment of whether the submitted response correctly states the necessary conclusions, and the evidence fields support task verification.

\zh{
\ourmethodname 是一个由六名硕士和博士层级 AI 研究人员组成的标注团队构建的真实世界多模态深度研究基准。其数据涵盖八个主题类别：media、people、organizations、geography、society、academia、technology 和 sports。每个条目还保留一个粗粒度 authored-template 标签，即 Linear 或 Multi-branch，并从 Constraint、Symbolic、Temporal 和 Numerical 中标注一个或多个 operation tag。该模板标签记录预期的构造方式，但不定义后续核定的依赖 DAG 的精确拓扑。除这些标签外，每个条目还包含自然语言问题、简短的可验证答案、来源证据、图像或其他非文本证据，以及由必要中间结论组成的不可约 checklist。该 schema 同时支持最终答案评测和事后回复分析：answer 字段用于答案层面的评分，checklist 用于评估提交回复是否正确陈述了必要结论，evidence 字段则用于任务验证。
}

The eight categories cover common real-world research settings: cultural media, public figures, organizations, geospatial reasoning, social and historical events, academic artifacts, technologies, and sports. Linear denotes an item authored around a predominantly sequential route, whereas Multi-branch denotes one intentionally authored around multiple evidence routes whose partial conclusions are later combined. These labels summarize construction intent. The verified checklist DAG is the authoritative representation of actual dependencies and may exhibit additional partial independence or local branching under either template. The operation tags describe local pressures within the chain rather than independent task families. Constraint marks candidate filtering or uniqueness conditions; Symbolic covers logos, abbreviations, names, colors, symbols, and format conversions; Temporal covers chronological ordering and relative-time constraints; Numerical covers numbers, rankings, coordinates, statistics, and unit calculations. Together, these labels support analysis of both an item's topic and the locations of structural pressure and demanding local operations within its research chain.

\zh{
八个类别覆盖真实研究中的常见场景，包括文化媒体、公众人物、组织机构、地理空间推理、社会和历史事件、学术成果与资料、技术以及体育。Linear 表示条目主要围绕顺序推进的路线构造；Multi-branch 表示条目在构造时有意设置多条证据路线，并在后续合并使用其局部结论。这些标签概括的是构造意图。经核定的 checklist DAG 才是实际依赖关系的权威表示，两种模板下都可能出现额外的局部独立关系或分支。Operation tag 描述的是链路内部的局部压力，而不是彼此独立的任务类型。Constraint 表示候选集过滤或唯一性约束；Symbolic 覆盖 logo、缩写、名称、颜色、符号和格式转换；Temporal 覆盖时间排序和相对时间约束；Numerical 覆盖数字、排名、坐标、统计和单位计算。这些标签支持分析条目的主题，以及结构压力和高要求局部操作在研究链中的位置。
}

\paragraph{Illustrative example.}

The case study in Appendix Table~\ref{tab:case_study_part3} provides a compact example. The intended chain identifies the pictured phone as an Apple iPhone 15 Pro Max, links it to the A17 Pro and the Apple Silicon team's relationship with Synopsys, and traces the phone's titanium frame to the A-12 and SR-71 aircraft. The branch corresponding to the later-retired aircraft proceeds through the SR-71 Flight Simulator to the Frontiers of Flight Museum, Smithsonian Affiliations, and the Smithsonian Institution. A U.S. Department of State video then supplies the first word, ``climate,'' on the third interviewee's shirt. Initially misidentifying the phone as a Samsung device redirects the processor, aircraft, and museum links; later guesses about Smithsonian-related nodes cannot repair the unsupported chain. Without reproducing the full case table, this example shows why early visual grounding and preservation of intermediate entities matter.

\zh{
Appendix Table~\ref{tab:case_study_part3} 中的案例提供了一个简洁示例。预期链路首先将图中手机识别为 Apple iPhone 15 Pro Max，继而关联到 A17 Pro 以及 Apple Silicon 团队与 Synopsys 的关系，再从手机的钛金属边框追踪到 A-12 和 SR-71 两种飞机。对应较晚退役机型的分支经由 SR-71 Flight Simulator 指向 Frontiers of Flight Museum、Smithsonian Affiliations 与 Smithsonian Institution。随后，从美国国务院的一段视频中读出第三位受访者衬衫上的首个单词 ``climate''。起初将手机误识别为 Samsung 设备，会使处理器、飞机和博物馆等后续连接偏离正确方向；之后对 Smithsonian 相关节点的猜测也无法修复缺乏支撑的链路。该示例无需复制完整案例表，即可说明早期基于视觉证据的实体识别与中间实体保留的重要性。
}

\subsection{Benchmark Construction}

\subsubsection{Node-Relation Design Principles}

Annotators do not begin \ourmethodname items by writing a natural-language question. Instead, they first construct a Node-Relation graph. A node pairs a core entity with searchable and verifiable properties, such as a person photo, landmark image, map location, paper PDF, official document, poster, logo, chart, timestamp, identity, work, affiliation, venue, or founder. Each item typically begins with five to six information-rich nodes. Nodes are selected not merely for searchable names, but for properties that support later verification and cross-modal bridging.

\zh{
标注者并非先撰写自然语言问题来构造 \ourmethodname 条目，而是先构建一个 Node-Relation graph。Node 将核心实体与可检索、可验证的属性配对，例如人物照片、地标图片、地图位置、论文 PDF、官方文档、海报、徽标、图表、时间戳、身份、作品、机构归属、场所或创始人。每道题通常从五到六个信息丰富的节点开始设计。选择节点不仅要考虑其名称是否可搜索，还要考虑其属性能否支撑后续验证和跨模态桥接。
}

A relation turns a verified property of one node into an entry condition for the next. Relations may take the form of shared attributes, such as the same location, organization, person, event, or work, or uniqueness constraints, such as a relative year, a founder relation, the first host city of an event, the sister city of a birthplace, or an organization indicated by a visual symbol. A valid relation must be natural and uniquely convergent: removing any key relation should visibly break the chain rather than merely shorten it.

\zh{
Relation 将一个节点中已验证的属性转化为下一个节点的入口条件。Relation 可以表现为共享属性，例如同一地点、同一机构、同一人物、同一事件或同一作品；也可以表现为唯一性约束，例如相对年份、创始人关系、某事件首次举办城市、出生地对应的友好城市，或由视觉符号指向的组织。有效 relation 必须自然且唯一收敛：删除任一关键 relation 后，链路应明显断裂，而非仅仅变短。
}


The long-horizon difficulty in \ourmethodname arises from dependency structure rather than verbose prompts, long outputs, or obscure trivia. Long-horizon requires preserving an evolving research state across dependent evidence steps: a conclusion established at one step becomes a constraint or entry condition for later investigation. Therefore, the final question should not be answerable through a single query, a single webpage, or a direct search for its final wording. Later steps must depend on earlier intermediate conclusions, requiring the agent to retain the research objective, resolved entities, and accumulated constraints across many hops. This design motivates checklist-based response evaluation: final-answer correctness alone can obscure whether the submitted response correctly states the necessary intermediate conclusions and their prerequisites.

\zh{
\ourmethodname 的 long-horizon 难度来自依赖结构，而非冗长题面、长输出或冷门事实。Long-horizon 要求在相互依赖的证据步骤中持续维护不断演化的研究状态：某一步得到的结论会成为后续调查的约束或入口条件。因此，最终问题不应能通过单次查询、单一网页或直接搜索最终题面措辞得到答案。后续步骤必须依赖前序中间结论，要求 agent 在多跳过程中持续保持研究目标、已确定实体和累积约束。这一设计也使基于 checklist 的回复评估成为必要：仅看最终答案正确性会掩盖提交回复是否正确陈述了必要中间结论及其前置依赖。
}

Multimodal evidence is a required part of the reasoning chain. The entry node is preferably grounded in a real-world non-text modality, such as an image, map, PDF, video screenshot, logo, product image, chart, or table. Each \ourmethodname item must contain non-text evidence that changes the reasoning state by identifying an entity, constraining candidates, confirming a visual relation, localizing a map object, or extracting information from a document, chart, or table.

\zh{
多模态证据是推理链的必要组成部分。入口节点优先由真实世界的非文本模态支撑，例如图片、地图、PDF、视频截图、logo、产品图、图表或表格。每个 \ourmethodname 条目都必须包含能够改变推理状态的非文本证据，例如用于识别实体、约束候选集、确认视觉关系、定位地图对象，或从文档、图表或表格中抽取信息的证据。
}

\subsubsection{AI-Assisted Construction Workflow}

\begin{figure}[t]
    \centering
    \includegraphics[width=\linewidth]{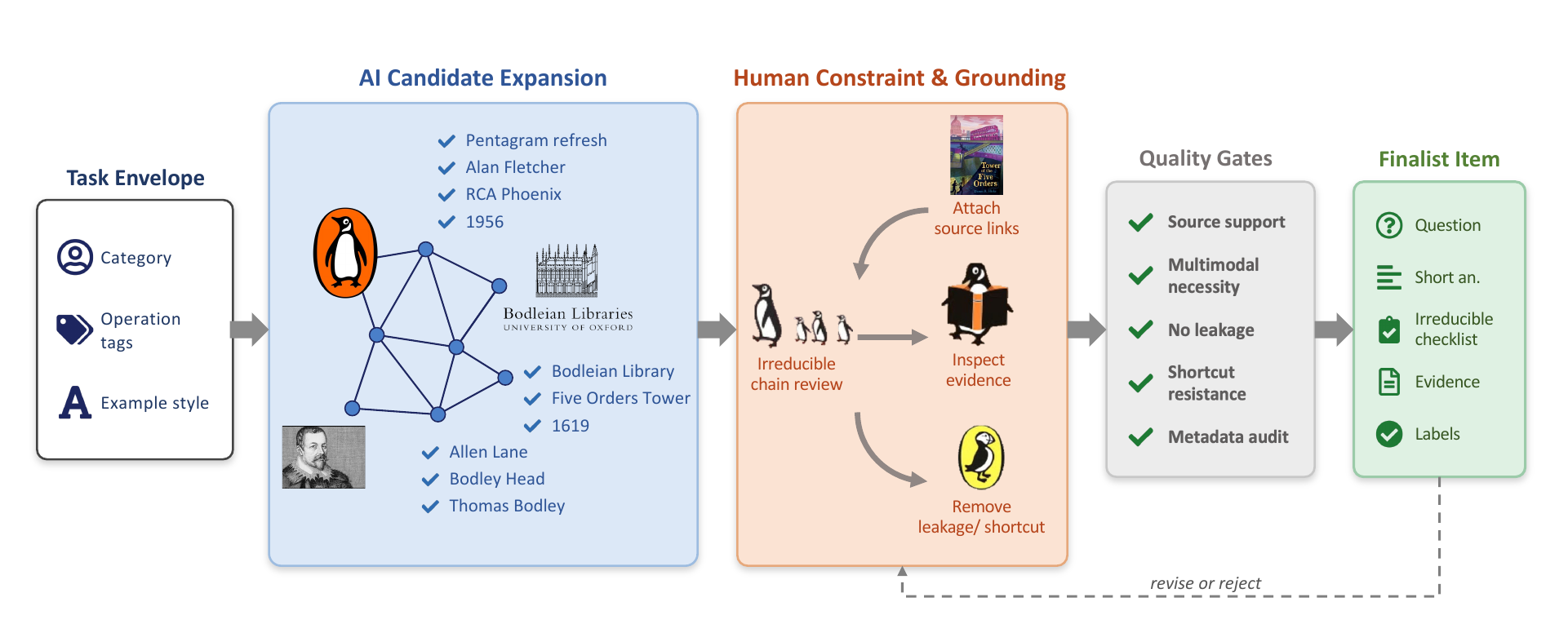}
    \caption{The overall construction pipeline for \ourmethodname. AI helps explore the candidate design space through Node-Relation graphs, over-complete checklists, draft fields, and question polishing. Human annotators then prune, revise, and validate the candidates, locking only items that pass source, checklist, multimodal, shortcut, and metadata checks.\zh{ \ourmethodname 的 AI-assisted 整体构造流程。AI 通过 Node-Relation graph、超量候选 checklist、字段草稿和题面润色来协助探索候选设计空间。人工标注者随后对候选内容进行剪枝、修订和核验，并只锁定通过 source、checklist、多模态、shortcut 和 metadata 检查的条目。}}
    \label{fig:ai_assisted_pipeline}
\end{figure}

Figure~\ref{fig:ai_assisted_pipeline} summarizes the AI-assisted construction workflow. AI serves as a candidate-expansion tool rather than an authority. Annotators first specify the target category, operation tags, and golden-example style, then use models to propose Node-Relation graphs, possible multimodal sources, and over-complete checklists. Human annotators prune these candidates into irreducible evidence chains, reject any wording or attribute ambiguity that undermines answer uniqueness or source support, and draft the final question, answer, sources, images, and metadata. Final admission remains under human control through checks of factual support, multimodal necessity, shortcut resistance, label correctness, and benchmark coverage. Detailed pipeline stages and annotation decisions are described in Appendix~\ref{sec:appendix_ai_assisted_pipeline}.

\zh{
Figure~\ref{fig:ai_assisted_pipeline} 总结了 AI-assisted construction workflow。AI 用作候选扩展工具，而非事实权威。标注者首先指定目标 category、operation tag 和 golden-example 风格，再使用模型提出 Node-Relation graph、潜在多模态来源和超量候选 checklist。人工标注者随后将这些候选内容剪枝为不可约证据链，拒绝任何会损害答案唯一性或来源支撑的模糊措辞或属性，并起草最终 question、answer、sources、images 和 metadata。最终准入仍由人工控制，并通过事实支撑、多模态必要性、捷径抵抗性、标签正确性和 benchmark 覆盖度检查完成。更详细的 pipeline 阶段和标注决策见 Appendix~\ref{sec:appendix_ai_assisted_pipeline}。
}

\subsubsection{Quality Standards and Validation Protocol}
\label{sec:quality_control}

\paragraph{Quality criteria.}

Candidate items must jointly satisfy four quality standards. First, the target answer must be concise, inspectable, unique under open-web evidence, and stable within the question's temporal scope. Second, the checklist must form an irreducible chain of verifiable conclusions rather than browsing actions. Every key node, bridge relation, checklist entry, and final answer requires inspectable source support, and removing any necessary step must break reliable derivation. Third, the question must hide intermediate nodes and resist direct-answer, single-query, single-page, and text-only shortcuts. Fourth, non-text evidence counts only if it changes the reasoning state by identifying an entity, constraining candidates, localizing an object, reading a document or chart, or bridging two nodes.

\zh{
候选条目必须同时满足四项质量标准。第一，目标答案必须简短、可检查，在开放网页证据下具有唯一性，并在题面限定的时间范围内保持稳定。第二，checklist 必须构成由可验证结论而非浏览动作组成的不可约链路。每个关键 node、bridge relation、checklist entry 和最终答案都需要可检查的 source 支撑，删除任一必要步骤都必须使可靠推导中断。第三，题面必须隐藏中间节点，并抵抗 direct-answer、single-query、single-page 和 text-only shortcut。第四，非文本证据只有在通过识别实体、约束候选、定位对象、读取文档或图表，或桥接两个节点来改变推理状态时才计入。
}

\paragraph{Validation protocol.}

We enforce these standards through a four-stage validation protocol.

\zh{
这些标准通过一个四阶段验证流程落实。
}

\textbf{Phase 1: Pilot and calibration.} Each of the six annotators first constructs a small pilot batch from golden examples. The core team audits graph naturalness, unique convergence, path leakage, checklist form, multimodal necessity, and answer format, aligning annotators on the boundary of admissible long-horizon tasks.

\zh{
\textbf{Phase 1: Pilot and calibration.} 六名标注者首先基于 golden examples 分别构造一小批 pilot 样本。核心团队检查 graph 自然性、唯一收敛、路径泄露、checklist 形式、多模态必要性和答案格式，以统一标注者对可接受 long-horizon task 边界的认识。
}

\textbf{Phase 2: Full-scale construction and structured secondary audit.} A separate auditor checks every field in a fixed order: whether the hidden chain can be skipped, whether the question reveals the path, whether the checklist is minimal, and whether sources support every key node, relation, conclusion, and answer. Items with unsupported conclusions, multiple valid answers, or reducible checklists are returned for revision.

\zh{
\textbf{Phase 2: Full-scale construction and structured secondary audit.} 独立审核者按固定顺序逐字段检查：隐藏链路能否被跳过、题面是否泄露路径、checklist 是否最小，以及 sources 是否支撑每个关键 node、relation、结论和答案。存在无支撑结论、多个有效答案或可约 checklist 的条目必须退回修改。
}

\textbf{Phase 3: Multimodal necessity and shortcut check.} Auditors verify that at least one key conclusion depends on non-text evidence and that this evidence participates in the chain rather than serving as decoration. They then submit the final question verbatim to a web-enabled model without providing the hidden graph, checklist, sources, or extra prompts. If a single response recovers the answer or a reproducible direct-search path, the item is rewritten, constrained further, or rejected.

\zh{
\textbf{Phase 3: Multimodal necessity and shortcut check.} 审核者确认至少一个关键结论依赖非文本证据，且该证据实际参与链路而非仅作装饰。随后，在不提供隐藏 graph、checklist、sources 或额外 prompt 的条件下，将最终题面原样提交给具备网页搜索能力的模型。若单次回复给出了答案或一条可复现的直接搜索路径，条目必须被改写、加强约束或拒绝。
}

\textbf{Phase 4: Factual and metadata audit.} Before locking an item, the core team verifies every URL and corresponding claim, checks that the chain is not dominated by Wikipedia or a single text-only page, confirms the presence of inspectable non-text evidence, and audits category, template, tag, image, and source metadata. Only items that pass every gate enter evaluation.

\zh{
\textbf{Phase 4: Factual and metadata audit.} 锁定条目前，核心团队核验每个 URL 及其对应主张，检查链路未被 Wikipedia 或单一纯文本页面主导，确认存在可检查的非文本证据，并审核 category、template、tag、image 和 source metadata。只有通过全部质量门的条目才进入评测。
}

\begin{table}[t]
\centering
\scriptsize
\setlength{\tabcolsep}{3pt}
\caption{Quality gates applied before an item enters the \ourmethodname evaluation set. These gates operationalize answer uniqueness, no path leakage, irreducibility, source support, multimodal necessity, shortcut filtering, and metadata consistency.\zh{ 条目进入 \ourmethodname 评测集前必须通过的质量门，具体落实答案唯一性、无路径泄露、不可约性、来源支撑、多模态必要性、捷径过滤和元数据一致性。}}
\label{tab:quality_gates}
\begin{tabular}{p{0.19\linewidth}p{0.31\linewidth}p{0.28\linewidth}p{0.13\linewidth}}
\toprule
Gate & What is checked & Failure condition & Action \\
\midrule
Answer uniqueness & The final answer is short, stable, unique, and verifiable. & Multiple equally valid answers or time-varying answers without a temporal constraint. & Revise or reject. \\
No path leakage & The question hides intermediate nodes, relations, checklist steps, and the answer. & The question can be followed as an explicit step list or directly searched as written. & Rewrite. \\
Irreducible checklist & Each checklist item is a necessary intermediate conclusion, not a browsing action. & Removing a step still leaves the answer derivable, or a step is merely operational. & Prune or rewrite. \\
Source support & Sources support key nodes, relations, checklist items, and the final answer. & A key claim lacks an inspectable URL or is supported only by a single fragile source. & Add evidence or reject. \\
Multimodal necessity & Non-text evidence changes the reasoning state. & The image, map, PDF, video, table, or chart is decorative or bypassable by text-only search. & Revise or reject. \\
Shortcut filtering & Strong web-enabled models cannot solve the item in one direct query. & The reference answer or full path is recovered by direct search. & Rewrite, strengthen constraints, or reject. \\
Metadata audit & Category, authored-template label, operation tags, image fields, and source fields are consistent. & Metadata is missing, inconsistent, or leaking. & Fix before locking. \\
\bottomrule
\end{tabular}
\end{table}

\subsection{Benchmark Statistics}

\begin{figure}[t]
\centering
\includegraphics[width=\linewidth]{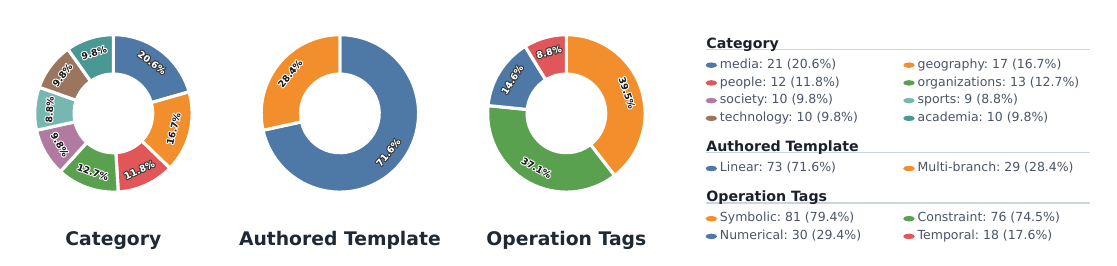}
\caption{Label distribution in \ourmethodname: categories, Linear/Multi-branch authored templates, and operation tags. Operation tags are multi-label, so their percentages need not sum to 100\%.\zh{ \ourmethodname 的标签分布：category、Linear/Multi-branch authored template 和 operation tag。Operation tag 是多标签，因此比例之和不必为 100\%。}}
\label{fig:dataset_labels}
\end{figure}

\ourmethodname contains 102 locked evaluation items, each supported by a multi-step checklist and multiple evidence sources. The author-verified checklist graphs exhibit substantial dependency depth, and parallel branches further increase the overall evidential workload. The benchmark spans eight real-world research categories, includes both Linear and Multi-branch authored templates, and covers symbolic, constraint-based, numerical, and temporal reasoning. Figure~\ref{fig:dataset_labels} summarizes the label distributions. Detailed checklist, dependency-depth, source, image, non-text-evidence, and split statistics are reported in Appendix~\ref{sec:appendix_dataset_statistics}.

\zh{
\ourmethodname 包含 102 个锁定的评测条目，每个条目均配有多步 checklist 和多项证据来源。经作者核定的 checklist graph 具有较深的依赖链，并行分支进一步增加了整体证据工作量。该基准覆盖八类真实研究主题，兼具 Linear 与 Multi-branch 两种 authored template，并涉及符号、约束、数值和时序推理。Figure~\ref{fig:dataset_labels} 总结了标签分布。详细的 checklist、dependency depth、source、image、non-text evidence 和 split 统计见 Appendix~\ref{sec:appendix_dataset_statistics}。
}

\section{Experiments}
\label{sec:experiments}



\subsection{Experimental Setup}

\subsubsection{Evaluation Set and Coverage}

All experiments use the fixed \ourmethodname evaluation set, which contains 102 benchmark items. Each item includes a natural-language question, a short reference answer, source evidence, an image or other non-text evidence, category and structure labels, operation tags, and an irreducible checklist of necessary intermediate conclusions. During evaluation, models receive the question and, when available, its associated image evidence.

\zh{
所有实验均使用固定的 \ourmethodname evaluation set，其中包含 102 个 benchmark item。每个 item 包含自然语言问题、简短 reference answer、source evidence、图像或其他非文本证据、category 和 structure 标签、operation tags，以及由必要中间结论组成的不可约 checklist。评测时，模型会收到问题，以及在可用时与问题关联的图像证据。
}

The evaluation set covers all eight \ourmethodname categories and both coarse authored-template labels. It contains 73 Linear-template items and 29 Multi-branch-template items. All 102 items have checklist annotations, comprising 1,231 checklist steps in total. These annotations define the denominators used to compute SA, CS, and DACS in the experiments below.

\zh{
该 evaluation set 覆盖 \ourmethodname 的全部八个 category 和两种粗粒度 authored-template 标签，其中包含 73 个 Linear-template item 和 29 个 Multi-branch-template item。102 个 item 均有 checklist annotation，总计包含 1,231 个 checklist step。这些 annotation 定义了后续实验计算 SA、CS 和 DACS 时使用的分母。
}

\subsubsection{Model Groups}

We evaluate \BREWmainNsystems{} systems, divided into four groups according to the research interface and tool access used during evaluation. Tool-augmented VLMs receive image input and access to provider-exposed web search, whereas tool-free VLMs receive image input but no search tool. The tool-free label refers to the evaluation configuration, not an inherent limitation of the model. Deep Research Systems use dedicated research endpoints, while framework-based agents run through DeerFlow~\citep{bytedance2026deerflow}. Each evaluation consists of a single request. Any tool use or multi-step behavior occurs within the provider's service or the framework, rather than in an agent loop that we built. Table~\ref{tab:model_scope_summary} summarizes these groups, and Appendix Table~\ref{tab:appendix_model_configuration} lists every model, route, and configuration.

\zh{
我们共评测 \BREWmainNsystems{} 个系统，并根据评测时使用的 research interface 和工具访问能力将其分为四组。Tool-augmented VLM 接收图像输入，并可使用 provider 提供的网页搜索；tool-free VLM 接收图像输入，但不配备搜索工具。Tool-free 这一名称描述的是评测配置，而非模型的固有能力限制。Deep Research Systems 使用专用 research endpoint，framework-based agents 则通过 DeerFlow~\citep{bytedance2026deerflow} 运行。每次评测仅包含一次请求。任何工具调用或多步行为都发生在 provider 服务或框架内部，而非我们自建的 agent 循环中。表~\ref{tab:model_scope_summary} 汇总这些分组，附录表~\ref{tab:appendix_model_configuration} 列出全部模型、路由和配置。
}

\begin{table}[t]
\centering
\small
\setlength{\tabcolsep}{6pt}
\caption{Summary of the model groups evaluated in \ourmethodname. The full model configuration is reported in the appendix.\zh{ \ourmethodname 评测中模型分组的概览。完整模型配置见 appendix。}}
\label{tab:model_scope_summary}
\begin{tabular}{@{}llr@{}}
\toprule
Group & Capability & Models \\
\midrule
Tool-augmented VLMs & image input + web search & 11 \\
Tool-free VLMs & image input only & 7 \\
Deep Research Systems & dedicated research endpoint & 4 \\
Framework-based agents & DeerFlow local agent framework & 3 \\
\bottomrule
\end{tabular}
\end{table}

\subsubsection{Metrics}
\label{sec:metrics}

We evaluate \ourmethodname along two complementary dimensions: final-answer correctness and the dependency consistency of the conclusions stated in a response. For a fixed model and item $i$, let $y_i\in\{0,1\}$ denote the judge's binary decision on final-answer correctness. Let $G_i=(V_i,E_i)$ denote the verified checklist dependency DAG, with its $m_i=|V_i|$ nodes indexed by $j$. Finally, let $a_{ij}\in\{0,1\}$ indicate whether the response correctly states checklist conclusion $j$ for item $i$. Over the fixed set of $n=102$ items, Overall Accuracy (OA), Strict Accuracy (SA), and Checklist Score (CS) are defined in Equations~\ref{eq:oa}--\ref{eq:cs}:

\zh{
我们从两个互补维度评估 \ourmethodname：最终答案正确性，以及回复中已陈述结论的依赖一致性。对于固定模型和 item $i$，令 $y_i\in\{0,1\}$ 表示 judge 对最终答案正确性作出的二元判定。令 $G_i=(V_i,E_i)$ 表示经核定的 checklist dependency DAG，并用 $j$ 索引其 $m_i=|V_i|$ 个节点。最后，令 $a_{ij}\in\{0,1\}$ 表示回复是否正确陈述了 item $i$ 的第 $j$ 个 checklist 结论。在固定的 $n=102$ 个 item 上，Overall Accuracy (OA)、Strict Accuracy (SA) 和 Checklist Score (CS) 分别由公式~\ref{eq:oa}--\ref{eq:cs} 定义：
}

\begin{align}
\mathrm{OA}
&= \frac{1}{n}\sum_{i=1}^{n}y_i,
\label{eq:oa}\\
\mathrm{SA}
&= \frac{1}{n}\sum_{i=1}^{n}y_i\prod_{j=1}^{m_i}a_{ij},
\label{eq:sa}\\
\mathrm{CS}
&= \frac{1}{n}\sum_{i=1}^{n}\frac{1}{m_i}\sum_{j=1}^{m_i}a_{ij}.
\label{eq:cs}
\end{align}

Thus, OA measures final-answer correctness. SA additionally requires the response to state every annotated necessary conclusion correctly. CS measures average checklist coverage without accounting for dependencies and is retained as an auxiliary metric.

\zh{
因此，OA 衡量最终答案正确性。SA 除要求最终答案正确外，还要求回复正确陈述全部已标注的必要结论。CS 衡量不考虑依赖关系时各 item 的 checklist coverage 平均值，并作为辅助指标保留。
}

Our primary process metric is the Dependency-Aware Checklist Score (DACS). Let $D_{ij}\subseteq V_i$ denote the direct prerequisites of node $j$ in item $i$. We recursively define its dependency-grounded credit $s_{ij}$ and the resulting DACS in Equations~\ref{eq:dependency-grounded-credit} and~\ref{eq:dacs}, respectively:

\zh{
我们的主要过程指标是 Dependency-Aware Checklist Score (DACS)。令 $D_{ij}\subseteq V_i$ 表示 item $i$ 中节点 $j$ 的直接前置依赖。其依赖支撑得分 $s_{ij}$ 和相应的 DACS 分别由公式~\ref{eq:dependency-grounded-credit} 和公式~\ref{eq:dacs} 定义：
}

\begin{align}
s_{ij}
&= a_{ij}\prod_{k\in D_{ij}}s_{ik},
\label{eq:dependency-grounded-credit}\\
\mathrm{DACS}
&= \frac{1}{n}\sum_{i=1}^{n}\frac{1}{m_i}\sum_{j=1}^{m_i}s_{ij}.
\label{eq:dacs}
\end{align}

where an empty product equals one. A node therefore receives credit only if both the node itself and all of its transitive prerequisites are satisfied. This recursion preserves credit for independent branches but withholds credit from a conclusion if any required ancestor is missing. DACS thus measures the dependency consistency of stated conclusions.

\zh{
其中空乘积等于一。因此，只有当节点本身及其全部传递前置依赖均满足时，该节点才能得分。这一递归会保留独立分支的得分，但如果缺少任一必要祖先，则不给相应结论计分。因此，DACS 衡量的是已陈述结论的依赖一致性。
}

\subsubsection{LLM Judge}
\label{sec:judge_reliability}

The LLM Judge compares each model response against the reference answer and checklist. It outputs a binary decision on final-answer correctness and a checklist completion vector. OA, SA, CS, and DACS are then computed deterministically from these outputs and the frozen dependency graphs, using the fixed denominator of 102 items. Failed or missing runs are assigned an incorrect final-answer decision and an all-zero checklist vector. The current evaluation pipeline uses Qwen3-VL-235B (Qwen/Qwen3-VL-235B-A22B-Instruct-FP8) as the judge model.

\zh{
LLM Judge 会将每个模型回复与 reference answer 和 checklist 进行比较。它输出一个关于最终答案正确性的二元判定和一个 checklist completion vector。随后，OA、SA、CS 和 DACS 根据这些输出与冻结的依赖图，在固定的 102 题分母下确定性计算。失败或缺失的运行会被判为最终答案错误，并分配全零 checklist vector。当前评测流程使用 Qwen3-VL-235B (Qwen/Qwen3-VL-235B-A22B-Instruct-FP8) 作为 judge model。
}

\subsection{Main Results}

\subsubsection{Overall Capability}

\ourmethodname remains challenging even for strong systems. As shown in Table~\ref{tab:main_results}, GPT-5.5 has the highest OA, with \BREWgptfiveOA\% OA, 34.3\% SA, 73.6\% CS, and 70.3\% DACS. These results remain far from saturation. Even this run solves only slightly more than two fifths of the benchmark items at the final-answer level and just over one third under the strict criterion that requires both the answer and all annotated necessary conclusions to be correct. Among dedicated deep research systems, o3 Deep Research has the highest OA, reaching 32.4\% OA, 19.6\% SA, 56.1\% CS, and 49.8\% DACS. These values are the highest OA point estimates, not a statistically separable ranking. With only 102 items, the confidence intervals of adjacent systems overlap substantially, and Appendix~\ref{sec:statistical_uncertainty} shows that most per-model differences are not statistically separable. Framework-based agents are substantially weaker in this evaluation: DeerFlow using Qwen3-VL-235B reaches 15.7\% OA, 9.8\% SA, 41.8\% CS, and 38.5\% DACS.

\zh{
\ourmethodname 对强系统而言仍然具有挑战性。如表~\ref{tab:main_results} 所示，GPT-5.5 的 OA 最高，达到 \BREWgptfiveOA\% OA、34.3\% SA、73.6\% CS 和 70.3\% DACS。这些结果距离饱和仍然很远。即使是该运行，在最终答案层面也仅解出略多于五分之二的 benchmark item；在要求答案与所有已标注必要结论均正确的严格口径下，则仅解出略多于三分之一。在专用 deep research system 中，o3 Deep Research 的 OA 最高，达到 32.4\% OA、19.6\% SA、56.1\% CS 和 49.8\% DACS。这些数值只是最高的 OA 点估计，并不构成统计上可区分的系统排名。在仅有 102 道题的情况下，相邻系统的置信区间大幅重叠；Appendix~\ref{sec:statistical_uncertainty} 进一步表明，多数 per-model 差异在统计上不可区分。Framework-based agent 在这一评测中明显更弱：使用 Qwen3-VL-235B 的 DeerFlow 达到 15.7\% OA、9.8\% SA、41.8\% CS 和 38.5\% DACS。
}

\begin{table*}[t]
\centering
\scriptsize
\setlength{\tabcolsep}{2pt}
\caption{Main \ourmethodname results for all 25 systems in the current evaluation scope, grouped by model capability. Overall columns report final-answer accuracy (OA), strict answer-and-process accuracy (SA), checklist score (CS), and dependency-aware checklist score (DACS). Category columns report diagnostic SA point estimates for each topical category. All metrics are percentages computed using Qwen3-VL-235B (Qwen/Qwen3-VL-235B-A22B-Instruct-FP8) as the judge.\zh{ 当前评测范围内全部 25 个系统的 \ourmethodname 主结果，并按模型能力分组。Overall 列报告 final-answer accuracy (OA)、strict answer-and-process accuracy (SA)、checklist score (CS) 和 dependency-aware checklist score (DACS)。Category 列报告各主题类别的诊断性 SA 点估计。所有指标均为百分比，并使用 Qwen3-VL-235B (Qwen/Qwen3-VL-235B-A22B-Instruct-FP8) 作为 judge 计算。}}
\label{tab:main_results}
\resizebox{\linewidth}{!}{%
\begin{tabular}{lrrrr|rrrrrrrr}
\toprule
 & \multicolumn{4}{c}{Overall} & \multicolumn{8}{c}{SA by category} \\
\cmidrule(lr){2-5}\cmidrule(lr){6-13}
Model & OA (\%) & SA (\%) & CS (\%) & DACS (\%) & Ac. & Geo. & Med. & Org. & Ppl. & Soc. & Sport & Tech. \\
\midrule
\multicolumn{13}{c}{\textbf{Tool-augmented VLMs}} \\
\cmidrule(lr){1-13}
GPT-5.5 & 43.1 & 34.3 & 73.6 & 70.3 & 50.0 & 17.6 & 42.9 & 46.2 & 41.7 & 30.0 & 22.2 & 20.0 \\
o4-mini & 35.3 & 20.6 & 49.9 & 44.9 & 20.0 & 17.6 & 23.8 & 15.4 & 33.3 & 20.0 & 22.2 & 10.0 \\
o3 & 34.3 & 15.7 & 39.4 & 34.1 & 50.0 & 11.8 & 19.0 & 0.0 & 25.0 & 10.0 & 0.0 & 10.0 \\
GPT-5.4 & 30.4 & 22.5 & 70.1 & 63.9 & 20.0 & 23.5 & 23.8 & 30.8 & 25.0 & 30.0 & 11.1 & 10.0 \\
Claude Opus 4.7 & 29.4 & 27.5 & 69.1 & 65.7 & 30.0 & 23.5 & 52.4 & 23.1 & 33.3 & 20.0 & 0.0 & 10.0 \\
Grok 4.20 & 22.5 & 17.6 & 69.3 & 62.4 & 40.0 & 11.8 & 28.6 & 15.4 & 16.7 & 20.0 & 0.0 & 0.0 \\
GPT-4o (2024-11) & 18.6 & 12.7 & 42.6 & 37.3 & 20.0 & 5.9 & 19.0 & 7.7 & 25.0 & 10.0 & 0.0 & 10.0 \\
Claude Sonnet 4.6 & 17.6 & 15.7 & 64.5 & 61.5 & 30.0 & 5.9 & 14.3 & 15.4 & 16.7 & 30.0 & 11.1 & 10.0 \\
GPT-4.1 & 16.7 & 13.7 & 48.4 & 41.7 & 20.0 & 5.9 & 33.3 & 0.0 & 8.3 & 10.0 & 11.1 & 10.0 \\
GPT-5.4 Mini & 14.7 & 4.9 & 25.0 & 21.0 & 20.0 & 0.0 & 0.0 & 0.0 & 8.3 & 10.0 & 0.0 & 10.0 \\
GPT-4o Mini (2024-07) & 9.8 & 3.9 & 19.0 & 14.2 & 20.0 & 5.9 & 4.8 & 0.0 & 0.0 & 0.0 & 0.0 & 0.0 \\
\midrule
\multicolumn{13}{c}{\textbf{Tool-free VLMs}} \\
\cmidrule(lr){1-13}
Gemini 3.1 Pro & 34.3 & 31.4 & 73.2 & 70.1 & 50.0 & 29.4 & 38.1 & 38.5 & 25.0 & 40.0 & 11.1 & 10.0 \\
Gemini 3 Flash & 30.4 & 24.5 & 74.4 & 68.6 & 40.0 & 23.5 & 33.3 & 30.8 & 16.7 & 30.0 & 0.0 & 10.0 \\
Gemini 2.5 Pro & 28.4 & 23.5 & 73.0 & 69.6 & 30.0 & 17.6 & 33.3 & 15.4 & 25.0 & 40.0 & 11.1 & 10.0 \\
Gemini 2.5 Flash & 26.5 & 18.6 & 65.0 & 58.0 & 40.0 & 17.6 & 23.8 & 15.4 & 16.7 & 20.0 & 0.0 & 10.0 \\
Llama 4 Maverick & 15.7 & 9.8 & 46.7 & 37.8 & 40.0 & 5.9 & 14.3 & 0.0 & 8.3 & 10.0 & 0.0 & 0.0 \\
Qwen3-VL-32B & 10.8 & 2.9 & 35.7 & 29.0 & 10.0 & 0.0 & 4.8 & 0.0 & 8.3 & 0.0 & 0.0 & 0.0 \\
Qwen2.5-VL-72B & 8.8 & 4.9 & 39.2 & 33.4 & 10.0 & 0.0 & 9.5 & 7.7 & 0.0 & 0.0 & 0.0 & 10.0 \\
\midrule
\multicolumn{13}{c}{\textbf{Deep Research Systems}} \\
\cmidrule(lr){1-13}
o3 Deep Research & 32.4 & 19.6 & 56.1 & 49.8 & 20.0 & 5.9 & 28.6 & 15.4 & 33.3 & 40.0 & 0.0 & 10.0 \\
o4-mini Deep Research & 27.5 & 11.8 & 38.0 & 32.5 & 10.0 & 5.9 & 19.0 & 7.7 & 8.3 & 20.0 & 11.1 & 10.0 \\
Perplexity Sonar Deep Research & 22.5 & 21.6 & 42.5 & 37.2 & 10.0 & 23.5 & 28.6 & 30.8 & 25.0 & 40.0 & 0.0 & 0.0 \\
Tongyi DeepResearch 30B & 6.9 & 2.9 & 22.8 & 19.0 & 10.0 & 0.0 & 9.5 & 0.0 & 0.0 & 0.0 & 0.0 & 0.0 \\
\midrule
\multicolumn{13}{c}{\textbf{Framework-based agents}} \\
\cmidrule(lr){1-13}
DeerFlow (minimax-m2.7) & 16.7 & 12.7 & 37.1 & 31.3 & 30.0 & 0.0 & 28.6 & 15.4 & 8.3 & 0.0 & 0.0 & 10.0 \\
DeerFlow (qwen3-vl-235b) & 15.7 & 9.8 & 41.8 & 38.5 & 30.0 & 5.9 & 14.3 & 7.7 & 8.3 & 0.0 & 0.0 & 10.0 \\
DeerFlow (kimi-k2.5) & 3.9 & 3.9 & 5.6 & 5.3 & 20.0 & 0.0 & 4.8 & 0.0 & 8.3 & 0.0 & 0.0 & 0.0 \\
\bottomrule
\end{tabular}
}%
\end{table*}



\subsubsection{Final Answers vs. Dependency-Consistent Stated Conclusions}

\begin{figure}[t]
\centering
\includegraphics[width=0.92\linewidth]{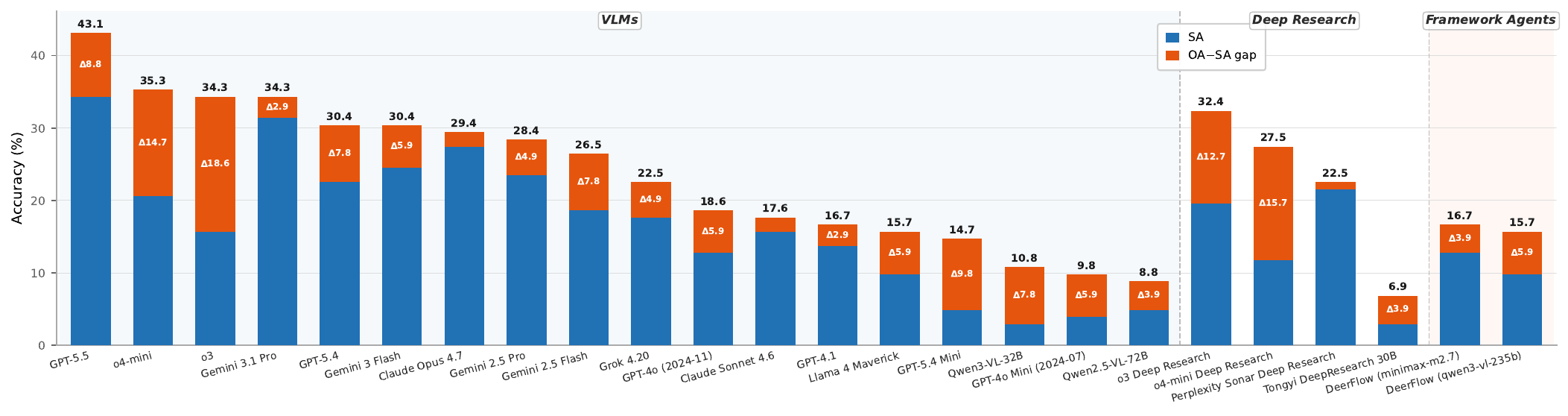}
\caption{Decomposition of final-answer accuracy for models with at least 5\% OA. Each bar separates strict answer-and-stated-conclusion correctness (SA) from the residual gap between OA and SA. The residual segment represents correct final answers for which the response does not state every required conclusion.\zh{ OA 不低于 5\% 的模型最终答案正确率分解。每个柱子将答案与已陈述结论均严格正确的部分（SA）和 OA 与 SA 之间的残差分开。残差段表示最终答案正确、但回复未陈述全部必要结论的情况。}}
\label{fig:final_answer_process_gap}
\end{figure}

Final-answer accuracy remains higher than strict answer-and-stated-conclusion correctness. Figure~\ref{fig:final_answer_process_gap} decomposes OA into SA and the residual gap between them. GPT-5.5 drops from \BREWgptfiveOA\% OA to 34.3\% SA, a gap of 8.8 points. Similarly, o4-mini drops from 35.3\% OA to 20.6\% SA, a gap of 14.7 points, and o3 Deep Research drops from 32.4\% OA to 19.6\% SA, a gap of 12.8 points. These measurements show that a correct final answer does not guarantee that the response states every necessary intermediate conclusion. Conversely, a high DACS indicates broad, dependency-consistent checklist coverage in the response, but does not imply a high OA. Neither metric alone verifies actual retrieval or source grounding.

\zh{
最终答案正确率仍高于答案与已陈述结论均严格正确的准确率。Figure~\ref{fig:final_answer_process_gap} 将 OA 分解为 SA 及两者之间的残差 gap。GPT-5.5 从 \BREWgptfiveOA\% OA 下降到 34.3\% SA，差距为 8.8 个百分点。类似地，o4-mini 从 35.3\% OA 下降到 20.6\% SA，差距为 14.7 个百分点；o3 Deep Research 从 32.4\% OA 下降到 19.6\% SA，差距为 12.8 个百分点。这些测量表明，最终答案正确并不能保证回复陈述了每个必要的中间结论。反过来，高 DACS 表明回复对 checklist 的覆盖广泛且符合依赖关系，但并不意味着 OA 较高。任何一个指标本身都不能验证是否发生了真实检索，也不能验证是否有来源支撑。
}

DACS further reveals dependency inconsistencies among the stated conclusions. CS measures how many checklist conclusions are stated correctly anywhere in a response, whereas DACS withholds credit from a conclusion if any required ancestor is missing. The gap is substantial for many systems. Llama 4 Maverick reaches 46.7\% CS but only 37.8\% DACS; Gemini 2.5 Flash reaches 65.0\% CS but 58.0\% DACS; and Grok 4.20 reaches 69.3\% CS but 62.4\% DACS. These differences identify downstream statements whose prerequisites are absent from the response, but do not establish how those statements were retrieved or grounded. Appendix~\ref{sec:appendix_split_failure_diagnostics} provides the complete split-level results and associated small-sample caveats.

\zh{
DACS 进一步揭示已陈述结论之间的依赖不一致。CS 衡量回复中有多少 checklist 结论在任意位置被正确陈述，而不考虑依赖关系；DACS 则在任一必要祖先缺失时不给相应结论计分。许多系统在两个指标之间存在明显差距。Llama 4 Maverick 达到 46.7\% CS，但 DACS 只有 37.8\%；Gemini 2.5 Flash 达到 65.0\% CS，但 DACS 为 58.0\%；Grok 4.20 达到 69.3\% CS，但 DACS 为 62.4\%。这些差异标识了回复中缺少必要前提的下游陈述，但不能说明这些陈述背后的信息是如何检索的，也不能说明陈述如何得到来源支撑。附录~\ref{sec:appendix_split_failure_diagnostics} 给出完整的 split-level 结果及相应的小样本解释。
}

\subsection{Analysis and Ablations}

\subsubsection{Image Ablation}
\label{sec:text_only}

\begin{figure}[t]
\centering
\includegraphics[width=\linewidth]{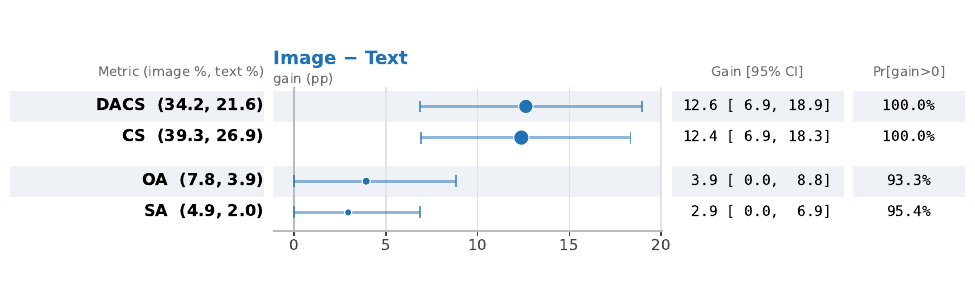}
\caption{Multimodal necessity: the same open source VLM (\texttt{Qwen3-VL-235B}) answering all 102 items \emph{with} vs \emph{without} the image, under the \ourmethodname metrics. Each row is one metric, sorted by the image$-$text gain: the marker is the gain's point estimate, the horizontal line its 95\% bootstrap CI over the 102 items ($B=10000$, seed 20260803), and marker area is proportional to the with-image score. The two right-hand columns restate the gain and the fraction of bootstrap resamples with a positive gain as exact numbers. Removing the image drops every metric; the effect is largest and direction-robust on the process metrics (CS, DACS), showing that images help the model state more of the dependency-constrained chain, not only the answer.\zh{ 多模态必要性：同一开源 VLM（\texttt{Qwen3-VL-235B}）在有图/无图两种设定下作答全部 102 题，并使用 \ourmethodname 指标评测。每行一个指标，按 image$-$text 增益排序：标记为增益点估计，横线为对 102 题做 $B=10000$ 次重采样（seed 20260803）的 95\% bootstrap 置信区间，标记面积正比于该指标的有图分数。右侧两列把增益与增益为正的重采样比例还原为精确数字。去掉图后所有指标下降，且在过程指标（CS、DACS）上幅度最大、方向稳健，说明图能帮助模型陈述更多受依赖约束的链路内容，而不仅影响最终答案。}}
\label{fig:text_only_baseline}
\end{figure}

We test the necessity of multimodal input by asking the same open VLM (Qwen3-VL-235B) to answer every item twice: once with the image and once with the image withheld. The question, reference answer, checklist, and judge remain fixed between the two conditions. Figure~\ref{fig:text_only_baseline} presents the results.

\zh{
我们让同一个开源 VLM（Qwen3-VL-235B）对每道题作答两次：一次提供图像，一次不提供图像。两种条件下的问题、reference answer、checklist 和 judge 保持不变，以检验多模态输入的必要性。结果见图~\ref{fig:text_only_baseline}。
}

The effect of the image is strongest on metrics that measure the evidence chain. CS rises from \BREWtoTextCS\% to \BREWtoImageCS\% (gain \BREWtoCSgain, 95\% CI [\BREWtoCSgainCIlo, \BREWtoCSgainCIhi]), while DACS rises from \BREWtoTextDACS\% to \BREWtoImageDACS\% (gain \BREWtoDACSgain, CI [\BREWtoDACSgainCIlo, \BREWtoDACSgainCIhi]). The CS and DACS gains are positive in \BREWtoCSgainProb\% and \BREWtoDACSgainProb\% of bootstrap resamples, respectively, and neither interval contains zero. The gain in final-answer accuracy has the same direction (\BREWtoImageOA\% versus \BREWtoTextOA\%). The fixed evaluation set includes both groups, which avoids post-hoc selection based on the ablation outcomes; Section~\ref{sec:limitations} characterizes the text-only routes. Under this probe, visual evidence is therefore necessary for correctly stating a larger portion of the dependency-constrained chain.

\zh{
图像的影响在衡量证据链的指标上最为明显。CS 从 \BREWtoTextCS\% 提升到 \BREWtoImageCS\%（增益 \BREWtoCSgain，95\% CI [\BREWtoCSgainCIlo, \BREWtoCSgainCIhi]），DACS 则从 \BREWtoTextDACS\% 提升到 \BREWtoImageDACS\%（增益 \BREWtoDACSgain，CI [\BREWtoDACSgainCIlo, \BREWtoDACSgainCIhi]）。CS 和 DACS 的增益分别在 \BREWtoCSgainProb\% 和 \BREWtoDACSgainProb\% 的 bootstrap 重采样中为正，且两个置信区间都不包含零。最终答案正确率的增益方向相同（\BREWtoImageOA\% 对 \BREWtoTextOA\%）。固定评测集同时包含两组题目，从而避免根据消融结果进行事后筛选；第~\ref{sec:limitations} 节进一步分析纯文本路径。因此，在该探测中，要正确陈述受依赖关系约束的链路中更大比例的内容，视觉证据是必要的。
}

This direction is stable across judges: under every judge in the pool, OA, CS, and DACS are higher with images, whereas SA is less stable. Appendix~\ref{sec:appendix_image_judge_robustness} provides the complete self-judge analysis.

\zh{
这一方向在不同 judge 下保持稳定：对于池中的每一个 judge，带图条件下的 OA、CS 和 DACS 都更高，而 SA 的方向较不稳定。完整的自评 judge 分析见附录~\ref{sec:appendix_image_judge_robustness}。
}

\subsubsection{Checklist-Length Pressure}
\label{sec:checklist_length_pressure}
\begin{figure}[t]
\centering
\includegraphics[width=0.98\linewidth]{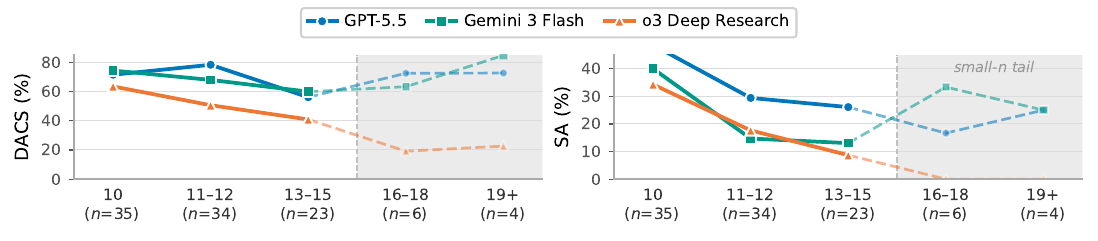}
\caption{DACS and SA for representative models across checklist-length ranges. The final two shaded ranges contain only six and four items, respectively, and are excluded from the main interpretation.\zh{ 代表性模型在不同 checklist 长度区间上的 DACS 与 SA。阴影标出的最后两个区间分别仅包含 6 个和 4 个样本，因此不纳入正文的主要分析。}}
\label{fig:checklist_length_pressure}
\end{figure}

We use checklist length as a proxy for evidential density. A longer checklist requires an answer to support more annotated conclusions, but does not directly measure browsing cost or the number of actions taken. We partition the 102 items into five ranges containing 10, 11--12, 13--15, 16--18, and 19+ statements. These ranges contain 35, 34, 23, 6, and 4 items, respectively. Figure~\ref{fig:checklist_length_pressure} reports DACS and SA for three representative models across the five ranges.

\zh{
我们将 checklist 长度作为证据密度的代理指标。更长的 checklist 要求答案支撑更多已标注结论，但并不直接衡量浏览成本或操作次数。我们将 102 个样本划分为五个区间，分别包含 10、11--12、13--15、16--18 和 19+ 个陈述。各区间分别包含 35、34、23、6 和 4 个样本。图~\ref{fig:checklist_length_pressure} 展示了三个代表性模型在这五个区间上的 DACS 和 SA。
}

Across the three ranges with more observations ($n\geq20$), SA decreases monotonically with checklist length for every model. DACS is less uniform: GPT-5.5 rises from the 10-step range to the 11--12 range before falling, whereas Gemini 3 Flash and o3 Deep Research decline across the first two ranges. All three models, however, reach their lowest DACS in the 13--15 range. The most consistent pattern therefore appears in strict satisfaction: as the number of required conclusions increases, models become progressively less likely to satisfy the entire checklist, even when they retain partial, dependency-consistent coverage.

\zh{
在样本量较多的三个区间（$n\geq20$）中，所有模型的 SA 均随 checklist 长度单调下降。DACS 则不那么整齐：GPT-5.5 从 10 步区间上升到 11--12 区间后再下降，而 Gemini 3 Flash 和 o3 Deep Research 在前两个区间上持续下降。不过，三个模型的 DACS 都在 13--15 区间达到最低值。因此，最一致的规律体现在严格满足率上：随着必要结论数量增加，模型完整满足整个 checklist 的可能性逐步降低，即使它们仍能保留部分具有依赖一致性的覆盖。
}

\subsubsection{High DACS in Tool-Free Gemini Models}
\label{sec:tool_free_dacs}

The Gemini models stand out within the tool-free group. Gemini 3.1 Pro records \BREWtfGemTopDACS\% DACS, and every tested Gemini model outperforms all non-Gemini tool-free models, whose scores range from \BREWtfCtrlWorstDACS\% to \BREWtfCtrlBestDACS\%. All models in this group are evaluated under the same input and tool conditions: they receive the question and its associated image but cannot use open-web search. The high Gemini scores therefore cannot be attributed to the tool-free setting alone. Moreover, tool-free does not mean evidence-free. The image ablation in Section~\ref{sec:text_only} shows that visual evidence improves dependency-consistent coverage, suggesting that strong multimodal understanding contributes to Gemini's results.

\zh{
Gemini 模型在 tool-free 组中表现突出。Gemini 3.1 Pro 的 DACS 为 \BREWtfGemTopDACS\%，所有被测 Gemini 模型都高于全部非 Gemini tool-free 模型，后者的得分范围为 \BREWtfCtrlWorstDACS\%--\BREWtfCtrlBestDACS\%。该组模型采用相同的输入和工具条件：它们会收到问题及其关联图像，但不能使用开放网页搜索。因此，Gemini 的高分不能仅归因于 tool-free 这一设置。此外，tool-free 并不意味着没有证据输入。第~\ref{sec:text_only} 节的图像消融表明，视觉证据能够提高依赖一致的结论覆盖，这说明较强的多模态理解能力有助于 Gemini 取得这些结果。
}

The score distribution provides a more direct explanation. First, Gemini 3.1 Pro has both a high CS and a DACS of \BREWtfGemTopDACS\%, with only a small gap between the two metrics. Many of the checklist conclusions that it states correctly therefore also satisfy their prerequisite dependencies. Second, its DACS is zero on only \BREWtfGemZero\% of items, while it achieves complete dependency-consistent coverage on \BREWtfGemFull\%. DACS awards partial credit for correctly recovered portions of a dependency graph and averages this credit across items. A model can therefore attain a high DACS by covering substantial, dependency-consistent portions of many items, even if it does not complete every item. Gemini's high score reflects this combination of broad coverage and consistent intermediate conclusions.

\zh{
得分分布给出了更直接的解释。首先，Gemini 3.1 Pro 的 CS 较高，DACS 也达到 \BREWtfGemTopDACS\%，两个指标之间的差距较小。因此，它正确陈述的许多 checklist 结论也满足相应的前置依赖。其次，它仅在 \BREWtfGemZero\% 的题目上 DACS 为零，而在 \BREWtfGemFull\% 的题目上实现了完整的依赖一致覆盖。DACS 会对依赖图中正确恢复的部分给予部分得分，再对所有题目的得分取平均。因此，即使没有完整解决每道题，只要在许多题目上覆盖了较多且依赖一致的结论，模型仍可获得较高的 DACS。Gemini 的高分反映了广泛覆盖与中间结论一致性之间的这种结合。
}

However, Gemini 3.1 Pro still has low OA and SA, and its rate of complete dependency-consistent coverage is only \BREWtfGemFull\%. It often recovers many intermediate conclusions but misses one or more links required to complete the evidence chain and support the final answer. The metrics capture this distinction: DACS measures dependency-consistent progress within the chain, whereas OA and SA measure end-to-end task completion. This pattern is also stable across judges. Under each of the \BREWtfJudgeN{} independent judges, at least one Gemini model ranks among the top \BREWtfJudgeWorstRank{} models by DACS.

\zh{
不过，Gemini 3.1 Pro 的 OA 和 SA 仍然较低，完整依赖一致覆盖率也仅为 \BREWtfGemFull\%。它通常能够恢复较多中间结论，但仍会遗漏完成证据链并支撑最终答案所需的一个或多个环节。不同指标反映了这一区别：DACS 衡量证据链中依赖一致的过程进展，而 OA 和 SA 衡量端到端任务完成情况。该模式在不同 judge 下也较为稳定。对于全部 \BREWtfJudgeN{} 个独立 judge，都至少有一个 Gemini 模型的 DACS 排名前 \BREWtfJudgeWorstRank{}。
}

\subsubsection{Failure-Mode Analysis}
\label{sec:failure_mode_analysis}
To characterize unsupported conclusions beyond the aggregate process scores, we assign each failed checklist statement a diagnostic label from the MM-BrowseComp taxonomy~\citep{li2025mmbrowsecompcomprehensivebenchmarkmultimodal}. For all 25 systems, the failure set is defined consistently using frozen verdicts from Qwen3.5-Flash, which differs from the current judge used for the main metrics. Qwen3.6 Plus then assigns the diagnostic labels. The analysis covers \BREWfailSteps{} checklist-step decisions from \BREWfailRuns{} model--item runs, of which \BREWfailFailedSteps{} receive failure labels. Appendix~\ref{sec:appendix_failure_mode_analysis} provides the taxonomy, coverage details, and complete counts.

\zh{
为了在聚合过程分数之外刻画未得到支撑的结论，我们按照 MM-BrowseComp taxonomy~\citep{li2025mmbrowsecompcomprehensivebenchmarkmultimodal}，为每个失败的 checklist statement 分配诊断标签。对于全部 25 个系统，failure set 均按同一口径，使用 Qwen3.5-Flash 的冻结 verdict 来定义；Qwen3.5-Flash 不同于主指标当前使用的 judge。随后，由 Qwen3.6 Plus 分配诊断标签。该分析覆盖来自 \BREWfailRuns{} 个 model--item run 的 \BREWfailSteps{} 个 checklist-step decision，其中 \BREWfailFailedSteps{} 个被分配 failure label。附录~\ref{sec:appendix_failure_mode_analysis} 给出 taxonomy、覆盖范围细节和完整计数。
}

\begin{figure}[t]
\centering
\includegraphics[width=\linewidth]{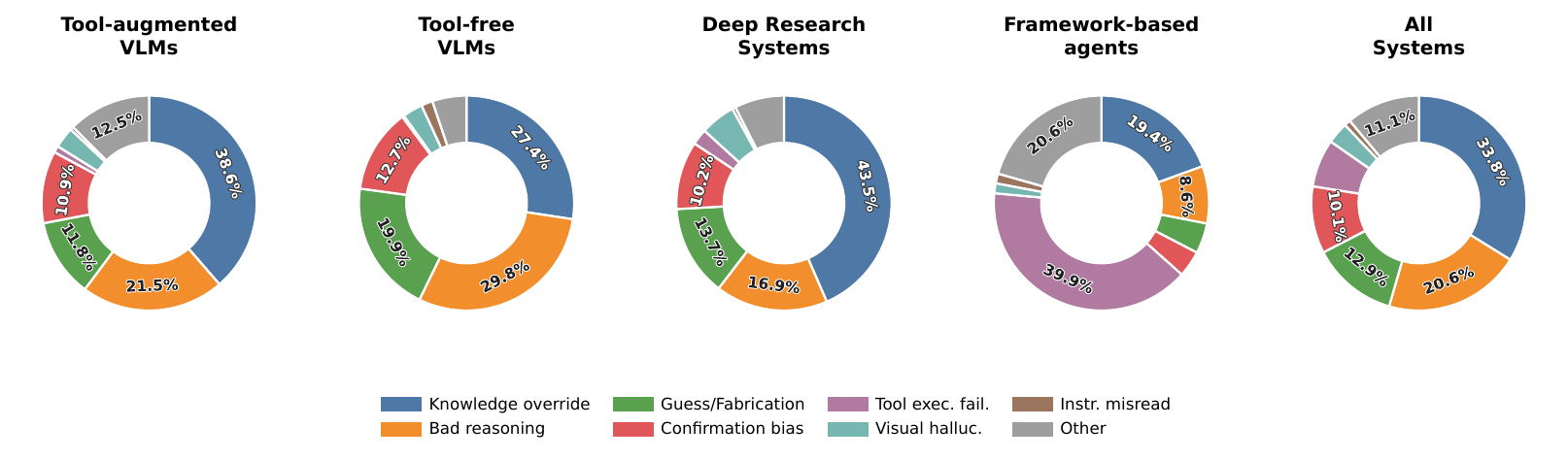}
\caption{Distribution of classifier-assigned failure labels among failed checklist items, aggregated by model group. Passed checklist items are excluded, and slices below 8\% are left unlabeled for readability.\zh{ 分类器所分配的失败标签在各模型组 failed checklist item 中的分布。通过的 checklist item 不计入；为便于阅读，低于 8\% 的小扇区不标注百分比。}}
\label{fig:failure_mode_analysis}
\end{figure}

\paragraph{Failure-label distribution.}
Figure~\ref{fig:failure_mode_analysis} reveals two broad patterns. First, failed steps are concentrated in cases where models override task evidence with prior knowledge or draw conclusions that their reasoning does not adequately support. This pattern indicates persistent weaknesses in evidence integration and reasoning fidelity. Second, failure profiles vary with system design. Knowledge override dominates the failures of tool-augmented VLMs and Deep Research Systems, whereas framework-based agents exhibit a broader mixture that also includes tool-execution and uncategorized failures. This contrast suggests that retrieval and tool access alone do not ensure evidence-grounded conclusions, while framework-based pipelines face a wider range of operational failure surfaces. Because these distributions are conditioned on failed checklist items and the labels are textual diagnostics, they characterize the composition of observed failures, not group failure rates or the causal effects of tool access, retrieval, or orchestration.

\zh{
\textbf{失败标签分布。}
图~\ref{fig:failure_mode_analysis} 呈现出两个总体规律。首先，failed step 主要集中在两类情形：模型以既有知识覆盖任务证据，或模型的推理不足以支撑所得结论。这一规律表明，证据整合与推理忠实度仍存在普遍不足。其次，失败构成随系统设计而变化。Tool-augmented VLM 和 Deep Research Systems 的失败以 knowledge override 为主，而 framework-based agents 的失败类型更为分散，还包括工具执行和未归类问题。这一差异表明，仅有检索和工具访问并不能保证结论以证据为基础，而 framework-based pipeline 面临的操作性失败环节更为多样。由于这些分布以 failed checklist item 为条件，且标签仅是文本诊断类别，因此它们描述的是已观察失败的构成，而不是各组的失败率，也不能据此推断工具访问、检索或编排的因果影响。
}

\paragraph{Failure rate by dependency depth.}
We locate unsupported conclusions within the annotated dependency structure rather than by their order in the serialized checklist. A root node has a dependency depth of zero; for every other node, the depth is one plus the maximum depth of its direct prerequisites. We normalize depth by the number of levels in each item and partition the nodes into the first, middle, and final thirds. Across all checklist-step decisions, the failure rate rises from \BREWfailDagEarlyRate\% in the first third to \BREWfailDagMiddleRate\% in the middle third and \BREWfailDagLateRate\% in the final third. The same monotonic pattern holds within each of the four system groups. Unsupported conclusions are therefore increasingly prevalent among nodes that lie deeper in the prerequisite structure. This result is a structural diagnostic, not a temporal trace of model behavior: dependency depth does not reveal when a conclusion was attempted, and the observed gradient does not establish that upstream failures caused downstream failures.

\zh{
\textbf{按依赖深度划分的失败率。}
我们按照已标注的依赖结构，而非节点在序列化 checklist 中的顺序，定位未支撑结论。根节点的依赖深度定义为零；对于其他节点，其深度定义为直接前置节点的最大深度加一。随后，我们按每道题的深度层数对深度进行归一化，并将节点划分为前、中、后三段。在全部 checklist-step decision 上，失败率从前段的 \BREWfailDagEarlyRate\% 上升至中段的 \BREWfailDagMiddleRate\%，并进一步上升至后段的 \BREWfailDagLateRate\%。这一单调趋势在四类系统内部均成立。因此，未支撑结论在依赖结构较深的节点中更为常见。该结果是结构性诊断，而非模型行为的时间轨迹：依赖深度不能揭示模型何时尝试某个结论，观察到的梯度也不能证明上游失败导致了下游失败。
}

\section{Limitations and Broader Impact}
\label{sec:limitations}

\paragraph{Limitations.}
Our agent-framework group is instantiated using a single open-source framework; its results therefore characterize this particular harness---including its planning loop, tool interface, and error handling---rather than agent frameworks in general. Framework runs may also fail for reasons unrelated to reasoning, such as tool-execution errors, whereas the metrics score submitted responses without distinguishing harness failures from reasoning failures. The evidence thus supports only the configuration-specific finding that this framework did not translate its additional machinery into higher process scores. Broader conclusions require instrumented tool-execution rates and a controlled comparison of the same underlying model with and without the framework.

\zh{
我们的 agent framework 组仅使用一个开源框架实例；因此，结果反映的是这一特定 harness（包括其规划循环、工具接口和错误处理），而非一般意义上的 agent framework。框架运行也可能因工具执行错误等与推理无关的原因而失败，但现有指标只对提交的回复评分，无法区分 harness 失败与推理失败。因此，现有证据仅支持针对该配置的结论：这一框架未能将其额外机制转化为更高的过程分数。要得出更广泛的结论，还需要记录工具执行率，并在同一底层模型上对使用和不使用该框架的情形进行受控比较。
}

\paragraph{Broader impact.}
\ourmethodname{} rewards responses that state necessary intermediate conclusions rather than providing only plausible final answers. This encourages more auditable research assistants and reveals failures obscured by final-answer accuracy, although the resulting process scores do not by themselves verify retrieval provenance. At the same time, linking evidence across sources and modalities is a dual-use capability that could be directed at private individuals. We therefore construct items around public works, organizations, and documented public figures, and do not release private personal data. Publishing questions that rely on the live web also creates risks of benchmark contamination and automated load on third-party sites; we withhold hidden reasoning paths and full source chains, encourage the reuse of cached responses, and recommend respecting robots directives and rate limits.

\zh{
\ourmethodname{} 奖励陈述必要中间结论的回复，而非仅提供看似合理的最终答案。这有助于提高研究助手的可审计性，并揭示被最终答案准确率所掩盖的失败，但由此得到的过程分数本身并不能验证检索来源。与此同时，跨来源和跨模态关联证据是一种可能被用于调查私人个体的双重用途能力。因此，我们仅围绕公开作品、机构和有公开记录的公众人物构建题目，并且不发布非公开个人数据。公开依赖实时网页的问题也会带来 benchmark 污染风险，并增加第三方网站的自动化访问负担；为此，我们不公开推理路径和完整来源链，鼓励复用缓存回复，并建议遵守 robots 指令和速率限制。
}

\section{Conclusion}

We introduced \ourmethodname, a long-horizon multimodal deep research benchmark that evaluates submitted responses to real-world research tasks requiring sustained evidence chains. Its 102 items are constructed from hidden Node-Relation graphs, grounded in multi-source and non-text evidence, and annotated with irreducible checklists of necessary intermediate conclusions. This design evaluates not only short-answer correctness but also whether a submitted response correctly states the necessary conclusions under their annotated dependencies.

\zh{
我们提出了 \ourmethodname，一个 long-horizon 多模态深度研究基准，用于评估智能体针对需要持续证据链的真实世界研究任务所提交的回复。其 102 个条目基于隐藏的 Node-Relation graph 构建，以多来源和非文本证据为基础，并标注了由必要中间结论组成的不可约 checklist。这一设计不仅评估短答案的正确性，还评估提交的回复是否在已标注依赖约束下正确陈述了必要结论。
}

Our evaluation suggests that responses submitted by current systems struggle to preserve dependency-consistent conclusions. \BREWtopModel{} attains the highest OA at \BREWtopOA\%, but only \BREWtopSA\% SA. The dedicated deep research system with the highest OA, \BREWtopDRModel, reaches \BREWtopDROA\% OA and \BREWtopDRSA\% SA. With 102 items, the wide confidence intervals around these point estimates preclude a complete ranking of systems. Across most systems, OA exceeds SA and CS exceeds DACS. This pattern indicates that a response can provide a correct final answer or isolated downstream facts without correctly stating all annotated prerequisites. Breakdowns by category, structure, checklist length, operation tag, and source density, together with the failure taxonomy, provide exploratory diagnostics rather than robust group-level conclusions. Taken together, these response-level patterns suggest difficulty keeping entities, constraints, and multimodal bridge conclusions aligned across a long dependency chain.

\zh{
我们的评测表明，当前系统提交的回复难以保持满足依赖约束的结论。\BREWtopModel{} 取得了最高的 OA（\BREWtopOA\%），但 SA 仅为 \BREWtopSA\%。在专用 deep research 系统中，OA 最高的 \BREWtopDRModel{} 达到 \BREWtopDROA\% OA 和 \BREWtopDRSA\% SA。在 102 道题的规模下，这些点估计的置信区间较宽，不足以支持对系统进行完整排名。对大多数系统而言，OA 高于 SA 且 CS 高于 DACS。这一模式表明，回复可以给出正确的最终答案或孤立的下游事实，却未能正确陈述全部已标注前置依赖。按 category、structure、checklist 长度、operation tag 和 source density 的拆分，以及 failure taxonomy，均属于探索性诊断，不能视为稳健的分组结论。综合来看，这些回复层面的模式表明，系统难以在长依赖链中持续保持实体、约束和多模态桥接结论彼此一致。
}

These findings suggest that deep research agents should be evaluated on both final-answer correctness and response-level dependency consistency. DACS measures whether conclusions are correctly stated under annotated dependencies; it does not prove that those conclusions came from retrieved evidence. Future evaluation should therefore incorporate citation- or trace-grounded verification. Future systems likewise need stronger research-state tracking, source and modality binding, branch-specific prerequisite preservation, and chain-level verification.

\zh{
这些发现表明，应当从最终答案正确性和回复层面的依赖一致性两个方面评估深度研究智能体。DACS 衡量结论是否在已标注依赖约束下得到正确陈述；它并不能证明这些结论确实来自检索到的证据。因此，未来的评测应当纳入基于引用或轨迹的验证。未来的系统同样需要更强的研究状态追踪、来源与模态绑定、各分支前置依赖保持以及整链级别验证能力。
}

\bibliography{main}

@inproceedings{yao2023react,
  title = {ReAct: Synergizing Reasoning and Acting in Language Models},
  author = {Yao, Shunyu and Zhao, Jeffrey and Yu, Dian and Du, Nan and Shafran, Izhak and Narasimhan, Karthik and Cao, Yuan},
  booktitle = {International Conference on Learning Representations},
  year = {2023},
  url = {https://openreview.net/forum?id=WE_vluYUL-X}
}

@inproceedings{deng2023mind2web,
  title = {Mind2Web: Towards a Generalist Agent for the Web},
  author = {Deng, Xiang and Gu, Yu and Zheng, Boyuan and Chen, Shijie and Stevens, Sam and Wang, Boshi and Sun, Huan and Su, Yu},
  booktitle = {Advances in Neural Information Processing Systems},
  year = {2023},
  url = {https://arxiv.org/abs/2306.06070}
}

@inproceedings{zhou2024webarena,
  title = {WebArena: A Realistic Web Environment for Building Autonomous Agents},
  author = {Zhou, Shuyan and Xu, Frank F. and Zhu, Hao and Zhou, Xuhui and Lo, Robert and Sridhar, Abishek and Cheng, Xianyi and Ou, Tianyue and Bisk, Yonatan and Fried, Daniel and Alon, Uri and Neubig, Graham},
  booktitle = {International Conference on Learning Representations},
  year = {2024},
  url = {https://openreview.net/forum?id=oKn9c6ytLx}
}

@inproceedings{he2024webvoyager,
  title = {WebVoyager: Building an End-to-End Web Agent with Large Multimodal Models},
  author = {He, Hongliang and Yao, Wenlin and Ma, Kaixin and Yu, Wenhao and Dai, Yong and Zhang, Hongming and Lan, Zhenzhong and Yu, Dong},
  booktitle = {Annual Meeting of the Association for Computational Linguistics},
  year = {2024},
  url = {https://aclanthology.org/2024.acl-long.371/}
}

@inproceedings{xie2024osworld,
  title = {OSWorld: Benchmarking Multimodal Agents for Open-Ended Tasks in Real Computer Environments},
  author = {Xie, Tianbao and Zhang, Danyang and Chen, Jixuan and Li, Xiaochuan and Zhao, Siheng and Cao, Ruisheng and Hua, Toh Jing and Cheng, Zhoujun and Shin, Dobin and Lei, Fangyu and Liu, Yitao and Xu, Yiheng and Zhou, Shuyan and Savarese, Silvio and Xiong, Caiming and Zhou, Victor Zhong and Wu, Tao and Hu, C. Lawrence},
  booktitle = {Advances in Neural Information Processing Systems},
  year = {2024},
  url = {https://arxiv.org/abs/2404.07972}
}

@inproceedings{mialon2024gaia,
  title = {GAIA: A Benchmark for General AI Assistants},
  author = {Mialon, Gr{\'e}goire and Fourrier, Cl{\'e}mentine and Wolf, Thomas and LeCun, Yann and Scialom, Thomas},
  booktitle = {International Conference on Learning Representations},
  year = {2024},
  url = {https://openreview.net/forum?id=fibxvahvs3}
}

@misc{wei2025browsecomp,
  title = {BrowseComp: A Simple Yet Challenging Benchmark for Browsing Agents},
  author = {Wei, Jason and Sun, Zhiqing and Papay, Spencer and McKinney, Scott and Han, Jeffrey and Fulford, Isa and Chung, Hyung Won and Passos, Alex Tachard and Fedus, William and Glaese, Amelia},
  year = {2025},
  eprint = {2504.12516},
  archivePrefix = {arXiv},
  primaryClass = {cs.CL},
  url = {https://arxiv.org/abs/2504.12516}
}

@inproceedings{liu2024mmbench,
  title = {MMBench: Is Your Multi-modal Model an All-Around Player?},
  author = {Liu, Yuan and Duan, Haodong and Zhang, Yuanhan and Li, Bo and Zhang, Songyang and Zhao, Wangbo and Yuan, Yike and Wang, Jiaqi and He, Conghui and Liu, Ziwei and Chen, Kai and Lin, Dahua},
  booktitle = {European Conference on Computer Vision},
  year = {2024},
  url = {https://arxiv.org/abs/2307.06281}
}

@misc{Jin2025SearchR1TL,
  title = {Search-R1: Training LLMs to Reason and Leverage Search Engines with Reinforcement Learning},
  author = {Jin, Bowen and Zeng, Hansi and Yue, Zhenrui and Yoon, Jinsung and Arik, Sercan and Wang, Dong and Zamani, Hamed and Han, Jiawei},
  year = {2025},
  eprint = {2503.09516},
  archivePrefix = {arXiv},
  primaryClass = {cs.CL},
  url = {https://arxiv.org/abs/2503.09516}
}

@misc{oaidr,
  title = {Introducing Deep Research},
  author = {{OpenAI}},
  year = {2025},
  url = {https://openai.com/index/introducing-deep-research/}
}

@misc{geminidr,
  title = {Gemini Deep Research},
  author = {{Google}},
  year = {2024},
  url = {https://gemini.google/overview/deep-research/}
}

@misc{perplexitydr,
  title = {Introducing Perplexity Deep Research},
  author = {{Perplexity AI}},
  year = {2025},
  url = {https://www.perplexity.ai/hub/blog/introducing-perplexity-deep-research}
}

@misc{grokdeepsearch,
  title = {Grok 3 Beta: The Age of Reasoning Agents},
  author = {{xAI}},
  year = {2025},
  url = {https://x.ai/news/grok-3}
}

@misc{msftdr,
  title = {Microsoft Copilot},
  author = {{Microsoft}},
  year = {2024},
  url = {https://copilot.microsoft.com/}
}

@misc{bytedance2026deerflow,
  title = {DeerFlow},
  author = {{ByteDance}},
  year = {2026},
  url = {https://github.com/bytedance/deer-flow}
}

@inproceedings{yao2022webshop,
  title = {WebShop: Towards Scalable Real-World Web Interaction with Grounded Language Agents},
  author = {Yao, Shunyu and Chen, Howard and Yang, John and Narasimhan, Karthik},
  booktitle = {Advances in Neural Information Processing Systems},
  year = {2022},
  url = {https://arxiv.org/abs/2207.01206}
}

@misc{Jiang2024MMSearchBT,
  title = {MMSearch: Benchmarking the Potential of Large Models as Multi-modal Search Engines},
  author = {Jiang, Dongzhi and Zhang, Renrui and Guo, Ziyu and Wu, Yanmin and Lei, Jiayi and Qiu, Pengshuo and Lu, Pan and Chen, Zehui and Fu, Chaoyou and Song, Guanglu and Gao, Peng and Liu, Yu and Li, Chunyuan and Li, Hongsheng},
  year = {2024},
  eprint = {2409.12959},
  archivePrefix = {arXiv},
  primaryClass = {cs.CV},
  url = {https://arxiv.org/abs/2409.12959}
}

@misc{Zhou2025BrowseCompZHBW,
  title = {BrowseComp-ZH: Benchmarking Web Browsing Ability of Large Language Models in Chinese},
  author = {Zhou, Peilin and Leon, Bruce and Ying, Xiang and Zhang, Can and Shao, Yifan and Ye, Qichen and Chong, Dading and Jin, Zhiling and Xie, Chenxuan and Cao, Meng and others},
  year = {2025},
  eprint = {2504.19314},
  archivePrefix = {arXiv},
  primaryClass = {cs.CL},
  url = {https://arxiv.org/abs/2504.19314}
}

@misc{Du2025DeepResearchBA,
  title = {DeepResearch Bench: A Comprehensive Benchmark for Deep Research Agents},
  author = {Du, Mingxuan and Xu, Benfeng and Zhu, Chiwei and Wang, Xiaorui and Mao, Zhendong},
  year = {2025},
  eprint = {2506.11763},
  archivePrefix = {arXiv},
  primaryClass = {cs.CL},
  url = {https://arxiv.org/abs/2506.11763}
}

@misc{li2025mmbrowsecompcomprehensivebenchmarkmultimodal,
      title={MM-BrowseComp: A Comprehensive Benchmark for Multimodal Browsing Agents}, 
      author={Shilong Li and Xingyuan Bu and Wenjie Wang and Jiaheng Liu and Jun Dong and Haoyang He and Hao Lu and Haozhe Zhang and Chenchen Jing and Zhen Li and Chuanhao Li and Jiayi Tian and Chenchen Zhang and Tianhao Peng and Yancheng He and Jihao Gu and Yuanxing Zhang and Jian Yang and Ge Zhang and Wenhao Huang and Wangchunshu Zhou and Zhaoxiang Zhang and Ruizhe Ding and Shilei Wen},
      year={2025},
      eprint={2508.13186},
      archivePrefix={arXiv},
      primaryClass={cs.CL},
      url={https://arxiv.org/abs/2508.13186}, 
}

@misc{lu2025deepdiveadvancingdeepsearch,
      title={DeepDive: Advancing Deep Search Agents with Knowledge Graphs and Multi-Turn RL},
      author={Rui Lu and Zhenyu Hou and Zihan Wang and Hanchen Zhang and Xiao Liu and Yujiang Li and Shi Feng and Jie Tang and Yuxiao Dong},
      year={2025},
      eprint={2509.10446},
      archivePrefix={arXiv},
      primaryClass={cs.CL},
      url={https://arxiv.org/abs/2509.10446},
}

@article{
  li2025llavaonevision,
  title={{LL}a{VA}-OneVision: Easy Visual Task Transfer},
  author={Bo Li and Yuanhan Zhang and Dong Guo and Renrui Zhang and Feng Li and Hao Zhang and Kaichen Zhang and Peiyuan Zhang and Yanwei Li and Ziwei Liu and Chunyuan Li},
  journal={Transactions on Machine Learning Research},
  issn={2835-8856},
  year={2025},
  url={https://openreview.net/forum?id=zKv8qULV6n},
  note={}
}

@article{guo2025seed1,
  title={Seed1. 5-vl technical report},
  author={Guo, Dong and Wu, Faming and Zhu, Feida and Leng, Fuxing and Shi, Guang and Chen, Haobin and Fan, Haoqi and Wang, Jian and Jiang, Jianyu and Wang, Jiawei and others},
  journal={arXiv preprint arXiv:2505.07062},
  year={2025}
}

@misc{vteam2025glm45vglm41vthinkingversatilemultimodal,
      title={GLM-4.5V and GLM-4.1V-Thinking: Towards Versatile Multimodal Reasoning with Scalable Reinforcement Learning},
      author={V Team and Wenyi Hong and Wenmeng Yu and Xiaotao Gu and Guo Wang and Guobing Gan and Haomiao Tang and Jiale Cheng and Ji Qi and Junhui Ji and Lihang Pan and Shuaiqi Duan and Weihan Wang and Yan Wang and Yean Cheng and Zehai He and Zhe Su and Zhen Yang and Ziyang Pan and Aohan Zeng and Baoxu Wang and Bin Chen and Boyan Shi and Changyu Pang and Chenhui Zhang and Da Yin and Fan Yang and Guoqing Chen and Jiazheng Xu and Jiale Zhu and Jiali Chen and Jing Chen and Jinhao Chen and Jinghao Lin and Jinjiang Wang and Junjie Chen and Leqi Lei and Letian Gong and Leyi Pan and Mingdao Liu and Mingde Xu and Mingzhi Zhang and Qinkai Zheng and Sheng Yang and Shi Zhong and Shiyu Huang and Shuyuan Zhao and Siyan Xue and Shangqin Tu and Shengbiao Meng and Tianshu Zhang and Tianwei Luo and Tianxiang Hao and Tianyu Tong and Wenkai Li and Wei Jia and Xiao Liu and Xiaohan Zhang and Xin Lyu and Xinyue Fan and Xuancheng Huang and Yanling Wang and Yadong Xue and Yanfeng Wang and Yanzi Wang and Yifan An and Yifan Du and Yiming Shi and Yiheng Huang and Yilin Niu and Yuan Wang and Yuanchang Yue and Yuchen Li and Yutao Zhang and Yuting Wang and Yu Wang and Yuxuan Zhang and Zhao Xue and Zhenyu Hou and Zhengxiao Du and Zihan Wang and Peng Zhang and Debing Liu and Bin Xu and Juanzi Li and Minlie Huang and Yuxiao Dong and Jie Tang},
      year={2025},
      eprint={2507.01006},
      archivePrefix={arXiv},
      primaryClass={cs.CV},
      url={https://arxiv.org/abs/2507.01006}, 
}

@article{Qwen3-VL,
      title={Qwen3-VL Technical Report}, 
      author={Shuai Bai and Yuxuan Cai and Ruizhe Chen and Keqin Chen and Xionghui Chen and Zesen Cheng and Lianghao Deng and Wei Ding and Chang Gao and Chunjiang Ge and Wenbin Ge and Zhifang Guo and Qidong Huang and Jie Huang and Fei Huang and Binyuan Hui and Shutong Jiang and Zhaohai Li and Mingsheng Li and Mei Li and Kaixin Li and Zicheng Lin and Junyang Lin and Xuejing Liu and Jiawei Liu and Chenglong Liu and Yang Liu and Dayiheng Liu and Shixuan Liu and Dunjie Lu and Ruilin Luo and Chenxu Lv and Rui Men and Lingchen Meng and Xuancheng Ren and Xingzhang Ren and Sibo Song and Yuchong Sun and Jun Tang and Jianhong Tu and Jianqiang Wan and Peng Wang and Pengfei Wang and Qiuyue Wang and Yuxuan Wang and Tianbao Xie and Yiheng Xu and Haiyang Xu and Jin Xu and Zhibo Yang and Mingkun Yang and Jianxin Yang and An Yang and Bowen Yu and Fei Zhang and Hang Zhang and Xi Zhang and Bo Zheng and Humen Zhong and Jingren Zhou and Fan Zhou and Jing Zhou and Yuanzhi Zhu and Ke Zhu},
	  journal={arXiv preprint arXiv:2511.21631},
      year={2025}
}

@inproceedings{duan2024vlmevalkit,
  title={Vlmevalkit: An open-source toolkit for evaluating large multi-modality models},
  author={Duan, Haodong and Yang, Junming and Qiao, Yuxuan and Fang, Xinyu and Chen, Lin and Liu, Yuan and Dong, Xiaoyi and Zang, Yuhang and Zhang, Pan and Wang, Jiaqi and others},
  booktitle={Proceedings of the 32nd ACM International Conference on Multimedia},
  pages={11198--11201},
  year={2024}
}

@article{Yue2023MMMUAM,
  title={MMMU: A Massive Multi-Discipline Multimodal Understanding and Reasoning Benchmark for Expert AGI},
  author={Xiang Yue and Yuansheng Ni and Kai Zhang and Tianyu Zheng and Ruoqi Liu and Ge Zhang and Samuel Stevens and Dongfu Jiang and Weiming Ren and Yuxuan Sun and Cong Wei and Botao Yu and Ruibin Yuan and Renliang Sun and Ming Yin and Boyuan Zheng and Zhenzhu Yang and Yibo Liu and Wenhao Huang and Huan Sun and Yu Su and Wenhu Chen},
  journal={2024 IEEE/CVF Conference on Computer Vision and Pattern Recognition (CVPR)},
  year={2023},
  pages={9556-9567},
  url={https://api.semanticscholar.org/CorpusID:265466525}
}

@inproceedings{masry-etal-2022-chartqa,
    title = "{C}hart{QA}: A Benchmark for Question Answering about Charts with Visual and Logical Reasoning",
    author = "Masry, Ahmed  and
      Long, Do Xuan  and
      Tan, Jia Qing  and
      Joty, Shafiq  and
      Hoque, Enamul",
    editor = "Muresan, Smaranda  and
      Nakov, Preslav  and
      Villavicencio, Aline",
    booktitle = "Findings of the Association for Computational Linguistics: ACL 2022",
    month = may,
    year = "2022",
    address = "Dublin, Ireland",
    publisher = "Association for Computational Linguistics",
    url = "https://aclanthology.org/2022.findings-acl.177/",
    doi = "10.18653/v1/2022.findings-acl.177",
    pages = "2263--2279"
}

@article{Mathew2020DocVQAAD,
  title={DocVQA: A Dataset for VQA on Document Images},
  author={Minesh Mathew and Dimosthenis Karatzas and R. Manmatha and C. V. Jawahar},
  journal={2021 IEEE Winter Conference on Applications of Computer Vision (WACV)},
  year={2020},
  pages={2199-2208},
  url={https://api.semanticscholar.org/CorpusID:220280200}
}

@article{Li2025RAVENEAAB,
  title={RAVENEA: A Benchmark for Multimodal Retrieval-Augmented Visual Culture Understanding},
  author={Jiaang Li and Yifei Yuan and Wenyan Li and Mohammad Aliannejadi and Daniel Hershcovich and Anders S{\o}gaard and Ivan Vuli'c and Wenxuan Zhang and Paul Pu Liang and Yang Deng and Serge J. Belongie},
  journal={ArXiv},
  year={2025},
  volume={abs/2505.14462},
  url={https://api.semanticscholar.org/CorpusID:278769446}
}

@article{Dong2025BenchmarkingRM,
  title={Benchmarking Retrieval-Augmented Multimomal Generation for Document Question Answering},
  author={Kuicai Dong and Yujing Chang and Shijie Huang and Yasheng Wang and Ruiming Tang and Yong Liu},
  journal={ArXiv},
  year={2025},
  volume={abs/2505.16470},
  url={https://api.semanticscholar.org/CorpusID:278788810}
}

@article{Jia2025BenchmarkingMK,
  title={Benchmarking Multimodal Knowledge Conflict for Large Multimodal Models},
  author={Yifan Jia and Kailin Jiang and Yuyang Liang and Qihan Ren and Yi Xin and Rui Yang and Fenze Feng and Mingcai Chen and Hengyang Lu and Haozhe Wang and Xiaoye Qu and Dongrui Liu and Lizhen Cui and Yuntao Du},
  journal={ArXiv},
  year={2025},
  volume={abs/2505.19509},
  url={https://api.semanticscholar.org/CorpusID:278904487}
}

@article{Nakano2021WebGPTBQ,
  title={WebGPT: Browser-assisted question-answering with human feedback},
  author={Reiichiro Nakano and Jacob Hilton and Suchir Balaji and Jeff Wu and Ouyang Long and Christina Kim and Christopher Hesse and Shantanu Jain and Vineet Kosaraju and William Saunders and Xu Jiang and Karl Cobbe and Tyna Eloundou and Gretchen Krueger and Kevin Button and Matthew Knight and Benjamin Chess and John Schulman},
  journal={ArXiv},
  year={2021},
  volume={abs/2112.09332},
  url={https://api.semanticscholar.org/CorpusID:245329531}
}

@article{Schick2023ToolformerLM,
  title={Toolformer: Language Models Can Teach Themselves to Use Tools},
  author={Timo Schick and Jane Dwivedi-Yu and Roberto Dess{\`i} and Roberta Raileanu and Maria Lomeli and Luke Zettlemoyer and Nicola Cancedda and Thomas Scialom},
  journal={ArXiv},
  year={2023},
  volume={abs/2302.04761},
  url={https://api.semanticscholar.org/CorpusID:256697342}
}

@inproceedings{Li2025Searcho1AS,
  title={Search-o1: Agentic Search-Enhanced Large Reasoning Models},
  author={Xiaoxi Li and Guanting Dong and Jiajie Jin and Yuyao Zhang and Yujia Zhou and Yutao Zhu and Peitian Zhang and Zhicheng Dou},
  booktitle={Conference on Empirical Methods in Natural Language Processing},
  year={2025},
  url={https://api.semanticscholar.org/CorpusID:275405676}
}

@article{Song2025R1SearcherIT,
  title={R1-Searcher: Incentivizing the Search Capability in LLMs via Reinforcement Learning},
  author={Huatong Song and Jinhao Jiang and Yingqian Min and Jie Chen and Zhipeng Chen and Wayne Xin Zhao and Lei Fang and Ji-Rong Wen},
  journal={ArXiv},
  year={2025},
  volume={abs/2503.05592},
  url={https://api.semanticscholar.org/CorpusID:276884818}
}

@inproceedings{Zheng2025DeepResearcherSD,
  title={DeepResearcher: Scaling Deep Research via Reinforcement Learning in Real-world Environments},
  author={Yuxiang Zheng and Dayuan Fu and Xiangkun Hu and Xiaojie Cai and Lyumanshan Ye and Pengrui Lu and Pengfei Liu},
  booktitle={Conference on Empirical Methods in Natural Language Processing},
  year={2025},
  url={https://api.semanticscholar.org/CorpusID:277596185}
}

@article{Zhang2025FromWS,
  title={From Web Search towards Agentic Deep Research: Incentivizing Search with Reasoning Agents},
  author={Weizhi Zhang and Yangning Li and Yuan-Qi Bei and Junyu Luo and Guancheng Wan and Liangwei Yang and Chenxuan Xie and Yuyao Yang and Wei-Chieh Huang and Chunyu Miao and Henry Peng Zou and Xiao Luo and Yusheng Zhao and Yankai Chen and Chunkit Chan and Peilin Zhou and Xinyang Zhang and Chenwei Zhang and Jingbo Shang and Ming Zhang and Yangqiu Song and Irwin King and Philip S. Yu},
  journal={ArXiv},
  year={2025},
  volume={abs/2506.18959},
  url={https://api.semanticscholar.org/CorpusID:279999700}
}

@article{Li2025WebSailorNS,
  title={WebSailor: Navigating Super-human Reasoning for Web Agent},
  author={Kuan Li and Zhongwang Zhang and Huifeng Yin and Liwen Zhang and Litu Ou and Jialong Wu and Wenbiao Yin and Baixuan Li and Zhengwei Tao and Xinyu Wang and Weizhou Shen and Junkai Zhang and Dingchu Zhang and Xixi Wu and Yong Jiang and Ming Yan and Pengjun Xie and Fei Huang and Jingren Zhou},
  journal={ArXiv},
  year={2025},
  volume={abs/2507.02592},
  url={https://api.semanticscholar.org/CorpusID:280078605}
}

@article{Huybrechts2025DocumentHA,
  title={Document Haystack: A Long Context Multimodal Image/Document Understanding Vision LLM Benchmark},
  author={Goeric Huybrechts and S. Ronanki and Sai Muralidhar Jayanthi and Jack G. M. Fitzgerald and Srinivasan R Veeravanallur},
  journal={2025 IEEE/CVF International Conference on Computer Vision Workshops (ICCVW)},
  year={2025},
  pages={4121-4129},
  url={https://api.semanticscholar.org/CorpusID:280166930}
}

@misc{xi2025infodeepseekbenchmarkingagenticinformation,
      title={InfoDeepSeek: Benchmarking Agentic Information Seeking for Retrieval-Augmented Generation}, 
      author={Yunjia Xi and Jianghao Lin and Menghui Zhu and Yongzhao Xiao and Zhuoying Ou and Jiaqi Liu and Tong Wan and Bo Chen and Weiwen Liu and Yasheng Wang and Ruiming Tang and Weinan Zhang and Yong Yu},
      year={2025},
      eprint={2505.15872},
      archivePrefix={arXiv},
      primaryClass={cs.IR},
      url={https://arxiv.org/abs/2505.15872}, 
}

@misc{gou2025mind2web2evaluatingagentic,
      title={Mind2Web 2: Evaluating Agentic Search with Agent-as-a-Judge}, 
      author={Boyu Gou and Zanming Huang and Yuting Ning and Yu Gu and Michael Lin and Weijian Qi and Andrei Kopanev and Botao Yu and Bernal Jiménez Gutiérrez and Yiheng Shu and Chan Hee Song and Jiaman Wu and Shijie Chen and Hanane Nour Moussa and Tianshu Zhang and Jian Xie and Yifei Li and Tianci Xue and Zeyi Liao and Kai Zhang and Boyuan Zheng and Zhaowei Cai and Viktor Rozgic and Morteza Ziyadi and Huan Sun and Yu Su},
      year={2025},
      eprint={2506.21506},
      archivePrefix={arXiv},
      primaryClass={cs.AI},
      url={https://arxiv.org/abs/2506.21506}, 
}

@misc{chen2025browsecompplusfairtransparentevaluation,
      title={BrowseComp-Plus: A More Fair and Transparent Evaluation Benchmark of Deep-Research Agent}, 
      author={Zijian Chen and Xueguang Ma and Shengyao Zhuang and Ping Nie and Kai Zou and Andrew Liu and Joshua Green and Kshama Patel and Ruoxi Meng and Mingyi Su and Sahel Sharifymoghaddam and Yanxi Li and Haoran Hong and Xinyu Shi and Xuye Liu and Nandan Thakur and Crystina Zhang and Luyu Gao and Wenhu Chen and Jimmy Lin},
      year={2025},
      eprint={2508.06600},
      archivePrefix={arXiv},
      primaryClass={cs.CL},
      url={https://arxiv.org/abs/2508.06600}, 
}

@misc{xu2025ravinerealityalignedevaluationagentic,
      title={RAVine: Reality-Aligned Evaluation for Agentic Search}, 
      author={Yilong Xu and Xiang Long and Zhi Zheng and Jinhua Gao},
      year={2025},
      eprint={2507.16725},
      archivePrefix={arXiv},
      primaryClass={cs.CL},
      url={https://arxiv.org/abs/2507.16725}, 
}

@misc{wong2025widesearchbenchmarkingagenticbroad,
      title={WideSearch: Benchmarking Agentic Broad Info-Seeking}, 
      author={Ryan Wong and Jiawei Wang and Junjie Zhao and Li Chen and Yan Gao and Long Zhang and Xuan Zhou and Zuo Wang and Kai Xiang and Ge Zhang and Wenhao Huang and Yang Wang and Ke Wang},
      year={2025},
      eprint={2508.07999},
      archivePrefix={arXiv},
      primaryClass={cs.CL},
      url={https://arxiv.org/abs/2508.07999}, 
}

@misc{lan2025deepwidesearchbenchmarkingdepthwidth,
      title={DeepWideSearch: Benchmarking Depth and Width in Agentic Information Seeking}, 
      author={Tian Lan and Bin Zhu and Qianghuai Jia and Junyang Ren and Haijun Li and Longyue Wang and Zhao Xu and Weihua Luo and Kaifu Zhang},
      year={2025},
      eprint={2510.20168},
      archivePrefix={arXiv},
      primaryClass={cs.CL},
      url={https://arxiv.org/abs/2510.20168}, 
}

@misc{zhang2026browsecompv3visualverticalverifiable,
      title={BrowseComp-$V^3$: A Visual, Vertical, and Verifiable Benchmark for Multimodal Browsing Agents}, 
      author={Huanyao Zhang and Jiepeng Zhou and Bo Li and Bowen Zhou and Yanzhe Shan and Haishan Lu and Zhiyong Cao and Jiaoyang Chen and Yuqian Han and Zinan Sheng and Zhengwei Tao and Hao Liang and Jialong Wu and Yang Shi and Yuanpeng He and Jiaye Lin and Qintong Zhang and Guochen Yan and Runhao Zhao and Zhengpin Li and Xiaohan Yu and Lang Mei and Chong Chen and Wentao Zhang and Bin Cui},
      year={2026},
      eprint={2602.12876},
      archivePrefix={arXiv},
      primaryClass={cs.AI},
      url={https://arxiv.org/abs/2602.12876}, 
}

@misc{ning2026mcsearchevaluatingenhancingmultimodal,
      title={MC-Search: Evaluating and Enhancing Multimodal Agentic Search with Structured Long Reasoning Chains}, 
      author={Xuying Ning and Dongqi Fu and Tianxin Wei and Mengting Ai and Jiaru Zou and Ting-Wei Li and Hanghang Tong and Yada Zhu and Hendrik Hamann and Jingrui He},
      year={2026},
      eprint={2603.00873},
      archivePrefix={arXiv},
      primaryClass={cs.AI},
      url={https://arxiv.org/abs/2603.00873}, 
}

@misc{gupta2026deepsearchqabridgingcomprehensivenessgap,
      title={DeepSearchQA: Bridging the Comprehensiveness Gap for Deep Research Agents}, 
      author={Nikita Gupta and Riju Chatterjee and Lukas Haas and Connie Tao and Andrew Wang and Chang Liu and Hidekazu Oiwa and Elena Gribovskaya and Jan Ackermann and John Blitzer and Sasha Goldshtein and Dipanjan Das},
      year={2026},
      eprint={2601.20975},
      archivePrefix={arXiv},
      primaryClass={cs.CL},
      url={https://arxiv.org/abs/2601.20975}, 
}

@misc{tao2026mmsearchplusbenchmarkingprovenanceawaresearch,
      title={MMSearch-Plus: Benchmarking Provenance-Aware Search for Multimodal Browsing Agents}, 
      author={Xijia Tao and Yihua Teng and Xinxing Su and Xinyu Fu and Jihao Wu and Chaofan Tao and Ziru Liu and Haoli Bai and Rui Liu and Lingpeng Kong},
      year={2026},
      eprint={2508.21475},
      archivePrefix={arXiv},
      primaryClass={cs.AI},
      url={https://arxiv.org/abs/2508.21475}, 
}

@misc{ye2026miroevalbenchmarkingmultimodaldeep,
      title={MiroEval: Benchmarking Multimodal Deep Research Agents in Process and Outcome}, 
      author={Fangda Ye and Yuxin Hu and Pengxiang Zhu and Yibo Li and Ziqi Jin and Yao Xiao and Yibo Wang and Lei Wang and Zhen Zhang and Lu Wang and Yue Deng and Bin Wang and Yifan Zhang and Liangcai Su and Xinyu Wang and He Zhao and Chen Wei and Qiang Ren and Bryan Hooi and An Bo and Shuicheng Yan and Lidong Bing},
      year={2026},
      eprint={2603.28407},
      archivePrefix={arXiv},
      primaryClass={cs.AI},
      url={https://arxiv.org/abs/2603.28407}, 
}

@misc{huang2026mmdeepresearchbenchbenchmarkmultimodaldeep,
      title={MMDeepResearch-Bench: A Benchmark for Multimodal Deep Research Agents}, 
      author={Peizhou Huang and Zixuan Zhong and Zhongwei Wan and Donghao Zhou and Samiul Alam and Xin Wang and Zexin Li and Zhihao Dou and Li Zhu and Jing Xiong and Chaofan Tao and Yan Xu and Dimitrios Dimitriadis and Tuo Zhang and Mi Zhang},
      year={2026},
      eprint={2601.12346},
      archivePrefix={arXiv},
      primaryClass={cs.CV},
      url={https://arxiv.org/abs/2601.12346}, 
}
\bibliographystyle{tmlr}

\newpage
\appendix
\section{Construction Workflow and Internal Interfaces}

This appendix provides supporting details on implementation and analysis, including the annotation and evaluation interfaces, the AI-assisted construction procedure, metric and judge specifications, model configurations, robustness checks, additional benchmark statistics, diagnostic analyses, case studies, and release considerations.

\zh{
本附录提供实现与分析方面的补充细节，包括标注与评测界面、AI-assisted 构造流程、指标与 judge 规范、模型配置、稳健性检验、额外 benchmark 统计、诊断分析、案例研究和发布相关考量。
}

\subsection{Annotation and Quality-Control Interface}

The internal annotation platform supports collaborative data construction and quality control. Within a single interface, annotators create and revise benchmark items by editing the question, short answer, hidden Node-Relation design, checklist, sources, images, category, authored-template label, and operation tags. The platform also supports Markdown rendering of the design field, image inspection, secondary-audit comments and ratings, and benchmark locking. Together, these functions operationalize the validation protocol: an independent auditor can check whether the question leaks the path, whether each checklist item constitutes a verifiable conclusion, whether each source supports a necessary node or relation, and whether uploaded images or other non-text evidence are used in the chain.

\zh{
内部标注平台支持协作式数据构造和质量控制。标注者可以在同一界面中编辑 question、short answer、隐藏的 Node-Relation design、checklist、sources、images、category、authored-template 标签和 operation tags，以创建和修改 benchmark item。平台还支持 design 字段的 Markdown 渲染、图像检查、二次审核的 comments 与 ratings，以及 benchmark locking。这些功能共同将验证协议落实为具体操作：独立审核者可以检查 question 是否泄露路径、每个 checklist item 是否构成可验证结论、每个 source 是否支撑必要的 node 或 relation，以及上传图像或其他非文本证据是否用于链路中。
}

\begin{figure}[t]
\centering
\includegraphics[width=0.9\linewidth]{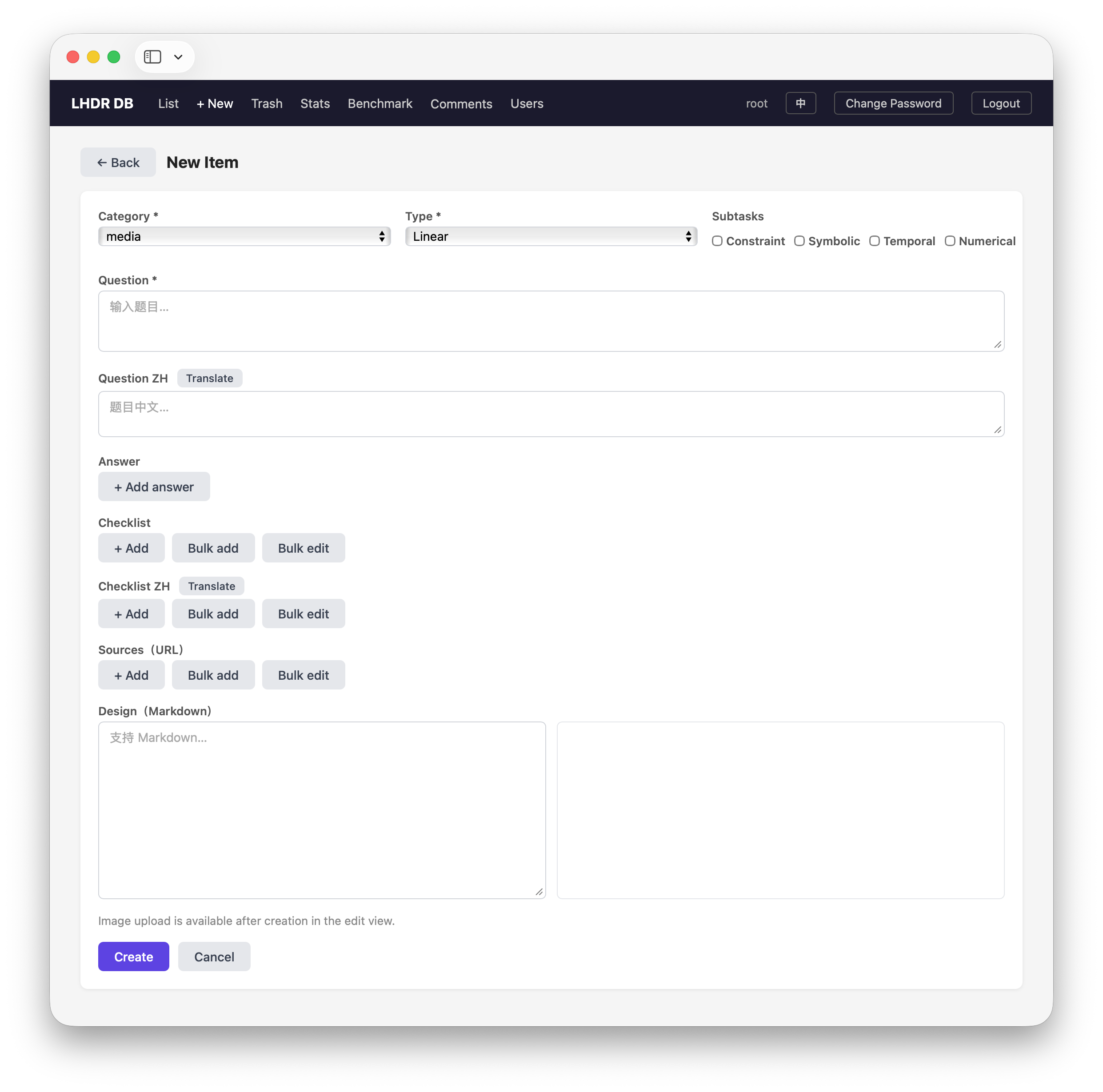}
\caption{Annotation interface for collaborative \ourmethodname construction. Before an item passes the secondary audit and is locked, the system allows annotators to inspect and edit its question, answer, design, checklist, sources, images, and metadata.\zh{ 用于协作式 \ourmethodname 构造的标注界面。在 item 通过二次审核并被锁定之前，系统允许标注者检查和编辑其 question、answer、design、checklist、sources、images 和 metadata。}}
\label{fig:appendix_annotation_interface}
\end{figure}

\subsection{Evaluation Interface}

The evaluation platform supports launching model runs, monitoring their status, inspecting model responses, and auditing judge outputs. For each item--model pair, the platform records whether the run completed or failed, the model response, the judge's final-answer decision, and the checklist completion vector.

\zh{
评测平台支持启动模型运行、监控运行状态、检查模型回复，以及审计 judge 输出。对于每个 item--model pair，平台记录该运行是完成还是失败、模型回复、judge 的最终答案判定，以及 checklist completion vector。
}

\begin{figure}[t]
\centering
\includegraphics[width=0.9\linewidth]{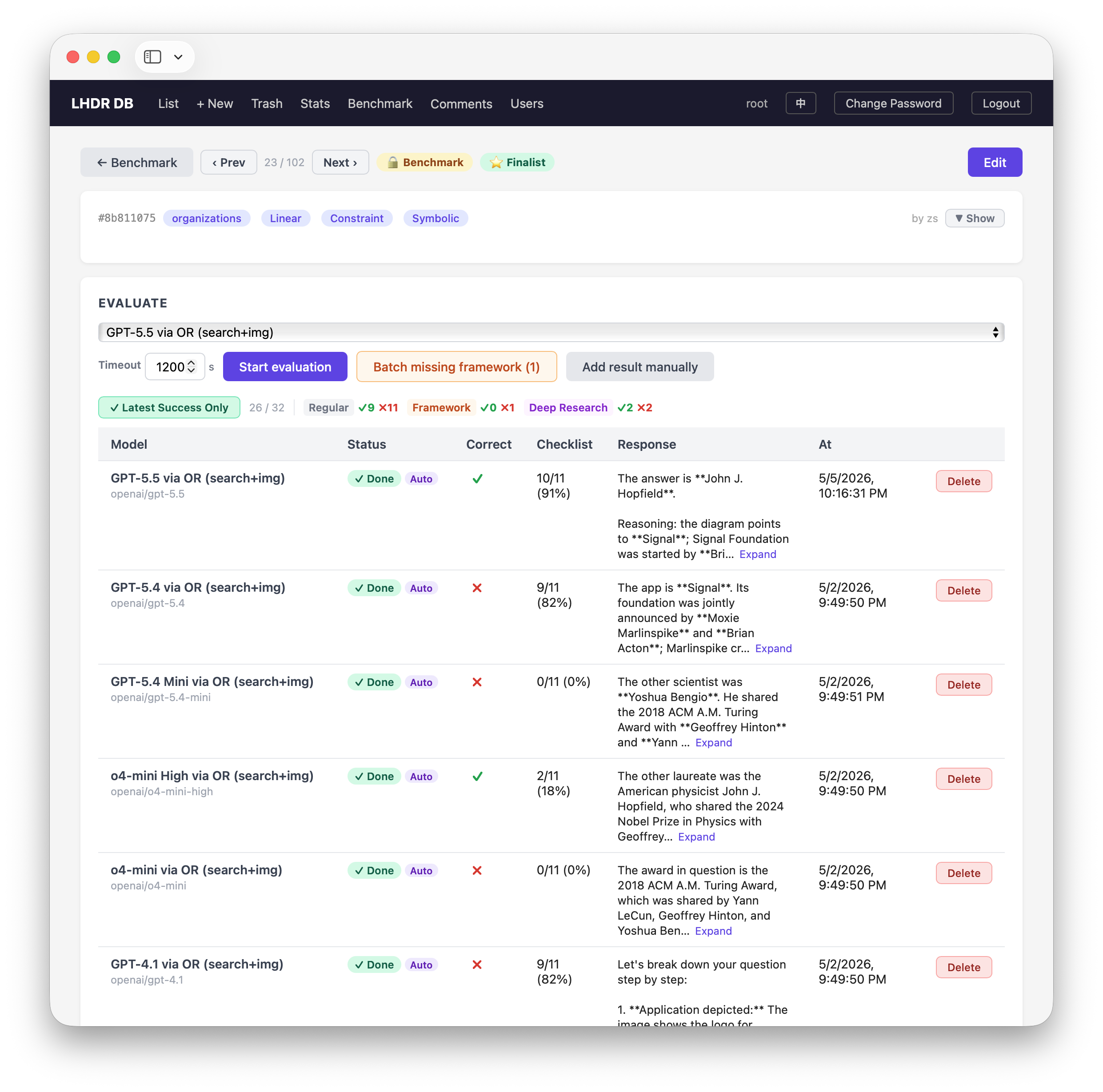}
\caption{Evaluation interface for launching runs and inspecting their status, responses, judge decisions, checklist completion, and aggregate metrics. The platform serves as an execution and audit tool. \zh{ 用于启动运行，并检查其状态、回复、judge 判定、checklist completion 和聚合指标的评测界面。该平台用于执行和审计。}}
\label{fig:appendix_evaluation_interface}
\end{figure}

\subsection{Dataset Schema and Construction Procedure}

\begin{table}[t]
\centering
\footnotesize
\setlength{\tabcolsep}{2pt}
\newcommand{\yesmark}{\ensuremath{\checkmark}}
\newcommand{\nomark}{\ensuremath{\times}}
\caption{Core fields and annotations in each \ourmethodname item. The \texttt{checklist} records necessary intermediate conclusions rather than browser actions; the checklist DAG records their verified prerequisite relations; \texttt{sources} support the nodes, relations, checklist items, and final answer; \texttt{images} provide non-textual entry points or key evidence.\zh{ 每个 \ourmethodname 条目的核心字段与标注。\texttt{checklist} 记录必要中间结论，而不是浏览器操作；checklist DAG 记录这些结论之间经核定的前置依赖关系；\texttt{sources} 支撑节点、关系、checklist item 和最终答案；\texttt{images} 提供非文本入口或关键证据。}}
\label{tab:dataset_schema}
\begin{tabularx}{\linewidth}{@{}ll>{\raggedright\arraybackslash}Xcc@{}}
\toprule
Field & Type & Role & Eval. & Verified \\
\midrule
\texttt{question} & text & Natural-language prompt shown to the agent, with the hidden path and answer masked. & \yesmark & \yesmark \\
\texttt{answer} & list/text & Short, unique, stable reference answer such as a person, place, organization, number, year, color, or short phrase. & \yesmark & \yesmark \\
\texttt{checklist} & list & Necessary intermediate conclusions required to derive the answer; these are not action logs. & \yesmark & \yesmark \\
\texttt{checklist DAG} & DAG & Verified prerequisite relations among checklist conclusions, used to measure dependency depth and compute DACS. & \yesmark & \yesmark \\
\texttt{sources} & list & URLs supporting nodes, relations, checklist items, and the final answer. & \yesmark & \yesmark \\
\texttt{images} & list & Non-textual entry points or key evidence, including images, maps, screenshots, logos, charts, or document pages. & \yesmark & \yesmark \\
\texttt{category} & label & Topical category used for coverage control and diagnostic analysis. & \nomark & \yesmark \\
\texttt{type} & label & Coarse authored template: Linear or Multi-branch; not the exact verified DAG topology. & \nomark & \yesmark \\
\texttt{subtasks} & labels & Operation tags indicating local transformations inside the chain. & \nomark & \yesmark \\
\bottomrule
\end{tabularx}
\end{table}

A node consists of a core entity paired with searchable and verifiable properties, such as a photo, map location, paper PDF, official document, logo, chart, timestamp, work, affiliation, venue, or founder. A relation uses a verified property of one node as the entry condition for another node.

\zh{
    Node 由一个核心实体及其可检索、可验证的属性组成，例如照片、地图位置、论文 PDF、官方文档、logo、图表、时间戳、作品、机构归属、场所或创始人。Relation 将一个 node 中已验证的属性用作另一个 node 的入口条件。
}

\subsection{AI-Assisted Construction Pipeline}
\label{sec:appendix_ai_assisted_pipeline}

AI assistance follows an expand-then-constrain process: the model broadens the candidate design space, while annotators determine which elements remain as benchmark evidence. The first layer consists of human-controlled task specification: before any model generation, annotators define the target category, operation tags, and golden-example style. Within these constraints, AI supports graph-level exploration by proposing Node-Relation graphs comprising five to six information-rich nodes, verifiable properties, possible multimodal sources, and bridge relations. AI is then used for checklist-level stress testing, expanding the graph into an intentionally over-complete set of candidate conclusions, often spanning 25--28 steps, to expose missing evidence, weak dependencies, duplicate facts, and potential shortcuts. Following DeepDive's \emph{Blur Attribute-Rich Path} strategy~\citep{lu2025deepdiveadvancingdeepsearch}, annotators also use AI to propose coarser-grained renderings of high-information attributes while preserving the unique convergence of the reasoning path: an exact year such as 1978 may be rewritten as ``the late 1970s,'' while overly direct place or affiliation cues may be abstracted into recoverable descriptors. Any blur is rejected if it introduces another valid answer or renders a necessary bridge unverifiable. Annotators complete this layer by compressing the candidate chain: they remove background facts, merge redundant conclusions, and retain only the intermediate conclusions necessary to derive the final answer.

\zh{
AI assistance 采用 expand-then-constrain 流程：模型扩大候选设计空间，标注者则决定哪些内容可以作为 benchmark evidence 保留下来。第一层是由人工控制的任务设定：在任何模型生成之前，标注者先确定目标 category、operation tag 和 golden-example 风格。在这些约束下，AI 用于 graph-level exploration，提出由五到六个信息丰富节点、可验证属性、潜在多模态来源和桥接 relation 构成的 Node-Relation graph。随后，AI 用于 checklist-level stress testing，将图结构扩展为有意超量的候选结论集合，通常涵盖 25--28 步，以暴露证据缺口、弱依赖、重复事实和潜在捷径。参考 DeepDive 的 \emph{Blur Attribute-Rich Path} 策略~\citep{lu2025deepdiveadvancingdeepsearch}，标注者还会在保持推理路径唯一收敛的前提下，使用 AI 为高信息量属性提出更粗粒度的表达：例如将精确年份 1978 改写为 ``the late 1970s''，或将过于直接的地点或机构线索抽象为仍可据以检索还原的描述。任何会引入另一个有效答案或使必要桥接关系不可验证的模糊化都会被拒绝。最后，标注者压缩候选链路以完成这一层：删除背景事实、合并冗余结论，并只保留推出最终答案所必需的中间结论。
}

The second layer converts the pruned chain into a grounded benchmark item. Annotators draft or revise the natural-language question, short answer, source list, image evidence, and metadata while verifying that inspectable evidence supports every necessary conclusion. Question polishing is treated as leakage control rather than stylistic rewriting: annotators remove intermediate-node hints, direct-answer cues, and wording that reveals the intended search path. The internal annotation platform facilitates collaborative editing, secondary-audit ratings and comments, image inspection, and the benchmark admission workflow. Thus, AI provides breadth during candidate exploration, whereas human annotators determine factual support, checklist irreducibility, multimodal necessity, shortcut resistance, label correctness, and final admission to the evaluation set.

\zh{
第二层将剪枝后的链路转化为有证据支撑的 benchmark item。标注者起草或修订自然语言问题、简短答案、source list、image evidence 和 metadata，同时验证每个必要结论都有可检查的证据支撑。Question polishing 被视为 leakage control，而非单纯的文风润色：标注者会移除中间节点提示、直接答案线索，以及暴露预期搜索路径的表述。内部标注平台支持多人协同编辑、二次审核的 ratings/comments、图片检查和 benchmark 准入流程。因此，AI 在候选探索阶段提供广度，而人工标注者负责判定事实支撑、checklist 不可约性、多模态必要性、捷径抵抗性、标签正确性，以及样本是否最终进入 evaluation set。
}

The pipeline is interactive and does not correspond to a single generation prompt. Model assistance is introduced at several distinct points---task specification, graph-level exploration, checklist stress testing, and blur rewriting. Each point is conditioned on the item under construction and guided by an annotator who accepts, edits, or discards the output. The expansion stage deliberately over-produces candidate checklists of 25--28 steps; the 102 admitted items retain 1,231 checklist steps in total, averaging 12.1 per item. Fewer than half of the proposed conclusions survive pruning, consistent with the expand-then-constrain design, in which annotators remove most model proposals.

\zh{
该 pipeline 是交互式流程，不对应某一条生成 prompt。模型辅助介入任务设定、graph-level exploration、checklist stress testing 与属性模糊化改写等多个不同环节。每个环节都以正在构造的条目为条件，并由标注者主导，由其接受、修改或丢弃模型输出。扩展阶段刻意超量生成包含 25--28 步的候选 checklist；最终入选的 102 个条目共保留 1,231 个 checklist step，平均每题 12.1 步。不到一半由模型提出的结论通过剪枝，这与 expand-then-constrain 设计一致，即由标注者删除大多数模型建议。
}

\section{Benchmark Statistics and Composition}
\label{sec:appendix_dataset_statistics}

\begin{table}[t]
\centering
\scriptsize
\setlength{\tabcolsep}{3pt}
\caption{Overview statistics for the \ourmethodname benchmark. Operation tags are multi-label, so their percentages need not sum to 100\%. Non-text evidence coverage counts either uploaded image assets or source URLs pointing to PDF, video, or map evidence.\zh{ \ourmethodname benchmark 的总体统计。Operation tag 是多标签，因此比例不需要相加为 100\%。Non-text evidence coverage 统计上传的 image asset，或指向 PDF、video、map evidence 的 source URL。}}
\label{tab:dataset_statistics}
\begin{tabular}{lp{0.62\linewidth}}
\toprule
Statistic & Value \\
\midrule
Benchmark items & 102 \\
Checklist annotations & 102 items (100.0\%); 1,231 checklist steps in total \\
Checklist length & Mean 12.1, median 11, range 10--21 \\
Dependency depth & Mean 10.4, median 10, range 6--18 \\
Label space & 8 categories, 2 authored templates, 4 operation tags \\
Authored templates & Linear: 73 (71.6\%); Multi-branch: 29 (28.4\%) \\
Operation tags & Constraint: 76 (74.5\%); Numerical: 30 (29.4\%); Symbolic: 81 (79.4\%); Temporal: 18 (17.6\%) \\
Checklist length buckets & 10--12: 69 (67.6\%); 13--15: 23 (22.5\%); 16+: 10 (9.8\%) \\
Dependency depth buckets & 6--9: 35 (34.3\%); 10--11: 38 (37.3\%); 12--18: 29 (28.4\%) \\
Source URLs & 818 total; mean 8.0 per item \\
Image assets & 101 total; mean 1.0 per item; 101 items (99.0\%) have an uploaded image asset \\
Non-text evidence coverage & 102 items (100.0\%), counting uploaded image assets or PDF/video/map source URLs \\
Source diagnostics & PDF/video/map source URLs appear in 45 items (44.1\%); Wikipedia URLs account for 241/818 source URLs (29.5\%) \\
\bottomrule
\end{tabular}
\end{table}

\begingroup
\scriptsize
\setlength{\tabcolsep}{3pt}
\begin{longtable}{llrrrr}
\caption{Benchmark split statistics. Operation tags are multi-label, so tag counts
and percentages do not sum to 102 items or 100\%.}\label{tab:appendix_dataset_splits}\\
\toprule
Scope & Group & Items & Percent & Avg. checklist & Avg. sources \\
\midrule
\endfirsthead
\caption[]{Benchmark split statistics. (continued)}\\
\toprule
Scope & Group & Items & Percent & Avg. checklist & Avg. sources \\
\midrule
\endhead
category & academia & 10 & 9.8 & 12.3 & 8.7 \\
category & geography & 17 & 16.7 & 14.4 & 9.2 \\
category & media & 21 & 20.6 & 11.9 & 8.7 \\
category & organizations & 13 & 12.7 & 11.2 & 7.2 \\
category & people & 12 & 11.8 & 11.0 & 6.6 \\
category & society & 10 & 9.8 & 11.2 & 6.7 \\
category & sports & 9 & 8.8 & 12.1 & 8.0 \\
category & technology & 10 & 9.8 & 11.6 & 8.1 \\
\midrule
authored template & Linear & 73 & 71.6 & 11.8 & 7.9 \\
authored template & Multi-branch & 29 & 28.4 & 12.8 & 8.3 \\
\midrule
operation tag & Constraint & 76 & 74.5 & 11.9 & 8.1 \\
operation tag & Numerical & 30 & 29.4 & 11.8 & 7.0 \\
operation tag & Symbolic & 81 & 79.4 & 12.2 & 8.2 \\
operation tag & Temporal & 18 & 17.6 & 11.6 & 7.2 \\
\midrule
checklist length & 10 & 35 & 34.3 & 10.0 & 7.3 \\
checklist length & 11--12 & 34 & 33.3 & 11.4 & 7.6 \\
checklist length & 13--15 & 23 & 22.5 & 13.7 & 9.1 \\
checklist length & 16--18 & 6 & 5.9 & 16.7 & 8.8 \\
checklist length & 19+ & 4 & 3.9 & 20.0 & 10.8 \\
\midrule
dependency depth & 6--9 & 35 & 34.3 & 11.1 & 8.1 \\
dependency depth & 10--11 & 38 & 37.3 & 11.1 & 7.7 \\
dependency depth & 12--18 & 29 & 28.4 & 14.4 & 8.3 \\
\midrule
source count & 1--6 & 27 & 26.5 & 11.6 & 5.0 \\
source count & 7--8 & 37 & 36.3 & 11.3 & 7.6 \\
source count & 9--10 & 25 & 24.5 & 12.4 & 9.6 \\
source count & 11+ & 13 & 12.7 & 14.7 & 12.5 \\
\bottomrule
\end{longtable}
\endgroup

Table~\ref{tab:dataset_statistics} summarizes the \ourmethodname evaluation set, while Table~\ref{tab:appendix_dataset_splits} presents benchmark diagnostics at the split level. The benchmark currently contains 102 items. Their checklists comprise 1,231 necessary intermediate conclusions in total, with a mean length of 12.1, a median length of 11, and a range of 10 to 21 steps. In the author-verified checklist DAGs, dependency depth---defined as the maximum number of conclusions along a prerequisite path, including both endpoints---has a mean of 10.4, a median of 10, and a range of 6 to 18. The evidence fields are similarly dense: the benchmark includes 818 source URLs, with an average of 8.0 URLs per item, as well as 101 image assets. Together, the short answers, complete checklist annotations, explicit dependencies, and evidence fields make \ourmethodname verifiable at both the final-answer and process levels.

\zh{
Table~\ref{tab:dataset_statistics} 总结了 \ourmethodname 的评测集，Table~\ref{tab:appendix_dataset_splits} 则给出了划分层面的基准诊断结果。当前基准包含 102 个条目，其 checklist 共涵盖 1,231 个必要的中间结论；checklist 长度的平均值为 12.1，中位数为 11，范围为 10 到 21 步。在经作者核验的 checklist DAG 中，dependency depth 定义为一条前置依赖路径上的最大结论数（包括路径的两个端点），其平均值为 10.4，中位数为 10，范围为 6 到 18。证据字段同样较为密集：基准共包含 818 个来源 URL，平均每个条目 8.0 个，另有 101 个图像资产。短答案、完整的 checklist 标注、显式依赖关系和证据字段共同使 \ourmethodname 能够在最终答案和过程两个层面接受验证。
}

The construction metadata covers all eight target categories and both coarse authored-template labels. The largest category is media, with 21 items, followed by geography with 17; the smallest is sports, with 9 items. Each remaining category contains between 10 and 13 items. Most items follow the Linear template: 73 items (71.6\%) use this template, while 29 items (28.4\%) use the Multi-branch template, which was designed to emphasize multiple evidence paths and their subsequent joint use. Operation tags characterize local demands within a chain rather than mutually exclusive task families: Symbolic appears in 81 items (79.4\%), Constraint in 76 (74.5\%), Numerical in 30 (29.4\%), and Temporal in 18 (17.6\%). Because these tags are multi-label, their percentages need not sum to 100\%.

\zh{
构造元数据覆盖全部八个目标类别和两种粗粒度 authored-template 标签。规模最大的类别是 media，包含 21 个条目；其次是 geography，包含 17 个条目；规模最小的是 sports，包含 9 个条目；其余每个类别包含 10 到 13 个条目。大多数条目采用 Linear 模板：73 个条目（71.6\%）采用该模板，另有 29 个条目（28.4\%）采用 Multi-branch 模板；后者旨在突出多条证据路径及其后续的联合使用。Operation tag 刻画链路内部的局部要求，而非相互排斥的任务类别：Symbolic 出现在 81 个条目中（79.4\%），Constraint 出现在 76 个条目中（74.5\%），Numerical 出现在 30 个条目中（29.4\%），Temporal 出现在 18 个条目中（17.6\%）。由于这些标签采用多标签形式，其比例之和不必为 100\%。
}

Dependency depth and checklist length capture complementary structural properties, rather than the length of browsing traces or the verbosity of prompts. Dependency depth measures the longest sequential prerequisite path, whereas checklist length measures the total number of necessary conclusions, including those on parallel branches. By construction, no benchmark item has fewer than 10 checklist steps. Most items fall into the 10--12-step bucket (69 items, 67.6\%); 23 items (22.5\%) contain 13--15 steps, and 10 items (9.8\%) contain 16 or more steps. Longer checklists also tend to have denser source evidence: items with 16 or more checklist steps use an average of 9.6 source URLs, compared with 7.4 URLs for items in the 10--12-step bucket.

\zh{
Dependency depth 与 checklist length 刻画了两种互补的结构属性，而非浏览轨迹的长度或 prompt 的冗长程度。Dependency depth 衡量最长的顺序前置依赖路径，checklist length 则衡量必要结论的总数，其中包括并行分支上的结论。根据构造标准，所有基准条目均包含至少 10 个 checklist step。多数条目位于 10--12 步区间（69 个，67.6\%）；23 个条目（22.5\%）包含 13--15 步，10 个条目（9.8\%）包含 16 步或更多。较长的 checklist 通常也具有更密集的来源证据：包含 16 个及以上 checklist step 的条目平均使用 9.6 个来源 URL，而 10--12 步区间内的条目平均使用 7.4 个。
}

For subsequent analysis, \ourmethodname tracks category, authored-template label, operation tags, checklist length, source count, image availability, coarse-grained non-text evidence coverage, and Wikipedia usage. Every item contains non-text evidence when uploaded images are considered together with PDF, video, or map source URLs. PDF, video, or map URLs appear in 45 items (44.1\%), while Wikipedia accounts for 241 of the 818 source URLs (29.5\%). These statistics support the claim that \ourmethodname cannot be reduced to a Wikipedia-only or text-only browsing benchmark.

\zh{
在后续分析中，\ourmethodname 跟踪类别、人工设计模板标签、operation tag、checklist 长度、来源数量、图像可用性、粗粒度非文本证据覆盖情况以及 Wikipedia 使用情况。将上传图像与 PDF、视频或地图来源 URL 一并考虑时，每个条目都包含非文本证据。PDF、视频或地图 URL 出现在 45 个条目中（44.1\%），Wikipedia 来源则占全部 818 个来源 URL 中的 241 个（29.5\%）。这些统计结果支持如下判断：\ourmethodname 不能被化约为仅使用 Wikipedia 或仅进行文本浏览的基准。
}

\section{Evaluation Configuration and Failure Handling}

\subsection{Model and Provider Configuration}

Table~\ref{tab:appendix_model_configuration} lists the complete model configuration used in the evaluation. The evaluation was conducted from April 30 to May 5, 2026 (UTC). Regular OpenRouter models in the search+image group were evaluated through the benchmark interface with image evidence and, when exposed by the provider route, web-search capability; image-only models received image input but no search tool. Deep research systems were evaluated through their dedicated research endpoints over the same OpenRouter route, whereas framework agents like \citet{bytedance2026deerflow} were evaluated through DeerFlow backends. We implemented neither a ReAct loop nor any harness-level optimizations. Each run consisted of a single request containing the question verbatim. For the tool-augmented group, the request additionally declared the provider's native server-side search tool, ensuring that all browsing was executed and completed within the provider's service before the response was returned to us. We provided no system prompt and set no sampling parameters. Because the providers' own routes were used, the deep-research endpoints did not accept inline image parts; these systems therefore received the image evidence as a URL in the prompt text rather than as attached bytes.

\zh{
Table~\ref{tab:appendix_model_configuration} 列出了本文评测所使用的完整模型配置。评测于 2026 年 4 月 30 日至 5 月 5 日（UTC）进行。Search+image 组中的常规 OpenRouter 模型通过 benchmark interface 评测，并使用图像证据；当 provider route 开放相应能力时，这些模型还可使用 web-search capability。Image-only 模型接收图像输入，但不提供搜索工具。Deep research systems 通过同一条 OpenRouter 路由访问其专用 research endpoint，而 framework agents 如 \citet{bytedance2026deerflow} 则通过 DeerFlow backends 评测。我们既未实现 ReAct 循环，也未进行任何 harness 层面的优化。每次运行仅发起一次请求，并原样传入问题。对于 tool-augmented 组，请求还会声明 provider 自有的服务端搜索工具，确保所有浏览操作均在 provider 服务内部执行完毕后，再将响应返回给我们。我们未设置 system prompt 或任何采样参数。由于使用 provider 自有路由，deep-research endpoint 不接受内联图像部件；因此，这些系统接收的图像证据是 prompt 文本中的 URL，而非直接附加的字节。
}

\begin{table}[t]
\centering
\scriptsize
\setlength{\tabcolsep}{3pt}
\caption{Full model configuration. All reported metrics use the same fixed 102-item denominator.}
\label{tab:appendix_model_configuration}
\begin{tabularx}{\linewidth}{@{}>{\raggedright\arraybackslash}Xlll>{\raggedright\arraybackslash}X@{}}
\toprule
Model & Provider/interface & Group & Capability & Version or API identifier \\
\midrule
GPT-5.5 & OpenRouter & regular & search+image & \url{openai/gpt-5.5} \\
GPT-5.4 & OpenRouter & regular & search+image & \url{openai/gpt-5.4} \\
GPT-5.4 Mini & OpenRouter & regular & search+image & \url{openai/gpt-5.4-mini} \\
o4-mini & OpenRouter & regular & search+image & \url{openai/o4-mini} \\
GPT-4.1 & OpenRouter & regular & search+image & \url{openai/gpt-4.1} \\
GPT-4o (2024-11) & OpenRouter & regular & search+image & \url{openai/gpt-4o-2024-11-20} \\
GPT-4o Mini (2024-07) & OpenRouter & regular & search+image & \url{openai/gpt-4o-mini-2024-07-18} \\
o3 & OpenRouter & regular & search+image & \url{openai/o3} \\
Claude Opus 4.7 & OpenRouter & regular & search+image & \url{anthropic/claude-opus-4.7} \\
Claude Sonnet 4.6 & OpenRouter & regular & search+image & \url{anthropic/claude-sonnet-4.6} \\
Grok 4.20 & OpenRouter & regular & search+image & \url{x-ai/grok-4.20} \\
Gemini 3.1 Pro & OpenRouter & regular & image & \url{google/gemini-3.1-pro-preview} \\
Gemini 3 Flash & OpenRouter & regular & image & \url{google/gemini-3-flash-preview} \\
Gemini 2.5 Pro & OpenRouter & regular & image & \url{google/gemini-2.5-pro-preview} \\
Gemini 2.5 Flash & OpenRouter & regular & image & \url{google/gemini-2.5-flash} \\
Llama 4 Maverick & OpenRouter & regular & image & \url{meta-llama/llama-4-maverick} \\
Qwen3-VL-32B & OpenRouter & regular & image & \url{qwen/qwen3-vl-32b-instruct} \\
Qwen2.5-VL-72B & OpenRouter & regular & image & \url{qwen/qwen2.5-vl-72b-instruct} \\
o3 Deep Research & OpenRouter & deep research & deep research & \begin{tabular}[t]{@{}l@{}}\url{openai/o3-deep-research}\\\url{2025-06-26}\end{tabular} \\
o4-mini Deep Research & OpenRouter & deep research & deep research & \begin{tabular}[t]{@{}l@{}}\url{openai/o4-mini-deep-research}\\\url{2025-06-26}\end{tabular} \\
Tongyi DeepResearch 30B & OpenRouter & deep research & deep research & \begin{tabular}[t]{@{}l@{}}\url{alibaba/tongyi-deepresearch}\\\url{30b-a3b}\end{tabular} \\
Perplexity Sonar Deep Research & OpenRouter & deep research & deep research & \url{perplexity/sonar-deep-research} \\
DeerFlow (kimi-k2.5) & DeerFlow & framework/agent & local agent framework & \url{deerflow/kimi-k2.5} \\
DeerFlow (minimax-m2.7) & DeerFlow & framework/agent & local agent framework & \url{deerflow/minimax-m2.7} \\
DeerFlow (qwen3-vl-235b) & DeerFlow & framework/agent & local agent framework & \url{deerflow/qwen3-vl-235b} \\
\bottomrule
\end{tabularx}
\end{table}

\section{Metrics and Judge Specification}
\label{sec:appendix_metric_definitions}

\subsection{Interpretation of CS and DACS}
\label{sec:appendix_cs_limitations}
Following prior checklist-based evaluations in browsing benchmarks, CS provides a useful measure of broad coverage but is not well aligned with long-horizon deep research because it treats checklist items as unordered and interchangeable. In \ourmethodname, downstream conclusions are not independent local facts: they are warranted only after the preceding entities, constraints, and disambiguating relations have been established. When these prerequisites are absent, downstream statements that match individual checklist items should not be interpreted as evidence of successful reasoning, because they may reflect unsupported guesses, search shortcuts, or chance agreement rather than a valid continuation of the evidence path. CS can therefore overestimate process reliability by aggregating local matches across a broken dependency graph. DACS addresses this limitation while retaining credit for genuinely independent branches.

\zh{
沿用既有 browsing benchmark 中的 checklist-based evaluation，CS 能够有效衡量宽泛覆盖度，但与 long-horizon deep research 的目标并不十分契合，因为它将 checklist item 视为无序且可交换的单元。在 \ourmethodname 中，下游结论并非相互独立的局部事实：只有在前序实体、约束和消歧关系均已建立后，这些结论才有充分依据。如果缺少这些前提，即使下游陈述与个别 checklist item 匹配，也不应视为成功推理的证据，因为这些陈述可能源于无支撑的猜测、搜索捷径或偶然命中，而非证据路径的有效延续。因此，CS 可能因聚合断裂依赖图上的局部匹配而高估过程可靠性。DACS 在保留真正独立分支得分的同时弥补了这一局限。
}

The DACS recursion defined in Section~\ref{sec:metrics} scores each verified dependency DAG directly, independently of its coarse authored-template label. Independent subgraphs retain credit separately, whereas a merge node receives credit only when the node itself and all required branch dependencies are satisfied. DACS measures whether the submitted response states dependency-consistent conclusions.

\zh{
第~\ref{sec:metrics} 节定义的 DACS 递归直接对每张经核定的依赖 DAG 计分，不受其粗粒度 authored-template 标签影响。相互独立的子图分别保留得分，而 merge node 只有在其自身及所有必要分支依赖均满足时才获得 credit。DACS 衡量提交的回复是否陈述了具有依赖一致性的结论。
}

\subsection{Automatic Judge Specification}

The automatic judge receives the user question, reference answer, reference checklist, and model response. The metric script reads only two machine-readable fields from the judge output: \texttt{OVERALL\_CORRECTNESS} and \texttt{CHECKLIST\_RESULT}. The judge model in the current pipeline is Qwen3-VL-235B (\texttt{Qwen/Qwen3-VL-235B-A22B-Instruct-FP8}). When the core answer is unchanged, answer normalization permits harmless differences in capitalization, punctuation, spelling variants, aliases, units, and brief paraphrases. A checklist item receives credit only when the response states the corresponding conclusion completely and associates it with the correct entity and constraint; vague mentions, incorrect entity associations, missing qualifiers, and descriptions of search actions do not count as completion. All reported verdicts are model-produced; neither manual result auditing nor calibration against human labels is performed.

\zh{
自动 judge 接收用户问题、reference answer、reference checklist 和模型回复。指标脚本只读取 judge 输出中的两个机器可读字段：\texttt{OVERALL\_CORRECTNESS} 和 \texttt{CHECKLIST\_RESULT}。当前流程中的 judge model 是 Qwen3-VL-235B（\texttt{Qwen/Qwen3-VL-235B-A22B-Instruct-FP8}）。在核心答案不变的情况下，答案规范化允许大小写、标点、拼写变体、别名、单位和简短改写等无害差异。只有当回复完整陈述对应结论，并将其关联到正确实体和约束时，checklist item 才获得 credit；模糊提及、错误实体关联、缺少限定条件以及对搜索动作的描述均不计为完成。本文报告的判定全部由模型产生，既未进行人工结果审计，也未依据人工标签进行校准。
}

\subsection{Exact LLM Judge Prompt}

Figure~\ref{fig:llm_judge_prompt} presents the exact prompt used by the automatic judge, including the required machine-readable fields for final-answer correctness and checklist completion.

\zh{
Figure~\ref{fig:llm_judge_prompt} 展示了 automatic judge 实际使用的完整 prompt，其中包含最终答案正确性和 checklist completion 所需的机器可读字段。
}

\begin{figure*}[t]
\centering
\fbox{%
\begin{minipage}{0.96\linewidth}
{\footnotesize\ttfamily\setlength{\parindent}{0pt}\obeylines
You are an AI evaluator. Your task is to evaluate the quality of an answer.
I will provide you with the user's question, the reference answer (ground truth), a checklist, and the answer to be evaluated.

--- USER QUESTION ---
\{question\}

--- REFERENCE ANSWER (Ground Truth) ---
\{reference\_answer\}
This reference answer is considered the correct and ideal response content-wise.

--- REFERENCE CHECKLIST ---
\{checklist\_lines\}

--- MODEL'S GENERATED ANSWER TO EVALUATE ---
\{generated\_answer\}

--- EVALUATION INSTRUCTIONS ---
Please provide your evaluation strictly in the following format on separate lines:
1. Checklist Score: Determine how many of the \{n\} items in the REFERENCE CHECKLIST have been correctly and completely addressed. The model's generated answer must fully comply with an item for it to be considered complete. State as 'CHECKLIST\_SCORE: [correct\_items]/\{n\}' (e.g., CHECKLIST\_SCORE: \{example\_score\}).
2. Checklist Result Vector: Provide a 0-1 vector indicating whether each checklist item passed, in order. '1' means fully satisfied, '0' means not fully satisfied. Output as 'CHECKLIST\_RESULT: [1,0,1]'.
3. Overall Correctness: Judge whether the generated answer is consistent with the reference answer in core content. Content consistency is key; minor wording differences are acceptable. State as 'OVERALL\_CORRECTNESS: [YES/NO]'.

Provide only these three formatted lines as your response.
}
\end{minipage}%
}
\caption{LLM judge prompt used to evaluate final-answer correctness and checklist completion. The judge outputs only the formatted fields shown above; OA, SA, CS, and DACS are computed deterministically from these outputs and the frozen dependency graphs.}
\label{fig:llm_judge_prompt}
\end{figure*}

\section{Statistical and Cross-Judge Robustness}
\label{sec:appendix_judge_robustness}

\subsection{Item-Bootstrap Uncertainty}
\label{sec:statistical_uncertainty}

Given an evaluation set of 102 items, Table~\ref{tab:main_results_ci} complements the per-model point estimates with 95\% confidence intervals obtained by resampling items ($B=\BREWbootB$, seed \BREWbootSeed). For GPT-5.5, the system with the highest OA, the estimate of \BREWgptfiveOA\% has an interval of [\BREWgptfiveOAlo, \BREWgptfiveOAhi], spanning twenty percentage points. The OA ordering of GPT-5.5 and o4-mini, which have the two highest point estimates, is preserved in \BREWgptVsOfourSep\% of resamples, exceeding 95\%. Among GPT-5.4, o3, and Gemini 3.1 Pro, the maximum pairwise separability is \BREWclusterMaxSep\%; by comparison, a difference that never reverses would yield 100\%. These results provide the strongest evidence of separation for the leading pair, while indicating that the remaining per-model rankings should be interpreted with appropriate caution.

\zh{
考虑到评测集包含 102 道题，表~\ref{tab:main_results_ci} 在给出各模型点估计的同时，还报告了通过题目重采样得到的 95\% 置信区间（$B=\BREWbootB$，随机种子为 \BREWbootSeed）。OA 最高的系统 GPT-5.5 的点估计为 \BREWgptfiveOA\%，对应区间为 [\BREWgptfiveOAlo, \BREWgptfiveOAhi]，跨度为 20 个百分点。点估计位列前二的 GPT-5.5 与 o4-mini，其 OA 排序在 \BREWgptVsOfourSep\% 的重采样中保持不变，比例超过 95\%。GPT-5.4、o3 和 Gemini 3.1 Pro 之间的最大两两可分性为 \BREWclusterMaxSep\%；相比之下，差异从不反转时该值为 100\%。这些结果为领先两个系统的区分提供了最充分的证据，同时表明其余系统的逐模型排名应谨慎解读。
}

\begin{table*}[t]
\centering
\scriptsize
\setlength{\tabcolsep}{4pt}
\caption{Full \ourmethodname results with 95\% bootstrap confidence intervals over the 25 systems. Point estimates match Table~\ref{tab:main_results}; intervals are the 2.5/97.5 percentiles of $B=10000$ bootstrap resamples of the 102 items (seed 20260731). Wide intervals and heavy overlap indicate that per-model rankings should not be over-interpreted; Figure~\ref{fig:bootstrap_gaps_forest} gives paired uncertainty for the within-system metric penalties.\zh{ 当前评测范围内全部 25 个系统的完整 \ourmethodname 结果及 95\% bootstrap 置信区间。点估计与表~\ref{tab:main_results} 一致，区间为对 102 题做 $B=10000$ 次重采样（seed 20260731）所得的 2.5/97.5 百分位。区间宽且大量重叠，说明 per-model 排名不应过度解读；图~\ref{fig:bootstrap_gaps_forest} 给出了同一系统内指标惩罚的配对不确定性。}}
\label{tab:main_results_ci}
\begin{tabular}{lcccc}
\toprule
Model & OA (\%) & SA (\%) & CS (\%) & DACS (\%) \\
\midrule
\multicolumn{5}{c}{\textbf{Tool-augmented VLMs}} \\
\cmidrule(lr){1-5}
GPT-5.5 & 43.1\,{\tiny[33.3,\,52.9]} & 34.3\,{\tiny[25.5,\,44.1]} & 73.6\,{\tiny[66.8,\,79.7]} & 70.3\,{\tiny[63.1,\,77.1]} \\
o4-mini & 35.3\,{\tiny[26.5,\,45.1]} & 20.6\,{\tiny[12.7,\,28.4]} & 49.9\,{\tiny[41.9,\,58.1]} & 44.9\,{\tiny[36.7,\,53.5]} \\
o3 & 34.3\,{\tiny[25.5,\,43.1]} & 15.7\,{\tiny[8.8,\,23.5]} & 39.4\,{\tiny[31.5,\,47.8]} & 34.1\,{\tiny[26.2,\,42.4]} \\
GPT-5.4 & 30.4\,{\tiny[21.6,\,39.2]} & 22.5\,{\tiny[14.7,\,31.4]} & 70.1\,{\tiny[64.2,\,75.9]} & 63.9\,{\tiny[57.1,\,70.7]} \\
Claude Opus 4.7 & 29.4\,{\tiny[20.6,\,38.2]} & 27.5\,{\tiny[19.6,\,36.3]} & 69.1\,{\tiny[62.6,\,75.3]} & 65.7\,{\tiny[59.0,\,72.3]} \\
Grok 4.20 & 22.5\,{\tiny[14.7,\,30.4]} & 17.6\,{\tiny[10.8,\,25.5]} & 69.3\,{\tiny[63.5,\,74.9]} & 62.4\,{\tiny[55.7,\,69.1]} \\
GPT-4o (2024-11) & 18.6\,{\tiny[11.8,\,26.5]} & 12.7\,{\tiny[6.9,\,19.6]} & 42.6\,{\tiny[35.0,\,50.2]} & 37.3\,{\tiny[29.6,\,45.0]} \\
Claude Sonnet 4.6 & 17.6\,{\tiny[10.8,\,25.5]} & 15.7\,{\tiny[8.8,\,23.5]} & 64.5\,{\tiny[58.2,\,70.5]} & 61.5\,{\tiny[55.0,\,67.8]} \\
GPT-4.1 & 16.7\,{\tiny[9.8,\,24.5]} & 13.7\,{\tiny[7.8,\,20.6]} & 48.4\,{\tiny[41.1,\,55.8]} & 41.7\,{\tiny[34.2,\,49.4]} \\
GPT-5.4 Mini & 14.7\,{\tiny[8.8,\,21.6]} & 4.9\,{\tiny[1.0,\,9.8]} & 25.0\,{\tiny[18.6,\,31.7]} & 21.0\,{\tiny[14.7,\,27.7]} \\
GPT-4o Mini (2024-07) & 9.8\,{\tiny[4.9,\,15.7]} & 3.9\,{\tiny[1.0,\,7.8]} & 19.0\,{\tiny[13.7,\,24.8]} & 14.2\,{\tiny[9.5,\,19.6]} \\
\midrule
\multicolumn{5}{c}{\textbf{Tool-free VLMs}} \\
\cmidrule(lr){1-5}
Gemini 3.1 Pro & 34.3\,{\tiny[25.5,\,44.1]} & 31.4\,{\tiny[22.5,\,40.2]} & 73.2\,{\tiny[66.9,\,79.6]} & 70.1\,{\tiny[63.3,\,76.7]} \\
Gemini 3 Flash & 30.4\,{\tiny[21.6,\,39.2]} & 24.5\,{\tiny[16.7,\,33.3]} & 74.4\,{\tiny[69.1,\,79.5]} & 68.6\,{\tiny[62.4,\,74.8]} \\
Gemini 2.5 Pro & 28.4\,{\tiny[20.6,\,37.3]} & 23.5\,{\tiny[15.7,\,32.4]} & 73.0\,{\tiny[67.4,\,78.4]} & 69.6\,{\tiny[63.4,\,75.6]} \\
Gemini 2.5 Flash & 26.5\,{\tiny[18.6,\,35.3]} & 18.6\,{\tiny[11.8,\,26.5]} & 65.0\,{\tiny[58.6,\,71.3]} & 58.0\,{\tiny[51.0,\,64.8]} \\
Llama 4 Maverick & 15.7\,{\tiny[8.8,\,22.5]} & 9.8\,{\tiny[4.9,\,15.7]} & 46.7\,{\tiny[40.2,\,53.4]} & 37.8\,{\tiny[31.0,\,44.7]} \\
Qwen3-VL-32B & 10.8\,{\tiny[4.9,\,17.6]} & 2.9\,{\tiny[0.0,\,6.9]} & 35.7\,{\tiny[29.0,\,42.6]} & 29.0\,{\tiny[22.9,\,35.5]} \\
Qwen2.5-VL-72B & 8.8\,{\tiny[3.9,\,14.7]} & 4.9\,{\tiny[1.0,\,9.8]} & 39.2\,{\tiny[32.3,\,46.3]} & 33.4\,{\tiny[26.7,\,40.5]} \\
\midrule
\multicolumn{5}{c}{\textbf{Deep Research Systems}} \\
\cmidrule(lr){1-5}
o3 Deep Research & 32.4\,{\tiny[23.5,\,42.2]} & 19.6\,{\tiny[11.8,\,27.5]} & 56.1\,{\tiny[48.2,\,63.7]} & 49.8\,{\tiny[41.5,\,58.0]} \\
o4-mini Deep Research & 27.5\,{\tiny[19.6,\,36.3]} & 11.8\,{\tiny[5.9,\,18.6]} & 38.0\,{\tiny[30.4,\,46.0]} & 32.5\,{\tiny[24.7,\,40.5]} \\
Perplexity Sonar Deep Research & 22.5\,{\tiny[14.7,\,30.4]} & 21.6\,{\tiny[13.7,\,29.4]} & 42.5\,{\tiny[34.2,\,50.9]} & 37.2\,{\tiny[28.9,\,45.5]} \\
Tongyi DeepResearch 30B & 6.9\,{\tiny[2.0,\,11.8]} & 2.9\,{\tiny[0.0,\,6.9]} & 22.8\,{\tiny[16.9,\,29.0]} & 19.0\,{\tiny[13.2,\,25.2]} \\
\midrule
\multicolumn{5}{c}{\textbf{Framework-based agents}} \\
\cmidrule(lr){1-5}
DeerFlow (minimax-m2.7) & 16.7\,{\tiny[9.8,\,24.5]} & 12.7\,{\tiny[6.9,\,19.6]} & 37.1\,{\tiny[29.1,\,45.4]} & 31.3\,{\tiny[23.4,\,39.5]} \\
DeerFlow (qwen3-vl-235b) & 15.7\,{\tiny[8.8,\,22.5]} & 9.8\,{\tiny[4.9,\,15.7]} & 41.8\,{\tiny[34.3,\,49.6]} & 38.5\,{\tiny[31.2,\,46.1]} \\
DeerFlow (kimi-k2.5) & 3.9\,{\tiny[1.0,\,7.8]} & 3.9\,{\tiny[1.0,\,7.8]} & 5.6\,{\tiny[1.8,\,10.1]} & 5.3\,{\tiny[1.6,\,9.8]} \\
\bottomrule
\end{tabular}
\end{table*}

\begin{figure*}[t]
\centering
\includegraphics[width=\linewidth]{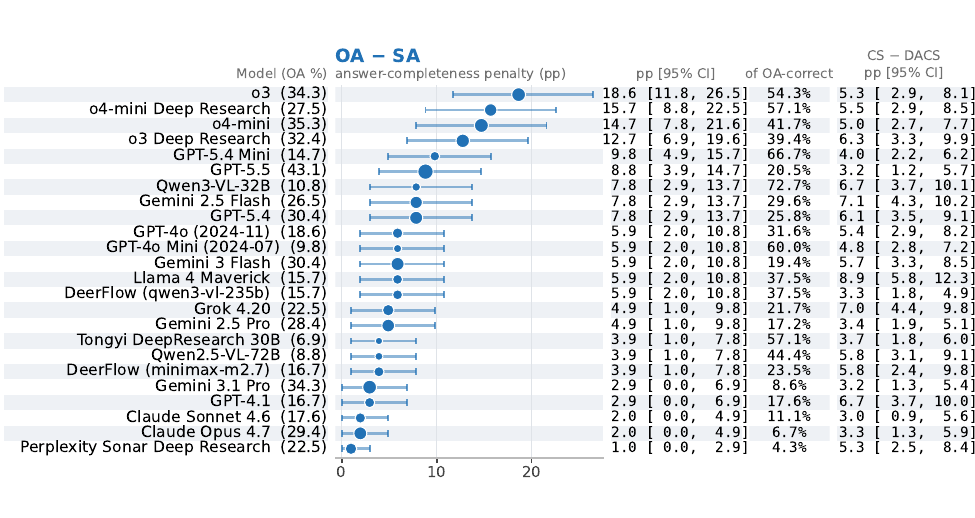}
\caption{Paired item-bootstrap intervals for the answer-completeness penalty OA$-$SA among systems with OA$\geq$5\%, ordered by the penalty itself. Each marker denotes a point estimate, and each line denotes the 95\% percentile interval across \BREWbootB{} resamples. Marker area is proportional to the system's OA, which is printed after the model name and serves as the denominator when the penalty is expressed as a conditional share. The third column presents this conditional share among items with correct final answers. Because the gap is non-negative by construction, an interval reaching zero means that the penalty may be absent, not that its sign is uncertain. For the same systems, the rightmost column reports the dependency penalty CS$-$DACS (point estimate and 95\% CI) as text rather than as a plotted panel because its intervals overlap too heavily to reveal separable differences.\zh{ 图中给出 OA$\geq$5\% 的系统之答案完整性惩罚（answer-completeness penalty，OA$-$SA）的配对题目 bootstrap 区间，并按惩罚本身排序。每个标记表示点估计，每条横线表示 \BREWbootB{} 次重采样所得的 95\% 百分位区间。标记面积与相应系统的 OA 成正比；OA 标注在模型名之后，也是将惩罚表示为条件占比时使用的分母。第三列给出仅以最终答案正确的题目为范围的这一条件占比。由于该差值按定义非负，区间触及 0 表示惩罚可能不存在，而非其符号存在不确定性。最右一列以文字而非绘图面板的形式，报告同一批系统的依赖性惩罚（dependency penalty，CS$-$DACS；点估计与 95\% CI），因为其区间重叠过多，无法显示可分的差异。}}
\label{fig:bootstrap_gaps_forest}
\end{figure*}

Figure~\ref{fig:bootstrap_gaps_forest} reports paired uncertainty for the answer-completeness penalty OA$-$SA: the proportion of all items for which the final answer is correct but the response still omits at least one required conclusion. Because SA is nested within OA, this difference is non-negative by construction; the intervals therefore quantify uncertainty in the \emph{magnitude} of the penalty, not evidence about its direction. When expressed only among items with correct final answers, the penalty reaches \BREWoasaMaxCond\%: for the most affected system, more than eight in ten correct answers are accompanied by responses that omit at least one required conclusion. The figure's rightmost column also reports the dependency penalty CS$-$DACS, namely the checklist credit removed when prerequisite conclusions are absent. We present this penalty as text rather than in a second plotted panel because its point estimates span only \BREWcsdacsMin{} to \BREWcsdacsMax{} percentage points and every interval overlaps every other interval; a second panel would therefore show no separable differences. Moreover, the two penalties use different units of evaluation---whole-item correctness and average checklist credit---so their magnitudes should not be interpreted as a direct statistical comparison.

\zh{
图~\ref{fig:bootstrap_gaps_forest} 报告答案完整性惩罚（answer-completeness penalty，OA$-$SA）的配对不确定性，即全部题目中最终答案正确、但回复仍遗漏至少一个必要结论的题目占比。由于 SA 嵌套于 OA，该差值按定义非负，因此区间刻画的是惩罚\emph{大小}的不确定性，而非其方向的证据。若仅以最终答案正确的题目为范围，该惩罚最高达 \BREWoasaMaxCond\%：对于受影响最严重的系统，超过八成正确答案对应的回复遗漏了至少一个必要结论。图中最右一列还报告了依赖性惩罚（dependency penalty，CS$-$DACS），即前置结论缺失时被扣除的 checklist credit。该惩罚以文字而非第二个绘图面板呈现，因为其点估计仅介于 \BREWcsdacsMin{} 到 \BREWcsdacsMax{} 个百分点之间，且任意两个区间均有重叠；因此，第二个面板不会显示可分的差异。此外，两种惩罚采用不同的评测单位---整题正确性与平均 checklist credit---故不应将其大小解读为直接的统计比较。
}

\subsection{Complete Judge Sampling Frame and Agreement Analysis}

Because SA and DACS depend on checklist-level judgments, every process-level claim in this paper depends on the judge. Reporting results from only one judge would leave this dependence untested. We therefore re-score cached responses from a subset of the evaluated systems (\BREWjudgeResponses{} responses, \BREWjudgeSteps{} checklist steps) using \BREWjudgeN{} judges drawn from \BREWjudgeFamilies{} model families. The scoring prompt and parsing procedure are held fixed across judges; only the model issuing the verdict varies.

\zh{
由于 SA 和 DACS 依赖 checklist 层面的判定，本文所有过程层面的结论都依赖 judge。若仅报告单个 judge 的结果，这种依赖将得不到检验。因此，我们使用来自 \BREWjudgeFamilies{} 个模型家族的 \BREWjudgeN{} 个 judge，对部分评测系统的缓存回复（\BREWjudgeResponses{} 条，\BREWjudgeSteps{} 个 checklist 步）重新判分。所有 judge 使用相同的评分 prompt 和解析流程，唯一变化的是给出判定的模型。
}

Because vendors contribute unequal numbers of systems, we report agreement on a family-balanced subsample. The stratified grid selects the most recent flagship from each model family as its prespecified representative and covers all \BREWitemsubsetItems{} items across all \BREWitemsubsetCats{} categories. This yields \BREWitemsubsetCells{} responses and \BREWitemsubsetSteps{} checklist steps scored by every judge (Table~\ref{tab:item_subset_agreement}).

\zh{
由于各厂商参与评测的系统数量不同，我们在按家族均衡的子样本上报告一致性。分层网格按照预先规定的标准，从每个模型家族中选取最新旗舰型号作为代表，并覆盖全部 \BREWitemsubsetItems{} 道题和 \BREWitemsubsetCats{} 个类别。由此得到 \BREWitemsubsetCells{} 条回复和 \BREWitemsubsetSteps{} 个 checklist 步，且每个 judge 均对其完成评分（表~\ref{tab:item_subset_agreement}）。
}

Agreement is high for both same-vendor and cross-vendor judge pairs. Pairwise final-answer agreement ranges from \BREWitemsubsetOAagreeMin\% to \BREWitemsubsetOAagreeMax\%, with Cohen's $\kappa$ ranging from \BREWitemsubsetOAkappaMin{} to \BREWitemsubsetOAkappaMax. The weakest pair is cross-vendor: among the \BREWitemsubsetXfamPairs{} cross-vendor pairs, the minimum agreement is \BREWitemsubsetXfamAgreeMin\% ($\kappa=\BREWitemsubsetXfamKappaMin$). Fleiss' $\kappa$ is \BREWitemsubsetFleissOA{} for final answers (95\% CI [\BREWitemsubsetFleissOACIlo, \BREWitemsubsetFleissOACIhi] over \BREWitemsubsetBootstrapB{} cell-level bootstrap resamples) and \BREWitemsubsetFleissStep{} for individual checklist steps; all judges are unanimous on \BREWitemsubsetUnanimous\% of responses. A sensitivity analysis that replaces each family's flagship with its \emph{weakest} member yields $\kappa=\BREWitemsubsetSensWeakestFleissOA$ [\BREWitemsubsetSensWeakestFleissOACIlo, \BREWitemsubsetSensWeakestFleissOACIhi]. We also pool \BREWjuryModels{} judges into a single reference verdict for \BREWjuryResponses{} responses, counting a response as correct when at least \BREWjuryThreshold{} judges in the pool agree. The pool is unanimous on \BREWjuryUnanimous\% of responses. The reported judge---which is not a member of the pool---matches this reference on \BREWjuryPaperOAagree\% of final answers ($\kappa=\BREWjuryPaperOAkappa$) and \BREWjuryPaperStepAgree\% of checklist steps ($\kappa=\BREWjuryPaperStepKappa$), as shown in Table~\ref{tab:jury_agreement}.

\zh{
同厂商和跨厂商 judge 配对的一致性均较高。两两最终答案一致率介于 \BREWitemsubsetOAagreeMin\% 与 \BREWitemsubsetOAagreeMax\% 之间，Cohen's $\kappa$ 介于 \BREWitemsubsetOAkappaMin{} 与 \BREWitemsubsetOAkappaMax{} 之间。最弱的配对来自跨厂商 judge：在 \BREWitemsubsetXfamPairs{} 个跨厂商配对中，最低一致率为 \BREWitemsubsetXfamAgreeMin\%（$\kappa=\BREWitemsubsetXfamKappaMin$）。最终答案的 Fleiss' $\kappa$ 为 \BREWitemsubsetFleissOA（基于 \BREWitemsubsetBootstrapB{} 次 cell 层面 bootstrap 重采样的 95\% CI 为 [\BREWitemsubsetFleissOACIlo, \BREWitemsubsetFleissOACIhi]），单个 checklist 步的 Fleiss' $\kappa$ 为 \BREWitemsubsetFleissStep；所有 judge 在 \BREWitemsubsetUnanimous\% 的回复上判定完全一致。敏感性分析将各家族的旗舰替换为其\emph{最弱}成员，得到 $\kappa=\BREWitemsubsetSensWeakestFleissOA$ [\BREWitemsubsetSensWeakestFleissOACIlo, \BREWitemsubsetSensWeakestFleissOACIhi]。此外，我们将 \BREWjuryModels{} 个 judge 在 \BREWjuryResponses{} 条回复上的判定汇聚为单一参考判定：当池中至少 \BREWjuryThreshold{} 个 judge 判为正确时，该回复计为正确。该汇聚池在 \BREWjuryUnanimous\% 的回复上判定完全一致。用于报告结果的 judge 不属于该汇聚池；其判定与参考判定在 \BREWjuryPaperOAagree\% 的最终答案上相符（$\kappa=\BREWjuryPaperOAkappa$），在 \BREWjuryPaperStepAgree\% 的 checklist 步上相符（$\kappa=\BREWjuryPaperStepKappa$），见表~\ref{tab:jury_agreement}。
}

\subsubsection{Detailed Agreement Results}
Table~\ref{tab:item_subset_agreement} reports all pairwise agreement results underlying this analysis: raw final-answer agreement appears below the diagonal and Cohen's $\kappa$ above it for all $\binom{\BREWitemsubsetJudges}{2}$ judge pairs on the family-stratified grid. The judge pool re-scored cached responses from a subset of the evaluated systems (\BREWjudgeResponses{} responses, \BREWjudgeSteps{} checklist steps), and these verdicts support the cross-judge checks reported elsewhere in the paper. The matrix is restricted to the stratified grid so that the agreement statistic does not depend on the number of systems contributed by each vendor.

\zh{
表~\ref{tab:item_subset_agreement} 报告了支撑上述分析的全部两两一致性结果：对角线以下为最终答案的原始一致率，对角线以上为 Cohen's $\kappa$，覆盖按家族分层网格上的全部 $\binom{\BREWitemsubsetJudges}{2}$ 个 judge 配对。judge 池对部分评测系统的缓存回复（\BREWjudgeResponses{} 条，\BREWjudgeSteps{} 个 checklist 步）重新判分，本文其他部分报告的跨 judge 检验均以这些判定为基础。该矩阵仅限于分层网格，从而避免一致性统计量受到各厂商参评系统数量的影响。
}

\begin{table}[t]
\centering
\small
\setlength{\tabcolsep}{3pt}
\caption{Inter-judge agreement on a family-stratified model subset: one representative per model family, selected as the family's most recent flagship (OpenAI (gpt-5.5), Anthropic (claude-opus-4.7), Google (gemini-3.1-pro-preview), Qwen (qwen3-vl-32b-instruct), Meta (llama-4-maverick), Perplexity (sonar-deep-research), xAI (grok-4.20), Alibaba (tongyi-deepresearch-30b-a3b)), covering all 8 model families and all 102 items (all 8 categories) in the benchmark. Of 799 candidate (item, representative) cells with a real, non-empty response, 97 lack a verdict from at least one judge and are excluded, leaving a 702-cell grid evaluated against 12 judges (the reported judge, one additional Qwen judge, and the 10-member cross-vendor jury pool). Columns are numbered as the rows; lower triangle: raw agreement on final answer (OA); upper triangle: Cohen's $\kappa$. Cell shading encodes magnitude (\textcolor[HTML]{4292C6}{blue} for OA over a fixed 95--100\% range; \textcolor[HTML]{F16913}{orange} for $\kappa$ over a fixed 0.80--1.00 range; darker = higher agreement); exact values stay printed in every cell. A representative-selection sensitivity check (each family's weakest member instead of its flagship) and bootstrap 95\% CIs are reported in the accompanying macros.\zh{{在一个按模型家族分层抽样的子集上度量的跨judge一致性：每个模型家族取其最新旗舰型号作为代表（OpenAI (gpt-5.5), Anthropic (claude-opus-4.7), Google (gemini-3.1-pro-preview), Qwen (qwen3-vl-32b-instruct), Meta (llama-4-maverick), Perplexity (sonar-deep-research), xAI (grok-4.20), Alibaba (tongyi-deepresearch-30b-a3b)），覆盖全部 8 个模型家族与全部 102 道题（8 个类别全部覆盖）。在 799 个"题目 x 代表模型"的候选格中（每格都有真实、非空回复），有 97 格缺少至少一位judge的判定被排除，最终形成 702 格的稠密网格，与 12 个judge（本文报告的 judge、一个额外的 Qwen judge，以及 10 名跨厂商 jury 判官池成员）比对。列按行同样编号；下三角为最终答案（OA）的原始一致率；上三角为 Cohen's $\kappa$。格子底色编码数值大小（OA 用固定 95--100\% 范围的\textcolor[HTML]{4292C6}{蓝色}，$\kappa$ 用固定 0.80--1.00 范围的\textcolor[HTML]{F16913}{橙色}，越深表示一致性越高）；每格仍保留精确数值。代表选取的敏感性检验（每家族取最弱成员而非旗舰）与 bootstrap 95\% 置信区间见配套宏。}}}
\label{tab:item_subset_agreement}
\begin{tabular}{lcccccccccccc}
\toprule
Judge & 1 & 2 & 3 & 4 & 5 & 6 & 7 & 8 & 9 & 10 & 11 & 12 \\
\midrule
1. Qwen3-VL-235B (Qwen) & \cellcolor[HTML]{F2F2F2}\textcolor[HTML]{8A8A8A}{--} & \cellcolor[HTML]{FDD6AD}\textcolor[HTML]{202124}{0.870} & \cellcolor[HTML]{FEE3C8}\textcolor[HTML]{202124}{0.845} & \cellcolor[HTML]{FEDFBF}\textcolor[HTML]{202124}{0.853} & \cellcolor[HTML]{FEDDBC}\textcolor[HTML]{202124}{0.856} & \cellcolor[HTML]{FDD7B0}\textcolor[HTML]{202124}{0.868} & \cellcolor[HTML]{FEE0C2}\textcolor[HTML]{202124}{0.851} & \cellcolor[HTML]{FDD6AE}\textcolor[HTML]{202124}{0.869} & \cellcolor[HTML]{FEDFC0}\textcolor[HTML]{202124}{0.852} & \cellcolor[HTML]{FEDDBB}\textcolor[HTML]{202124}{0.857} & \cellcolor[HTML]{FEDFC0}\textcolor[HTML]{202124}{0.852} & \cellcolor[HTML]{FEDBB9}\textcolor[HTML]{202124}{0.859} \\
2. Qwen3.5-Flash (Qwen) & \cellcolor[HTML]{D0E2F2}\textcolor[HTML]{202124}{96.6} & \cellcolor[HTML]{F2F2F2}\textcolor[HTML]{8A8A8A}{--} & \cellcolor[HTML]{FDC590}\textcolor[HTML]{202124}{0.893} & \cellcolor[HTML]{FDC895}\textcolor[HTML]{202124}{0.890} & \cellcolor[HTML]{FDBB80}\textcolor[HTML]{202124}{0.904} & \cellcolor[HTML]{FDBB7F}\textcolor[HTML]{202124}{0.905} & \cellcolor[HTML]{FDD2A5}\textcolor[HTML]{202124}{0.877} & \cellcolor[HTML]{FDBA7E}\textcolor[HTML]{202124}{0.906} & \cellcolor[HTML]{FDBF86}\textcolor[HTML]{202124}{0.900} & \cellcolor[HTML]{FDC48F}\textcolor[HTML]{202124}{0.894} & \cellcolor[HTML]{FDBF86}\textcolor[HTML]{202124}{0.900} & \cellcolor[HTML]{FDB97D}\textcolor[HTML]{202124}{0.907} \\
3. composer-2.5 (Cursor) & \cellcolor[HTML]{E1EDF8}\textcolor[HTML]{202124}{95.9} & \cellcolor[HTML]{BAD6EB}\textcolor[HTML]{202124}{97.3} & \cellcolor[HTML]{F2F2F2}\textcolor[HTML]{8A8A8A}{--} & \cellcolor[HTML]{F9812F}\textcolor[HTML]{202124}{0.973} & \cellcolor[HTML]{F87D2A}\textcolor[HTML]{202124}{0.978} & \cellcolor[HTML]{FB8735}\textcolor[HTML]{202124}{0.967} & \cellcolor[HTML]{F67824}\textcolor[HTML]{202124}{0.984} & \cellcolor[HTML]{FD9040}\textcolor[HTML]{202124}{0.957} & \cellcolor[HTML]{F67824}\textcolor[HTML]{202124}{0.984} & \cellcolor[HTML]{F87D29}\textcolor[HTML]{202124}{0.978} & \cellcolor[HTML]{F67824}\textcolor[HTML]{202124}{0.984} & \cellcolor[HTML]{FB8634}\textcolor[HTML]{202124}{0.968} \\
4. gemini-3.1-pro (Google) & \cellcolor[HTML]{DEEBF7}\textcolor[HTML]{202124}{96.0} & \cellcolor[HTML]{C0D8ED}\textcolor[HTML]{202124}{97.2} & \cellcolor[HTML]{5FA6D1}\textcolor[HTML]{202124}{99.3} & \cellcolor[HTML]{F2F2F2}\textcolor[HTML]{8A8A8A}{--} & \cellcolor[HTML]{F9812F}\textcolor[HTML]{202124}{0.973} & \cellcolor[HTML]{F9812F}\textcolor[HTML]{202124}{0.973} & \cellcolor[HTML]{FB8634}\textcolor[HTML]{202124}{0.968} & \cellcolor[HTML]{FD9C52}\textcolor[HTML]{202124}{0.941} & \cellcolor[HTML]{F77C29}\textcolor[HTML]{202124}{0.979} & \cellcolor[HTML]{F67724}\textcolor[HTML]{202124}{0.984} & \cellcolor[HTML]{F77C29}\textcolor[HTML]{202124}{0.979} & \cellcolor[HTML]{F9812E}\textcolor[HTML]{202124}{0.974} \\
5. gemini-3.5-flash (Google) & \cellcolor[HTML]{DAE9F6}\textcolor[HTML]{202124}{96.2} & \cellcolor[HTML]{AFD1E7}\textcolor[HTML]{202124}{97.6} & \cellcolor[HTML]{59A2CF}\textcolor[HTML]{202124}{99.4} & \cellcolor[HTML]{5FA6D1}\textcolor[HTML]{202124}{99.3} & \cellcolor[HTML]{F2F2F2}\textcolor[HTML]{8A8A8A}{--} & \cellcolor[HTML]{F87D29}\textcolor[HTML]{202124}{0.978} & \cellcolor[HTML]{F9822F}\textcolor[HTML]{202124}{0.973} & \cellcolor[HTML]{FD994D}\textcolor[HTML]{202124}{0.946} & \cellcolor[HTML]{F9822F}\textcolor[HTML]{202124}{0.973} & \cellcolor[HTML]{FB8735}\textcolor[HTML]{202124}{0.967} & \cellcolor[HTML]{F9822F}\textcolor[HTML]{202124}{0.973} & \cellcolor[HTML]{FB8634}\textcolor[HTML]{202124}{0.968} \\
6. gemini-3.6-flash (Google) & \cellcolor[HTML]{D3E4F3}\textcolor[HTML]{202124}{96.4} & \cellcolor[HTML]{AFD1E7}\textcolor[HTML]{202124}{97.6} & \cellcolor[HTML]{65AAD4}\textcolor[HTML]{202124}{99.1} & \cellcolor[HTML]{5FA6D1}\textcolor[HTML]{202124}{99.3} & \cellcolor[HTML]{59A2CF}\textcolor[HTML]{202124}{99.4} & \cellcolor[HTML]{F2F2F2}\textcolor[HTML]{8A8A8A}{--} & \cellcolor[HTML]{F9822F}\textcolor[HTML]{202124}{0.973} & \cellcolor[HTML]{FDA159}\textcolor[HTML]{202124}{0.935} & \cellcolor[HTML]{F67824}\textcolor[HTML]{202124}{0.984} & \cellcolor[HTML]{FB8634}\textcolor[HTML]{202124}{0.967} & \cellcolor[HTML]{F9812F}\textcolor[HTML]{202124}{0.973} & \cellcolor[HTML]{F77C29}\textcolor[HTML]{202124}{0.979} \\
7. glm-5.2 (Zhipu) & \cellcolor[HTML]{DEEBF7}\textcolor[HTML]{202124}{96.0} & \cellcolor[HTML]{C9DDF0}\textcolor[HTML]{202124}{96.9} & \cellcolor[HTML]{549ECD}\textcolor[HTML]{202124}{99.6} & \cellcolor[HTML]{65AAD4}\textcolor[HTML]{202124}{99.1} & \cellcolor[HTML]{5FA6D1}\textcolor[HTML]{202124}{99.3} & \cellcolor[HTML]{5FA6D1}\textcolor[HTML]{202124}{99.3} & \cellcolor[HTML]{F2F2F2}\textcolor[HTML]{8A8A8A}{--} & \cellcolor[HTML]{FD9D53}\textcolor[HTML]{202124}{0.940} & \cellcolor[HTML]{F87D29}\textcolor[HTML]{202124}{0.978} & \cellcolor[HTML]{F9822F}\textcolor[HTML]{202124}{0.973} & \cellcolor[HTML]{F87D29}\textcolor[HTML]{202124}{0.978} & \cellcolor[HTML]{FC8B3A}\textcolor[HTML]{202124}{0.962} \\
8. gpt-5.1 (OpenAI) & \cellcolor[HTML]{D3E4F3}\textcolor[HTML]{202124}{96.4} & \cellcolor[HTML]{AFD1E7}\textcolor[HTML]{202124}{97.6} & \cellcolor[HTML]{72B2D8}\textcolor[HTML]{202124}{98.9} & \cellcolor[HTML]{88BEDC}\textcolor[HTML]{202124}{98.4} & \cellcolor[HTML]{81BADB}\textcolor[HTML]{202124}{98.6} & \cellcolor[HTML]{8FC2DE}\textcolor[HTML]{202124}{98.3} & \cellcolor[HTML]{88BEDC}\textcolor[HTML]{202124}{98.4} & \cellcolor[HTML]{F2F2F2}\textcolor[HTML]{8A8A8A}{--} & \cellcolor[HTML]{FD9446}\textcolor[HTML]{202124}{0.952} & \cellcolor[HTML]{FD984C}\textcolor[HTML]{202124}{0.946} & \cellcolor[HTML]{FD9D52}\textcolor[HTML]{202124}{0.941} & \cellcolor[HTML]{FD984B}\textcolor[HTML]{202124}{0.947} \\
9. gpt-5.5 (OpenAI) & \cellcolor[HTML]{DEEBF7}\textcolor[HTML]{202124}{96.0} & \cellcolor[HTML]{B5D4E9}\textcolor[HTML]{202124}{97.4} & \cellcolor[HTML]{549ECD}\textcolor[HTML]{202124}{99.6} & \cellcolor[HTML]{59A2CF}\textcolor[HTML]{202124}{99.4} & \cellcolor[HTML]{5FA6D1}\textcolor[HTML]{202124}{99.3} & \cellcolor[HTML]{549ECD}\textcolor[HTML]{202124}{99.6} & \cellcolor[HTML]{59A2CF}\textcolor[HTML]{202124}{99.4} & \cellcolor[HTML]{79B6D9}\textcolor[HTML]{202124}{98.7} & \cellcolor[HTML]{F2F2F2}\textcolor[HTML]{8A8A8A}{--} & \cellcolor[HTML]{F9812F}\textcolor[HTML]{202124}{0.973} & \cellcolor[HTML]{F4731E}\textcolor[HTML]{202124}{0.989} & \cellcolor[HTML]{F67723}\textcolor[HTML]{202124}{0.984} \\
10. gpt-5.6-terra (OpenAI) & \cellcolor[HTML]{DAE9F6}\textcolor[HTML]{202124}{96.2} & \cellcolor[HTML]{BAD6EB}\textcolor[HTML]{202124}{97.3} & \cellcolor[HTML]{59A2CF}\textcolor[HTML]{202124}{99.4} & \cellcolor[HTML]{549ECD}\textcolor[HTML]{202124}{99.6} & \cellcolor[HTML]{65AAD4}\textcolor[HTML]{202124}{99.1} & \cellcolor[HTML]{65AAD4}\textcolor[HTML]{202124}{99.1} & \cellcolor[HTML]{5FA6D1}\textcolor[HTML]{202124}{99.3} & \cellcolor[HTML]{81BADB}\textcolor[HTML]{202124}{98.6} & \cellcolor[HTML]{5FA6D1}\textcolor[HTML]{202124}{99.3} & \cellcolor[HTML]{F2F2F2}\textcolor[HTML]{8A8A8A}{--} & \cellcolor[HTML]{F9812F}\textcolor[HTML]{202124}{0.973} & \cellcolor[HTML]{FB8634}\textcolor[HTML]{202124}{0.968} \\
11. kimi-k2.7-code (Moonshot) & \cellcolor[HTML]{DEEBF7}\textcolor[HTML]{202124}{96.0} & \cellcolor[HTML]{B5D4E9}\textcolor[HTML]{202124}{97.4} & \cellcolor[HTML]{549ECD}\textcolor[HTML]{202124}{99.6} & \cellcolor[HTML]{59A2CF}\textcolor[HTML]{202124}{99.4} & \cellcolor[HTML]{5FA6D1}\textcolor[HTML]{202124}{99.3} & \cellcolor[HTML]{5FA6D1}\textcolor[HTML]{202124}{99.3} & \cellcolor[HTML]{59A2CF}\textcolor[HTML]{202124}{99.4} & \cellcolor[HTML]{88BEDC}\textcolor[HTML]{202124}{98.4} & \cellcolor[HTML]{4E9ACB}\textcolor[HTML]{202124}{99.7} & \cellcolor[HTML]{5FA6D1}\textcolor[HTML]{202124}{99.3} & \cellcolor[HTML]{F2F2F2}\textcolor[HTML]{8A8A8A}{--} & \cellcolor[HTML]{F9812E}\textcolor[HTML]{202124}{0.973} \\
12. kimi-k3 (Moonshot) & \cellcolor[HTML]{DAE9F6}\textcolor[HTML]{202124}{96.2} & \cellcolor[HTML]{AFD1E7}\textcolor[HTML]{202124}{97.6} & \cellcolor[HTML]{65AAD4}\textcolor[HTML]{202124}{99.1} & \cellcolor[HTML]{5FA6D1}\textcolor[HTML]{202124}{99.3} & \cellcolor[HTML]{65AAD4}\textcolor[HTML]{202124}{99.1} & \cellcolor[HTML]{59A2CF}\textcolor[HTML]{202124}{99.4} & \cellcolor[HTML]{6BAED6}\textcolor[HTML]{202124}{99.0} & \cellcolor[HTML]{81BADB}\textcolor[HTML]{202124}{98.6} & \cellcolor[HTML]{549ECD}\textcolor[HTML]{202124}{99.6} & \cellcolor[HTML]{65AAD4}\textcolor[HTML]{202124}{99.1} & \cellcolor[HTML]{5FA6D1}\textcolor[HTML]{202124}{99.3} & \cellcolor[HTML]{F2F2F2}\textcolor[HTML]{8A8A8A}{--} \\
\bottomrule
\end{tabular}
\end{table}

Table~\ref{tab:jury_agreement} compares each individual judge against the pooled LLM reference verdict.

\zh{
表~\ref{tab:jury_agreement} 比较每个 judge 与 LLM 汇聚参考判定的一致性。
}

\begin{table}[t]
\centering
\small
\caption{Each judge against the pooled verdict of 10 models scoring the same responses with the identical prompt. A response counts as correct for the pool only when at least 50\% of the judges pass it. Judges inside the pool agree with it by construction; the informative rows are the two that are not.}
\label{tab:jury_agreement}
\begin{tabular}{lccccc}
\toprule
Judge & In pool & OA agr. & OA $\kappa$ & Step agr. & Step $\kappa$ \\
\midrule
Qwen3-VL-235B (reported) & no & 97.1 & 0.890 & 86.6 & 0.728 \\
composer-2.5 & yes & 99.4 & 0.979 & 96.1 & 0.919 \\
gemini-3.1-pro & yes & 99.8 & 0.991 & 96.6 & 0.930 \\
gemini-3.5-flash & yes & 99.5 & 0.982 & 97.5 & 0.948 \\
gemini-3.6-flash & yes & 99.6 & 0.984 & 97.9 & 0.956 \\
glm-5.2 & yes & 99.4 & 0.977 & 93.9 & 0.875 \\
gpt-5.1 & yes & 99.0 & 0.963 & 96.4 & 0.926 \\
gpt-5.5 & yes & 99.8 & 0.993 & 97.2 & 0.942 \\
gpt-5.6-terra & yes & 99.0 & 0.964 & 95.7 & 0.910 \\
kimi-k2.7-code & yes & 99.6 & 0.984 & 96.2 & 0.922 \\
kimi-k3 & yes & 99.8 & 0.993 & 97.5 & 0.949 \\
Qwen3.5-Flash & no & 97.4 & 0.899 & 87.1 & 0.732 \\
\bottomrule
\end{tabular}
\end{table}

\subsubsection{Judge Overlap with an Evaluated System}

\BREWjudgeBackendModel{} in Table~\ref{tab:main_results} uses the same base model as the judge that issues the reported verdicts, and Section~\ref{sec:text_only} evaluates this model directly. Three measurements characterize the resulting risk of self-evaluation. First, \BREWjudgeBackendModel{} achieves \BREWjudgeBackendOA\% OA and \BREWjudgeBackendSA\% SA, both among the lowest values in the table. Second, in the ablation, this judge reports a DACS gain of \BREWtoXselfDACS{} points, below the \BREWtoXmedDACS{}-point median across the \BREWtoXcrossN{} judges from other vendors. Third, the judge is excluded from the pooled reference used for comparison. Within the full judge pool, the reported judge is the most lenient of the \BREWjudgeN: it marks \BREWitemsubsetPaperOArate\% of final answers as correct and passes \BREWitemsubsetPaperStepRate\% of checklist steps, whereas the next-highest rates are \BREWitemsubsetOtherOArateMax\% and \BREWitemsubsetOtherStepRateMax\%, respectively. This leniency also explains the judge's relative isolation in Table~\ref{tab:item_subset_agreement}: every pair involving it has $\kappa \leq \BREWitemsubsetPaperKappaMax$, below the minimum of $\BREWitemsubsetOtherKappaMin$ among pairs that exclude it. Consequently, the reported absolute process scores lie at the lenient end of the pool's range, whereas the comparative conclusions rely on directions that are stable across judges.

\zh{
\textbf{Judge 与被测系统的底座重合。} 表~\ref{tab:main_results} 中的 \BREWjudgeBackendModel{} 与给出报告判定的 judge 使用相同的底座模型，第~\ref{sec:text_only} 节也直接评测了该模型。三项测量刻画了由此产生的自评风险。第一，\BREWjudgeBackendModel{} 的 OA 为 \BREWjudgeBackendOA\%，SA 为 \BREWjudgeBackendSA\%，两者均处于全表最低之列。第二，在消融实验中，该 judge 报告的 DACS 增益为 \BREWtoXselfDACS{} 个百分点，低于来自其他厂商的 \BREWtoXcrossN{} 个 judge 的中位数 \BREWtoXmedDACS{} 个百分点。第三，该 judge 被排除在用于比较的汇聚参考池之外。在完整 judge 池中，本文用于报告结果的 judge 是 \BREWjudgeN{} 个 judge 中最宽松的一个：它将 \BREWitemsubsetPaperOArate\% 的最终答案判为正确，并让 \BREWitemsubsetPaperStepRate\% 的 checklist 步通过；相应的次高比率分别为 \BREWitemsubsetOtherOArateMax\% 和 \BREWitemsubsetOtherStepRateMax\%。这种宽松也解释了该 judge 在表~\ref{tab:item_subset_agreement} 中的相对孤立：凡涉及它的配对，$\kappa$ 均不超过 \BREWitemsubsetPaperKappaMax，低于不涉及它的配对下界 \BREWitemsubsetOtherKappaMin。因此，所报告过程分数的绝对值位于 judge 池取值范围中较宽松的一端，而比较性结论依赖于跨 judge 保持稳定的方向。
}

\subsection{Image Ablation: Cross-Judge Self-Evaluation Check}
\label{sec:appendix_image_judge_robustness}

Because the ablation judge and the evaluated model share the same base model, we examine whether self-preference affects the measured image gain. We re-score every frozen ablation response using all \BREWtoXjudgeN{} judges from \BREWtoXfamilyN{} vendor families and independently recompute the gain under each judge (Table~\ref{tab:text_only_cross_judge}). OA, CS, and DACS increase under every judge in the pool (\BREWtoXposOA, \BREWtoXposCS{}, and \BREWtoXposDACS{} out of \BREWtoXjudgeN, respectively). For DACS, the self-judge reports a \BREWtoXselfDACS{}-point gain, below the \BREWtoXmedDACS{}-point median across the \BREWtoXcrossN{} judges from other vendors. The reported DACS effect therefore lies at the low end of the cross-judge range. SA is less stable: only \BREWtoXposSA{} of the \BREWtoXjudgeN{} judges report a positive gain. This is consistent with the small number of items successful under this metric; accordingly, the conclusion about multimodality is based on OA, CS, and DACS.

\zh{
由于消融实验的 judge 与被测模型共享同一底座模型，我们进一步检验自我偏好是否会影响测得的图像增益。我们使用来自 \BREWtoXfamilyN{} 个厂商家族的全部 \BREWtoXjudgeN{} 个 judge，对每条冻结的消融回复重新判分，并在每个 judge 下独立重新计算增益（表~\ref{tab:text_only_cross_judge}）。OA、CS 和 DACS 在池中每个 judge 下均有所提升（在 \BREWtoXjudgeN{} 个 judge 中分别有 \BREWtoXposOA、\BREWtoXposCS 和 \BREWtoXposDACS{} 个给出正增益）。对于 DACS，自评 judge 报告的增益为 \BREWtoXselfDACS{} 个百分点，低于来自其他厂商的 \BREWtoXcrossN{} 个 judge 的中位数 \BREWtoXmedDACS{} 个百分点。因此，正文报告的 DACS 效应位于跨 judge 取值范围的低端。SA 的方向稳定性较弱：\BREWtoXjudgeN{} 个 judge 中仅 \BREWtoXposSA{} 个报告正增益。这与满足该指标的题目数量较少一致；因此，关于多模态的结论基于 OA、CS 和 DACS。
}

\begin{table}[t]
\centering
\small
\setlength{\tabcolsep}{5pt}
\caption{Robustness of the image$-$text gain to the choice of judge. The ablation responses are frozen; each row re-scores all of them with a different judge and recomputes the gain. The pool spans 6 vendor families and includes the subject model itself ($\dagger$), whose verdicts produced the numbers in Figure~\ref{fig:text_only_baseline}. OA, CS and DACS gains are positive under every judge. The self-judge reports a \emph{smaller} DACS gain than the cross-family median, placing the reported effect at the low end of the cross-judge range. SA is less stable because its base rate is a handful of items and single-verdict changes can flip its sign; the multimodality conclusion is therefore based on OA, CS, and DACS.\zh{ image$-$text 增益对裁判选择的稳健性。消融回答已冻结，每一行用不同裁判重新判分并重算增益。裁判池覆盖 6 个厂商家族，并包含被测模型自身（$\dagger$）——图~\ref{fig:text_only_baseline} 的数字正是由它判出的。OA、CS、DACS 的增益在所有裁判下均为正。自评裁判给出的 DACS 增益低于跨家族中位数，因此报告效应位于跨 judge 范围的低端。SA 的稳定性较低：其基数只有个位数题目，单条判定翻转即可改变符号；多模态结论因此基于 OA、CS 和 DACS。}}
\label{tab:text_only_cross_judge}
\begin{tabular}{llrrrr}
\toprule
Judge & Family & OA & SA & CS & DACS \\
\midrule
Qwen3-VL-235B$^{\dagger}$ & Qwen & $+3.9$ & $+2.9$ & $+12.4$ & $+12.6$ \\
composer-2.5 & Cursor & $+3.9$ & $0.0$ & $+12.4$ & $+15.9$ \\
gemini-3.1-pro & Google & $+3.9$ & $0.0$ & $+12.0$ & $+15.1$ \\
gemini-3.5-flash & Google & $+3.9$ & $+1.0$ & $+12.1$ & $+16.3$ \\
gemini-3.6-flash & Google & $+4.9$ & $+1.0$ & $+12.0$ & $+16.6$ \\
kimi-k2.7-code & Moonshot & $+4.9$ & $+1.0$ & $+13.8$ & $+16.0$ \\
kimi-k3 & Moonshot & $+3.9$ & $0.0$ & $+13.2$ & $+14.9$ \\
gpt-5.1 & OpenAI & $+2.9$ & $-1.0$ & $+12.7$ & $+12.5$ \\
gpt-5.5 & OpenAI & $+3.9$ & $-1.0$ & $+12.1$ & $+13.7$ \\
gpt-5.6-terra & OpenAI & $+3.9$ & $+1.0$ & $+10.4$ & $+10.5$ \\
Qwen3.5-Flash & Qwen & $+4.9$ & $+2.0$ & $+9.5$ & $+7.3$ \\
glm-5.2 & Zhipu & $+4.9$ & $+1.0$ & $+13.2$ & $+14.0$ \\
\midrule
\multicolumn{2}{l}{Median, cross-family ($n=10$)} & $+3.9$ & $+0.5$ & $+12.2$ & $+15.0$ \\
\multicolumn{2}{l}{Judges with a positive gain} & 12/12 & 7/12 & 12/12 & 12/12 \\
\bottomrule
\end{tabular}
\end{table}

\section{Split-Level Performance and Failure Diagnostics}
\label{sec:appendix_split_failure_diagnostics}

This appendix reports complete split-level results by category, coarse authored-template label (Linear vs. Multi-branch), operation tag (Constraint, Symbolic, Numerical, and Temporal), checklist length, and source count. It also provides the protocol and full counts for the failure-mode analysis. Because operation tags are multi-label, the corresponding splits should not be interpreted as disjoint partitions.

\zh{
本附录报告按 category、粗粒度 authored-template 标签（Linear vs. Multi-branch）、operation tag（Constraint、Symbolic、Numerical 和 Temporal）、checklist length 和 source count 划分的完整结果，并给出 failure-mode analysis 的 protocol 和完整计数。由于 operation tag 是多标签的，相应 split 不应被解读为互斥划分。
}

\subsection{Performance by Category}

Table~\ref{tab:appendix_results_by_category} reports complete results for all evaluated systems across the eight benchmark categories.

\zh{
Table~\ref{tab:appendix_results_by_category} 报告所有参评系统在八个 benchmark category 上的完整结果。
}

\begin{sidewaystable}[p]
\centering
\scriptsize
\setlength{\tabcolsep}{1.5pt}
\caption{Model results by category. Metrics are percentages over the items in each split; failed, missing, and invalid responses count as zero. \zh{ 按 category 分解的模型结果。各项指标均为相应 split 内条目的百分比；失败、缺失和无效回答均计为零。}}
\label{tab:appendix_results_by_category}
\begin{adjustbox}{max width=\textheight,center}
\begin{tabular}{lrrrrrrrrrrrrrrrrrrrrrrrrrrrrrrrr}
\toprule
Model & \multicolumn{4}{c}{academia} & \multicolumn{4}{c}{geography} & \multicolumn{4}{c}{media} & \multicolumn{4}{c}{organizations} & \multicolumn{4}{c}{people} & \multicolumn{4}{c}{society} & \multicolumn{4}{c}{sports} & \multicolumn{4}{c}{technology} \\
\cmidrule(lr){2-5}\cmidrule(lr){6-9}\cmidrule(lr){10-13}\cmidrule(lr){14-17}\cmidrule(lr){18-21}\cmidrule(lr){22-25}\cmidrule(lr){26-29}\cmidrule(lr){30-33}
 & OA & SA & CS & DACS & OA & SA & CS & DACS & OA & SA & CS & DACS & OA & SA & CS & DACS & OA & SA & CS & DACS & OA & SA & CS & DACS & OA & SA & CS & DACS & OA & SA & CS & DACS \\
\midrule
GPT-5.5 & 60.0 & 50.0 & 79.1 & 72.5 & 35.3 & 17.6 & 71.0 & 68.7 & 52.4 & 42.9 & 75.8 & 74.4 & 53.8 & 46.2 & 76.5 & 71.2 & 50.0 & 41.7 & 74.9 & 69.0 & 40.0 & 30.0 & 86.1 & 81.1 & 22.2 & 22.2 & 53.7 & 53.7 & 20.0 & 20.0 & 67.8 & 66.9 \\
GPT-5.4 & 30.0 & 20.0 & 66.4 & 59.1 & 23.5 & 23.5 & 72.7 & 71.7 & 42.9 & 23.8 & 71.9 & 60.7 & 30.8 & 30.8 & 76.2 & 73.2 & 33.3 & 25.0 & 62.0 & 48.4 & 50.0 & 30.0 & 81.7 & 72.7 & 11.1 & 11.1 & 57.1 & 57.1 & 10.0 & 10.0 & 67.3 & 66.3 \\
GPT-5.4 Mini & 20.0 & 20.0 & 38.8 & 30.8 & 5.9 & 0.0 & 17.6 & 13.1 & 14.3 & 0.0 & 20.0 & 15.6 & 15.4 & 0.0 & 32.3 & 26.3 & 8.3 & 8.3 & 17.2 & 16.4 & 30.0 & 10.0 & 42.7 & 39.7 & 11.1 & 0.0 & 13.9 & 11.1 & 20.0 & 10.0 & 26.3 & 24.5 \\
o4-mini & 50.0 & 20.0 & 63.6 & 45.3 & 29.4 & 17.6 & 43.8 & 41.3 & 47.6 & 23.8 & 52.7 & 45.8 & 30.8 & 15.4 & 43.1 & 34.6 & 41.7 & 33.3 & 47.6 & 46.8 & 40.0 & 20.0 & 64.5 & 63.5 & 22.2 & 22.2 & 46.3 & 46.3 & 10.0 & 10.0 & 40.9 & 40.0 \\
GPT-4.1 & 20.0 & 20.0 & 28.7 & 28.7 & 5.9 & 5.9 & 45.9 & 42.1 & 38.1 & 33.3 & 59.3 & 51.7 & 0.0 & 0.0 & 48.2 & 38.0 & 16.7 & 8.3 & 47.2 & 32.0 & 20.0 & 10.0 & 53.5 & 48.5 & 11.1 & 11.1 & 46.8 & 43.9 & 10.0 & 10.0 & 47.4 & 41.1 \\
GPT-4o (2024-11) & 20.0 & 20.0 & 35.8 & 31.4 & 5.9 & 5.9 & 43.2 & 42.2 & 28.6 & 19.0 & 39.0 & 33.1 & 15.4 & 7.7 & 50.8 & 43.1 & 41.7 & 25.0 & 49.0 & 32.6 & 20.0 & 10.0 & 50.0 & 47.0 & 0.0 & 0.0 & 37.5 & 34.8 & 10.0 & 10.0 & 34.9 & 34.0 \\
GPT-4o Mini (2024-07) & 20.0 & 20.0 & 25.9 & 20.0 & 11.8 & 5.9 & 24.6 & 16.8 & 9.5 & 4.8 & 19.1 & 15.6 & 7.7 & 0.0 & 12.6 & 11.3 & 8.3 & 0.0 & 3.8 & 0.0 & 0.0 & 0.0 & 20.9 & 18.4 & 11.1 & 0.0 & 30.6 & 16.2 & 10.0 & 0.0 & 17.0 & 16.0 \\
o3 & 60.0 & 50.0 & 66.2 & 62.2 & 29.4 & 11.8 & 37.1 & 36.9 & 47.6 & 19.0 & 41.5 & 36.6 & 30.8 & 0.0 & 19.5 & 6.3 & 41.7 & 25.0 & 37.4 & 32.1 & 30.0 & 10.0 & 47.6 & 41.8 & 0.0 & 0.0 & 31.4 & 29.6 & 20.0 & 10.0 & 39.3 & 30.6 \\
Claude Opus 4.7 & 50.0 & 30.0 & 71.5 & 69.5 & 23.5 & 23.5 & 66.1 & 62.8 & 52.4 & 52.4 & 83.0 & 77.6 & 23.1 & 23.1 & 64.8 & 62.7 & 33.3 & 33.3 & 64.9 & 64.3 & 20.0 & 20.0 & 82.3 & 72.3 & 0.0 & 0.0 & 52.5 & 50.4 & 10.0 & 10.0 & 54.9 & 54.9 \\
Claude Sonnet 4.6 & 30.0 & 30.0 & 70.3 & 68.2 & 5.9 & 5.9 & 60.5 & 60.5 & 19.0 & 14.3 & 75.2 & 69.8 & 15.4 & 15.4 & 61.6 & 55.6 & 25.0 & 16.7 & 51.8 & 45.7 & 30.0 & 30.0 & 74.2 & 74.2 & 11.1 & 11.1 & 59.9 & 59.3 & 10.0 & 10.0 & 56.1 & 54.7 \\
Grok 4.20 & 50.0 & 40.0 & 68.2 & 67.5 & 11.8 & 11.8 & 68.8 & 61.2 & 28.6 & 28.6 & 70.6 & 62.5 & 30.8 & 15.4 & 73.9 & 62.7 & 25.0 & 16.7 & 63.7 & 58.5 & 20.0 & 20.0 & 80.6 & 74.0 & 11.1 & 0.0 & 53.3 & 45.0 & 0.0 & 0.0 & 72.6 & 67.3 \\
Gemini 3.1 Pro & 50.0 & 50.0 & 76.5 & 76.5 & 29.4 & 29.4 & 68.0 & 65.4 & 42.9 & 38.1 & 70.6 & 67.5 & 38.5 & 38.5 & 77.5 & 74.0 & 33.3 & 25.0 & 80.7 & 73.7 & 40.0 & 40.0 & 86.0 & 81.8 & 22.2 & 11.1 & 58.0 & 57.0 & 10.0 & 10.0 & 70.8 & 67.2 \\
Gemini 3 Flash & 50.0 & 40.0 & 75.0 & 70.8 & 29.4 & 23.5 & 75.7 & 66.5 & 42.9 & 33.3 & 79.2 & 72.2 & 30.8 & 30.8 & 75.1 & 69.2 & 25.0 & 16.7 & 70.8 & 64.9 & 30.0 & 30.0 & 82.7 & 78.5 & 0.0 & 0.0 & 56.6 & 54.8 & 20.0 & 10.0 & 72.4 & 69.0 \\
Gemini 2.5 Pro & 40.0 & 30.0 & 70.5 & 63.2 & 17.6 & 17.6 & 67.6 & 65.3 & 47.6 & 33.3 & 79.8 & 75.3 & 15.4 & 15.4 & 66.9 & 62.4 & 25.0 & 25.0 & 76.4 & 75.8 & 40.0 & 40.0 & 85.6 & 83.1 & 11.1 & 11.1 & 61.9 & 59.7 & 20.0 & 10.0 & 71.7 & 69.1 \\
Gemini 2.5 Flash & 50.0 & 40.0 & 71.0 & 55.0 & 23.5 & 17.6 & 68.1 & 64.4 & 33.3 & 23.8 & 64.3 & 58.0 & 23.1 & 15.4 & 66.0 & 59.0 & 25.0 & 16.7 & 67.9 & 58.2 & 40.0 & 20.0 & 77.7 & 67.5 & 0.0 & 0.0 & 49.8 & 45.7 & 10.0 & 10.0 & 51.6 & 49.8 \\
Llama 4 Maverick & 40.0 & 40.0 & 48.6 & 46.6 & 5.9 & 5.9 & 42.3 & 28.7 & 23.8 & 14.3 & 41.3 & 33.6 & 15.4 & 0.0 & 55.8 & 40.9 & 16.7 & 8.3 & 45.2 & 36.1 & 20.0 & 10.0 & 56.4 & 45.8 & 0.0 & 0.0 & 44.8 & 41.1 & 0.0 & 0.0 & 45.7 & 40.3 \\
Qwen3-VL-32B & 20.0 & 10.0 & 41.8 & 38.8 & 5.9 & 0.0 & 20.4 & 9.0 & 19.0 & 4.8 & 38.0 & 31.5 & 15.4 & 0.0 & 42.3 & 35.4 & 16.7 & 8.3 & 34.4 & 28.7 & 0.0 & 0.0 & 47.1 & 45.5 & 0.0 & 0.0 & 30.2 & 14.0 & 0.0 & 0.0 & 37.0 & 37.0 \\
Qwen2.5-VL-72B & 20.0 & 10.0 & 38.1 & 30.4 & 5.9 & 0.0 & 34.5 & 24.3 & 14.3 & 9.5 & 39.1 & 34.8 & 7.7 & 7.7 & 45.7 & 42.2 & 0.0 & 0.0 & 37.1 & 26.3 & 0.0 & 0.0 & 46.0 & 42.0 & 11.1 & 0.0 & 35.7 & 31.1 & 10.0 & 10.0 & 39.3 & 39.3 \\
o3 Deep Research & 50.0 & 20.0 & 52.2 & 34.6 & 17.6 & 5.9 & 31.2 & 29.9 & 38.1 & 28.6 & 73.2 & 65.2 & 38.5 & 15.4 & 75.2 & 64.6 & 41.7 & 33.3 & 51.1 & 50.3 & 50.0 & 40.0 & 76.1 & 73.1 & 0.0 & 0.0 & 28.0 & 28.0 & 20.0 & 10.0 & 53.1 & 43.4 \\
o4-mini Deep Research & 30.0 & 10.0 & 26.9 & 16.0 & 29.4 & 5.9 & 32.8 & 30.4 & 42.9 & 19.0 & 46.7 & 35.7 & 30.8 & 7.7 & 38.3 & 33.3 & 16.7 & 8.3 & 27.0 & 20.8 & 20.0 & 20.0 & 57.4 & 57.4 & 11.1 & 11.1 & 34.8 & 34.8 & 20.0 & 10.0 & 36.3 & 31.7 \\
Tongyi DeepResearch 30B & 20.0 & 10.0 & 18.4 & 16.4 & 0.0 & 0.0 & 14.8 & 14.8 & 19.0 & 9.5 & 20.2 & 16.6 & 0.0 & 0.0 & 25.1 & 16.8 & 0.0 & 0.0 & 24.4 & 15.2 & 0.0 & 0.0 & 37.9 & 34.6 & 11.1 & 0.0 & 12.4 & 8.7 & 0.0 & 0.0 & 35.2 & 35.2 \\
Perplexity Sonar Deep Research & 10.0 & 10.0 & 22.6 & 22.6 & 23.5 & 23.5 & 39.3 & 39.3 & 28.6 & 28.6 & 46.1 & 43.5 & 30.8 & 30.8 & 52.6 & 40.8 & 25.0 & 25.0 & 39.2 & 28.8 & 40.0 & 40.0 & 70.9 & 67.6 & 0.0 & 0.0 & 28.9 & 19.3 & 10.0 & 0.0 & 35.1 & 26.3 \\
DeerFlow (kimi-k2.5) & 20.0 & 20.0 & 20.0 & 20.0 & 0.0 & 0.0 & 1.8 & 0.0 & 4.8 & 4.8 & 4.8 & 4.8 & 0.0 & 0.0 & 0.0 & 0.0 & 8.3 & 8.3 & 8.3 & 8.3 & 0.0 & 0.0 & 0.0 & 0.0 & 0.0 & 0.0 & 15.4 & 15.4 & 0.0 & 0.0 & 0.0 & 0.0 \\
DeerFlow (minimax-m2.7) & 40.0 & 30.0 & 42.4 & 40.4 & 5.9 & 0.0 & 27.6 & 17.7 & 33.3 & 28.6 & 43.6 & 43.6 & 15.4 & 15.4 & 38.5 & 27.0 & 8.3 & 8.3 & 33.3 & 26.2 & 10.0 & 0.0 & 53.3 & 46.5 & 0.0 & 0.0 & 16.0 & 16.0 & 10.0 & 10.0 & 39.5 & 29.5 \\
DeerFlow (qwen3-vl-235b) & 40.0 & 30.0 & 59.8 & 57.7 & 5.9 & 5.9 & 36.0 & 30.9 & 28.6 & 14.3 & 45.3 & 40.7 & 7.7 & 7.7 & 49.2 & 47.9 & 16.7 & 8.3 & 43.1 & 37.3 & 10.0 & 0.0 & 41.4 & 37.8 & 0.0 & 0.0 & 34.2 & 33.3 & 10.0 & 10.0 & 21.7 & 21.7 \\
\bottomrule
\end{tabular}
\end{adjustbox}
\end{sidewaystable}

\subsection{Performance by Authored Template}
\label{sec:appendix_authored_template_performance}

\begin{figure}[t]
\centering
\includegraphics[width=0.8\linewidth]{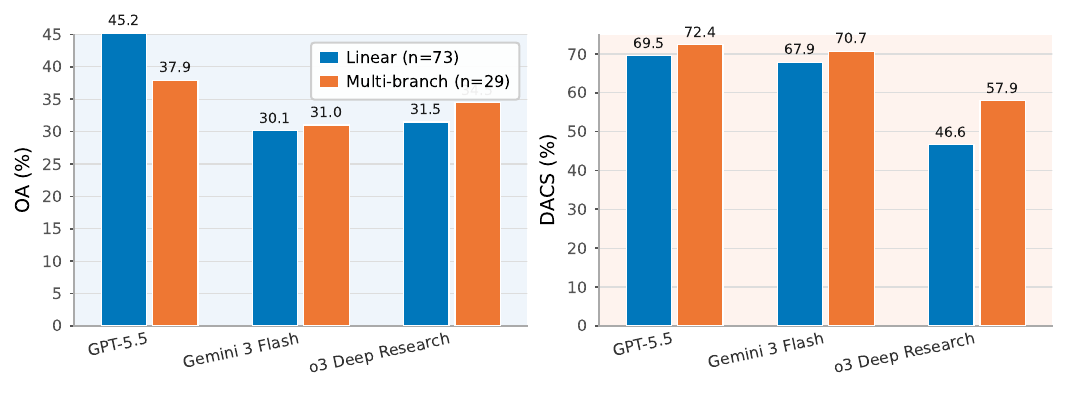}
\caption{Performance of representative models by coarse authored template. The bars compare OA and DACS for Linear- and Multi-branch-template items. \zh{ 代表模型按粗粒度 authored template 分解的表现。柱状图比较 Linear-template 与 Multi-branch-template item 的 OA 和 DACS。}}
\label{fig:linear_vs_multibranch}
\end{figure}

We first partition \ourmethodname items by their coarse authored-template labels. The evaluation set contains 73 Linear-template items, which account for 71.6\% of the benchmark, and 29 Multi-branch-template items, which account for 28.4\%. On average, Multi-branch-template items have slightly longer checklists: 12.8 steps, compared with 11.8 steps for Linear-template items. Although this difference is consistent with the Multi-branch template's construction goal of combining multiple evidence routes, the label does not strictly determine the topology of an item's verified DAG.

\zh{
我们首先按照粗粒度 authored-template 标签对 \ourmethodname item 进行划分。Evaluation set 包含 73 个 Linear-template item，占 benchmark 的 71.6\%；另有 29 个 Multi-branch-template item，占 28.4\%。Multi-branch-template item 的 checklist 平均略长，为 12.8 步，而 Linear-template item 为 11.8 步。尽管这一差异与 Multi-branch 模板组合多条证据路线的构造目标一致，但该标签并不严格决定逐题核定 DAG 的拓扑。
}

As shown in Figure~\ref{fig:linear_vs_multibranch}, the authored-template split does not produce a uniform OA gap: GPT-5.5 is lower on Multi-branch-template items, whereas Gemini 3 Flash and o3 Deep Research are slightly higher. We interpret the split as a diagnostic rather than a standalone difficulty ranking or an exact partition of DAG topologies. Although the Linear- and Multi-branch-template subsets differ in graph topology on average, they also differ in category mix, source availability, and the positions of multimodal bridge steps. The relevant diagnostic signal is that the Multi-branch template tends to reveal whether an agent can maintain multiple partial conclusions until they are jointly needed, whereas the Linear template more directly tests whether an agent can avoid early drift along a predominantly sequential route.

\zh{
如 Figure~\ref{fig:linear_vs_multibranch} 所示，authored-template split 并未产生一致的 OA 差距：GPT-5.5 在 Multi-branch-template item 上更低，而 Gemini 3 Flash 和 o3 Deep Research 略高。我们将该 split 视为一种诊断，而不是独立的难度排序或 DAG 拓扑的精确划分。尽管 Linear-template 和 Multi-branch-template 子集的 graph topology 在总体上存在差异，但二者在 category mix、source availability 以及 multimodal bridge step 的位置上也有所不同。这里相关的诊断信号在于：Multi-branch 模板倾向于揭示 agent 能否保留多个局部结论，直至需要将它们结合使用；而 Linear 模板更直接地测试 agent 能否在主要按顺序推进的路线中避免早期偏移。
}

Table~\ref{tab:appendix_results_by_type} extends this comparison to all evaluated systems.

\zh{
Table~\ref{tab:appendix_results_by_type} 将这一比较扩展到所有参评系统。
}

\begin{table*}[t]
\centering
\scriptsize
\setlength{\tabcolsep}{1.5pt}
\caption{Model results by coarse authored template. Metrics are percentages over the items in each split; failed, missing, and invalid responses count as zero. \zh{ 按粗粒度 authored template 分解的模型结果。各项指标均为相应 split 内条目的百分比；失败、缺失和无效回答均计为零。}}
\label{tab:appendix_results_by_type}
\begin{adjustbox}{max width=\linewidth,center}
\begin{tabular}{lrrrrrrrr}
\toprule
Model & \multicolumn{4}{c}{Linear} & \multicolumn{4}{c}{Multi-branch} \\
\cmidrule(lr){2-5}\cmidrule(lr){6-9}
 & OA & SA & CS & DACS & OA & SA & CS & DACS \\
\midrule
GPT-5.5 & 45.2 & 35.6 & 73.6 & 69.5 & 37.9 & 31.0 & 73.6 & 72.4 \\
GPT-5.4 & 32.9 & 23.3 & 71.3 & 65.3 & 24.1 & 20.7 & 66.9 & 60.5 \\
GPT-5.4 Mini & 12.3 & 1.4 & 20.9 & 17.1 & 20.7 & 13.8 & 35.2 & 30.8 \\
o4-mini & 32.9 & 19.2 & 47.2 & 42.9 & 41.4 & 24.1 & 56.7 & 49.9 \\
GPT-4.1 & 12.3 & 9.6 & 46.0 & 38.0 & 27.6 & 24.1 & 54.3 & 51.0 \\
GPT-4o (2024-11) & 16.4 & 9.6 & 36.8 & 31.4 & 24.1 & 20.7 & 57.2 & 52.0 \\
GPT-4o Mini (2024-07) & 6.8 & 0.0 & 11.8 & 7.5 & 17.2 & 13.8 & 37.2 & 31.1 \\
o3 & 34.2 & 12.3 & 35.9 & 29.4 & 34.5 & 24.1 & 48.2 & 45.7 \\
Claude Opus 4.7 & 30.1 & 28.8 & 70.0 & 66.3 & 27.6 & 24.1 & 66.8 & 64.3 \\
Claude Sonnet 4.6 & 13.7 & 11.0 & 62.9 & 59.1 & 27.6 & 27.6 & 68.4 & 67.5 \\
Grok 4.20 & 20.5 & 15.1 & 71.4 & 63.6 & 27.6 & 24.1 & 64.2 & 59.2 \\
Gemini 3.1 Pro & 32.9 & 30.1 & 73.6 & 69.4 & 37.9 & 34.5 & 72.3 & 71.7 \\
Gemini 3 Flash & 30.1 & 23.3 & 74.6 & 67.9 & 31.0 & 27.6 & 73.7 & 70.7 \\
Gemini 2.5 Pro & 26.0 & 20.5 & 73.4 & 70.0 & 34.5 & 31.0 & 72.0 & 68.8 \\
Gemini 2.5 Flash & 23.3 & 17.8 & 63.7 & 55.6 & 34.5 & 20.7 & 68.3 & 63.9 \\
Llama 4 Maverick & 11.0 & 5.5 & 43.1 & 32.0 & 27.6 & 20.7 & 55.9 & 52.4 \\
Qwen3-VL-32B & 9.6 & 2.7 & 32.1 & 27.2 & 13.8 & 3.4 & 44.7 & 33.5 \\
Qwen2.5-VL-72B & 5.5 & 4.1 & 34.6 & 29.3 & 17.2 & 6.9 & 50.8 & 43.7 \\
o3 Deep Research & 31.5 & 17.8 & 54.3 & 46.6 & 34.5 & 24.1 & 60.5 & 57.9 \\
o4-mini Deep Research & 27.4 & 8.2 & 32.3 & 26.0 & 27.6 & 20.7 & 52.4 & 48.9 \\
Tongyi DeepResearch 30B & 6.8 & 1.4 & 18.7 & 14.4 & 6.9 & 6.9 & 33.0 & 30.6 \\
Perplexity Sonar Deep Research & 23.3 & 23.3 & 44.1 & 38.4 & 20.7 & 17.2 & 38.4 & 34.1 \\
DeerFlow (kimi-k2.5) & 1.4 & 1.4 & 1.8 & 1.4 & 10.3 & 10.3 & 15.1 & 15.1 \\
DeerFlow (minimax-m2.7) & 12.3 & 9.6 & 32.9 & 26.3 & 27.6 & 20.7 & 47.5 & 43.9 \\
DeerFlow (qwen3-vl-235b) & 12.3 & 5.5 & 34.4 & 30.7 & 24.1 & 20.7 & 60.4 & 58.0 \\
\bottomrule
\end{tabular}
\end{adjustbox}
\end{table*}

\subsection{Performance by Operation Tag}

Table~\ref{tab:appendix_results_by_subtask} reports complete results by operation tag. Because the tags are multi-label, these columns describe overlapping subsets rather than a partition of the evaluation set.

\zh{
Table~\ref{tab:appendix_results_by_subtask} 报告按 operation tag 划分的完整结果。由于这些标签采用多标签形式，各列描述的是相互重叠的子集，而不是对 evaluation set 的互斥划分。
}

\begin{table*}[t]
\centering
\scriptsize
\setlength{\tabcolsep}{1.5pt}
\caption{Model results by operation tag. Operation tags are multi-label, so split counts do not sum to 102 items. Metrics are percentages over the tagged items; failed, missing, and invalid responses count as zero. \zh{ 按 operation tag 分解的模型结果。Operation tag 为多标签，因此各 split 的条目数之和不等于 102。各项指标均为带有相应标签的条目的百分比；失败、缺失和无效回答均计为零。}}
\label{tab:appendix_results_by_subtask}
\begin{adjustbox}{max width=\linewidth,center}
\begin{tabular}{lrrrrrrrrrrrrrrrr}
\toprule
Model & \multicolumn{4}{c}{Constraint} & \multicolumn{4}{c}{Numerical} & \multicolumn{4}{c}{Symbolic} & \multicolumn{4}{c}{Temporal} \\
\cmidrule(lr){2-5}\cmidrule(lr){6-9}\cmidrule(lr){10-13}\cmidrule(lr){14-17}
 & OA & SA & CS & DACS & OA & SA & CS & DACS & OA & SA & CS & DACS & OA & SA & CS & DACS \\
\midrule
GPT-5.5 & 42.1 & 32.9 & 70.1 & 65.7 & 56.7 & 40.0 & 79.1 & 74.7 & 46.9 & 37.0 & 74.7 & 72.1 & 50.0 & 44.4 & 74.5 & 71.9 \\
GPT-5.4 & 34.2 & 26.3 & 67.5 & 60.7 & 30.0 & 23.3 & 65.1 & 62.5 & 34.6 & 25.9 & 71.6 & 65.7 & 22.2 & 22.2 & 57.5 & 51.3 \\
GPT-5.4 Mini & 15.8 & 6.6 & 26.2 & 22.1 & 10.0 & 6.7 & 31.6 & 28.8 & 16.0 & 4.9 & 23.9 & 19.6 & 11.1 & 5.6 & 21.2 & 19.8 \\
o4-mini & 36.8 & 23.7 & 52.9 & 48.4 & 43.3 & 26.7 & 58.8 & 50.0 & 35.8 & 22.2 & 49.8 & 45.8 & 50.0 & 33.3 & 59.5 & 48.3 \\
GPT-4.1 & 17.1 & 13.2 & 48.2 & 42.0 & 20.0 & 13.3 & 46.8 & 40.7 & 16.0 & 13.6 & 50.7 & 43.7 & 22.2 & 22.2 & 50.0 & 45.7 \\
GPT-4o (2024-11) & 18.4 & 13.2 & 43.1 & 37.3 & 16.7 & 13.3 & 47.6 & 42.5 & 19.8 & 13.6 & 44.2 & 37.9 & 22.2 & 16.7 & 50.7 & 42.1 \\
GPT-4o Mini (2024-07) & 9.2 & 3.9 & 19.2 & 14.4 & 16.7 & 10.0 & 29.9 & 22.2 & 9.9 & 3.7 & 17.2 & 12.3 & 16.7 & 11.1 & 30.9 & 21.8 \\
o3 & 35.5 & 17.1 & 42.1 & 36.6 & 43.3 & 30.0 & 50.1 & 44.9 & 34.6 & 13.6 & 36.4 & 30.9 & 38.9 & 22.2 & 47.6 & 38.3 \\
Claude Opus 4.7 & 28.9 & 26.3 & 64.7 & 61.7 & 33.3 & 26.7 & 67.8 & 66.1 & 30.9 & 28.4 & 70.1 & 68.3 & 33.3 & 33.3 & 56.2 & 55.8 \\
Claude Sonnet 4.6 & 22.4 & 19.7 & 63.1 & 59.9 & 26.7 & 20.0 & 66.9 & 62.6 & 19.8 & 17.3 & 66.7 & 63.7 & 33.3 & 27.8 & 60.2 & 54.9 \\
Grok 4.20 & 25.0 & 19.7 & 67.5 & 60.4 & 26.7 & 20.0 & 65.5 & 59.7 & 24.7 & 18.5 & 71.8 & 64.9 & 16.7 & 16.7 & 56.1 & 46.2 \\
Gemini 3.1 Pro & 35.5 & 31.6 & 71.7 & 67.4 & 36.7 & 33.3 & 69.5 & 64.6 & 38.3 & 34.6 & 75.4 & 72.8 & 44.4 & 44.4 & 66.7 & 63.6 \\
Gemini 3 Flash & 30.3 & 23.7 & 71.8 & 65.8 & 36.7 & 30.0 & 70.4 & 66.2 & 33.3 & 25.9 & 75.2 & 70.5 & 33.3 & 33.3 & 66.0 & 62.1 \\
Gemini 2.5 Pro & 31.6 & 25.0 & 70.2 & 65.8 & 33.3 & 30.0 & 67.1 & 65.2 & 28.4 & 23.5 & 72.6 & 69.5 & 33.3 & 33.3 & 69.4 & 66.4 \\
Gemini 2.5 Flash & 30.3 & 21.1 & 64.8 & 57.7 & 30.0 & 23.3 & 65.9 & 62.2 & 28.4 & 19.8 & 67.0 & 60.5 & 22.2 & 22.2 & 64.7 & 57.5 \\
Llama 4 Maverick & 14.5 & 9.2 & 46.3 & 37.7 & 30.0 & 20.0 & 52.8 & 44.3 & 16.0 & 9.9 & 47.7 & 38.2 & 22.2 & 16.7 & 48.8 & 41.1 \\
Qwen3-VL-32B & 10.5 & 3.9 & 33.5 & 27.5 & 10.0 & 3.3 & 39.4 & 34.1 & 11.1 & 1.2 & 36.5 & 28.1 & 5.6 & 5.6 & 33.3 & 29.6 \\
Qwen2.5-VL-72B & 7.9 & 6.6 & 37.4 & 30.5 & 10.0 & 3.3 & 38.9 & 34.4 & 7.4 & 3.7 & 39.5 & 32.2 & 16.7 & 5.6 & 38.5 & 32.5 \\
o3 Deep Research & 32.9 & 19.7 & 55.1 & 48.9 & 36.7 & 23.3 & 63.1 & 56.4 & 33.3 & 18.5 & 54.1 & 48.6 & 38.9 & 33.3 & 58.4 & 54.0 \\
o4-mini Deep Research & 28.9 & 13.2 & 38.4 & 33.2 & 26.7 & 16.7 & 44.5 & 39.5 & 32.1 & 13.6 & 39.8 & 33.9 & 16.7 & 16.7 & 34.1 & 34.1 \\
Tongyi DeepResearch 30B & 6.6 & 3.9 & 25.3 & 21.0 & 10.0 & 6.7 & 36.1 & 31.7 & 4.9 & 1.2 & 18.5 & 15.0 & 11.1 & 11.1 & 31.9 & 25.9 \\
Perplexity Sonar Deep Research & 23.7 & 22.4 & 42.7 & 37.1 & 20.0 & 20.0 & 39.6 & 37.8 & 24.7 & 23.5 & 42.2 & 36.2 & 16.7 & 16.7 & 34.3 & 27.7 \\
DeerFlow (kimi-k2.5) & 3.9 & 3.9 & 5.8 & 5.8 & 10.0 & 10.0 & 12.9 & 11.9 & 2.5 & 2.5 & 4.2 & 4.2 & 22.2 & 22.2 & 27.1 & 25.5 \\
DeerFlow (minimax-m2.7) & 17.1 & 13.2 & 36.8 & 30.9 & 20.0 & 16.7 & 35.0 & 31.6 & 17.3 & 12.3 & 36.9 & 30.1 & 16.7 & 16.7 & 26.8 & 22.9 \\
DeerFlow (qwen3-vl-235b) & 17.1 & 9.2 & 43.4 & 39.1 & 20.0 & 13.3 & 42.2 & 40.0 & 16.0 & 8.6 & 43.6 & 39.9 & 22.2 & 11.1 & 47.7 & 43.5 \\
\bottomrule
\end{tabular}
\end{adjustbox}
\end{table*}

\subsection{Performance by Checklist Length}

The main checklist-length analysis appears in Section~\ref{sec:checklist_length_pressure}; Table~\ref{tab:appendix_results_by_checklist_bucket} reports the complete results, including the small tail bins.

\zh{
Checklist length 的主分析见第~\ref{sec:checklist_length_pressure} 节；Table~\ref{tab:appendix_results_by_checklist_bucket} 报告完整结果，其中包括小样本尾部桶。
}

\begin{table*}[t]
\centering
\scriptsize
\setlength{\tabcolsep}{1.5pt}
\caption{Model results by checklist-length bucket. Metrics are percentages over the items in each split; failed, missing, and invalid responses count as zero. \zh{ 按 checklist 长度分桶的模型结果。各项指标均为相应 split 内条目的百分比；失败、缺失和无效回答均计为零。}}
\label{tab:appendix_results_by_checklist_bucket}
\begin{adjustbox}{max width=\linewidth,center}
\begin{tabular}{lrrrrrrrrrrrrrrrrrrrr}
\toprule
Model & \multicolumn{4}{c}{10} & \multicolumn{4}{c}{11--12} & \multicolumn{4}{c}{13--15} & \multicolumn{4}{c}{16--18} & \multicolumn{4}{c}{19+} \\
\cmidrule(lr){2-5}\cmidrule(lr){6-9}\cmidrule(lr){10-13}\cmidrule(lr){14-17}\cmidrule(lr){18-21}
 & OA & SA & CS & DACS & OA & SA & CS & DACS & OA & SA & CS & DACS & OA & SA & CS & DACS & OA & SA & CS & DACS \\
\midrule
GPT-5.5 & 62.9 & 48.6 & 76.9 & 71.4 & 32.4 & 29.4 & 80.0 & 78.2 & 26.1 & 26.1 & 59.6 & 56.0 & 50.0 & 16.7 & 72.4 & 72.4 & 50.0 & 25.0 & 72.6 & 72.6 \\
GPT-5.4 & 57.1 & 42.9 & 73.1 & 66.9 & 17.6 & 11.8 & 72.7 & 64.8 & 8.7 & 4.3 & 58.2 & 53.3 & 33.3 & 33.3 & 66.2 & 62.0 & 25.0 & 25.0 & 94.9 & 94.9 \\
GPT-5.4 Mini & 25.7 & 11.4 & 37.1 & 33.4 & 2.9 & 0.0 & 20.1 & 15.8 & 13.0 & 4.3 & 19.0 & 18.0 & 16.7 & 0.0 & 10.8 & 0.0 & 25.0 & 0.0 & 15.5 & 4.8 \\
o4-mini & 54.3 & 34.3 & 53.4 & 50.3 & 26.5 & 17.6 & 50.8 & 44.6 & 17.4 & 8.7 & 44.2 & 37.8 & 33.3 & 16.7 & 46.9 & 40.9 & 50.0 & 0.0 & 48.7 & 47.4 \\
GPT-4.1 & 28.6 & 25.7 & 52.0 & 48.0 & 11.8 & 5.9 & 50.1 & 40.1 & 8.7 & 8.7 & 39.1 & 34.0 & 0.0 & 0.0 & 37.7 & 26.5 & 25.0 & 25.0 & 72.1 & 68.2 \\
GPT-4o (2024-11) & 25.7 & 17.1 & 45.4 & 37.4 & 14.7 & 11.8 & 39.9 & 36.3 & 8.7 & 8.7 & 34.2 & 32.4 & 16.7 & 0.0 & 45.5 & 39.5 & 50.0 & 25.0 & 85.2 & 69.3 \\
GPT-4o Mini (2024-07) & 11.4 & 5.7 & 24.6 & 20.0 & 5.9 & 0.0 & 12.7 & 8.4 & 13.0 & 4.3 & 15.5 & 9.9 & 0.0 & 0.0 & 12.3 & 7.1 & 25.0 & 25.0 & 54.2 & 48.0 \\
o3 & 48.6 & 25.7 & 43.4 & 39.7 & 26.5 & 8.8 & 42.3 & 32.5 & 21.7 & 17.4 & 40.4 & 37.1 & 33.3 & 0.0 & 21.1 & 21.1 & 50.0 & 0.0 & 1.3 & 0.0 \\
Claude Opus 4.7 & 45.7 & 45.7 & 78.6 & 77.7 & 23.5 & 20.6 & 73.0 & 65.7 & 21.7 & 17.4 & 53.2 & 52.9 & 0.0 & 0.0 & 45.1 & 38.2 & 25.0 & 25.0 & 80.2 & 76.3 \\
Claude Sonnet 4.6 & 22.9 & 22.9 & 69.1 & 68.0 & 20.6 & 14.7 & 64.8 & 59.3 & 13.0 & 13.0 & 58.2 & 54.9 & 0.0 & 0.0 & 57.2 & 57.2 & 0.0 & 0.0 & 68.5 & 67.2 \\
Grok 4.20 & 34.3 & 31.4 & 72.9 & 67.4 & 17.6 & 8.8 & 69.7 & 59.9 & 13.0 & 13.0 & 61.0 & 55.1 & 16.7 & 0.0 & 74.1 & 70.2 & 25.0 & 25.0 & 75.5 & 69.3 \\
Gemini 3.1 Pro & 45.7 & 42.9 & 76.0 & 74.6 & 32.4 & 26.5 & 79.3 & 72.2 & 21.7 & 21.7 & 60.6 & 59.2 & 16.7 & 16.7 & 54.8 & 54.8 & 50.0 & 50.0 & 97.5 & 97.5 \\
Gemini 3 Flash & 42.9 & 40.0 & 77.4 & 74.3 & 20.6 & 14.7 & 75.1 & 67.9 & 21.7 & 13.0 & 67.1 & 59.9 & 33.3 & 33.3 & 69.1 & 63.2 & 50.0 & 25.0 & 91.1 & 84.7 \\
Gemini 2.5 Pro & 40.0 & 37.1 & 74.9 & 73.4 & 20.6 & 17.6 & 76.8 & 72.9 & 26.1 & 13.0 & 67.7 & 62.3 & 16.7 & 16.7 & 51.6 & 47.4 & 25.0 & 25.0 & 87.3 & 84.7 \\
Gemini 2.5 Flash & 37.1 & 28.6 & 66.3 & 62.9 & 11.8 & 8.8 & 63.9 & 53.5 & 26.1 & 13.0 & 57.3 & 48.0 & 50.0 & 33.3 & 77.1 & 73.2 & 25.0 & 25.0 & 89.9 & 87.2 \\
Llama 4 Maverick & 17.1 & 11.4 & 50.3 & 42.0 & 14.7 & 2.9 & 47.9 & 39.1 & 17.4 & 17.4 & 42.1 & 30.2 & 0.0 & 0.0 & 32.8 & 25.7 & 25.0 & 25.0 & 52.8 & 51.6 \\
Qwen3-VL-32B & 11.4 & 5.7 & 32.6 & 27.1 & 8.8 & 2.9 & 40.7 & 34.9 & 8.7 & 0.0 & 30.2 & 25.3 & 16.7 & 0.0 & 33.1 & 29.2 & 25.0 & 0.0 & 55.7 & 16.2 \\
Qwen2.5-VL-72B & 11.4 & 11.4 & 37.7 & 34.0 & 2.9 & 0.0 & 41.6 & 36.5 & 8.7 & 0.0 & 34.8 & 26.6 & 16.7 & 16.7 & 39.8 & 32.7 & 25.0 & 0.0 & 56.7 & 41.7 \\
o3 Deep Research & 51.4 & 34.3 & 70.6 & 63.4 & 23.5 & 17.6 & 57.7 & 50.6 & 17.4 & 8.7 & 45.7 & 40.8 & 33.3 & 0.0 & 24.0 & 19.1 & 25.0 & 0.0 & 23.9 & 22.6 \\
o4-mini Deep Research & 42.9 & 20.0 & 45.4 & 37.1 & 20.6 & 8.8 & 39.5 & 36.3 & 13.0 & 8.7 & 27.3 & 25.1 & 33.3 & 0.0 & 32.2 & 21.2 & 25.0 & 0.0 & 30.5 & 18.5 \\
Tongyi DeepResearch 30B & 14.3 & 8.6 & 28.0 & 24.0 & 0.0 & 0.0 & 30.1 & 24.1 & 8.7 & 0.0 & 12.1 & 10.4 & 0.0 & 0.0 & 6.9 & 6.9 & 0.0 & 0.0 & 0.0 & 0.0 \\
Perplexity Sonar Deep Research & 37.1 & 34.3 & 52.9 & 46.0 & 14.7 & 14.7 & 40.4 & 36.3 & 8.7 & 8.7 & 29.9 & 24.4 & 16.7 & 16.7 & 31.4 & 31.4 & 50.0 & 50.0 & 59.2 & 50.0 \\
DeerFlow (kimi-k2.5) & 8.6 & 8.6 & 11.7 & 10.9 & 0.0 & 0.0 & 1.7 & 1.7 & 4.3 & 4.3 & 4.3 & 4.3 & 0.0 & 0.0 & 0.0 & 0.0 & 0.0 & 0.0 & 0.0 & 0.0 \\
DeerFlow (minimax-m2.7) & 28.6 & 20.0 & 43.4 & 38.3 & 5.9 & 5.9 & 30.7 & 28.9 & 13.0 & 13.0 & 34.4 & 23.1 & 16.7 & 16.7 & 29.4 & 29.4 & 25.0 & 0.0 & 62.9 & 39.1 \\
DeerFlow (qwen3-vl-235b) & 22.9 & 11.4 & 44.0 & 41.1 & 8.8 & 2.9 & 44.7 & 41.9 & 13.0 & 13.0 & 30.5 & 25.5 & 16.7 & 16.7 & 41.7 & 38.7 & 25.0 & 25.0 & 61.7 & 60.5 \\
\bottomrule
\end{tabular}
\end{adjustbox}
\end{table*}

The 16--18 and 19+ checklist-length buckets contain only 6 and 4 items, respectively. A single item can therefore change these cells by tens of percentage points, and a system can obtain high DACS without solving any items. Consequently, per-split orderings cannot establish that the longest checklist ranges are easier. The difficulty trend reported in Section~\ref{sec:checklist_length_pressure} instead uses the better-populated ranges, where individual items do not dominate the estimate.

\zh{
16--18 与 19+ 两个 checklist-length bucket 分别只有 6 道和 4 道题。因此，单独一道题就能使这些单元格的结果变动数十个百分点，一个系统甚至可能在未解决任何题目的情况下获得很高的 DACS。由此，按 split 的排序无法证明最长的 checklist 区间更容易。第~\ref{sec:checklist_length_pressure} 节报告的难度趋势取自样本较充足、单题变动不会主导估计的区间。
}

\subsection{Performance by Source Count}

Table~\ref{tab:appendix_results_by_source_count} reports complete results by source-count bucket. Source count describes annotated evidence availability, not the number of browsing actions taken by a system or the amount of evidence it actually uses.

\zh{
Table~\ref{tab:appendix_results_by_source_count} 报告按 source-count bucket 划分的完整结果。Source count 描述的是已标注证据的可用性，而不是系统执行的浏览操作数，也不代表系统实际使用的证据量。
}

\begin{table*}[t]
\centering
\scriptsize
\setlength{\tabcolsep}{1.5pt}
\caption{Model results by source-count bucket. Metrics are percentages over the items in each split; failed, missing, and invalid responses count as zero. \zh{ 按 source count 分桶的模型结果。各项指标均为相应 split 内条目的百分比；失败、缺失和无效回答均计为零。}}
\label{tab:appendix_results_by_source_count}
\begin{adjustbox}{max width=\linewidth,center}
\begin{tabular}{lrrrrrrrrrrrrrrrr}
\toprule
Model & \multicolumn{4}{c}{1--6} & \multicolumn{4}{c}{7--8} & \multicolumn{4}{c}{9--10} & \multicolumn{4}{c}{11+} \\
\cmidrule(lr){2-5}\cmidrule(lr){6-9}\cmidrule(lr){10-13}\cmidrule(lr){14-17}
 & OA & SA & CS & DACS & OA & SA & CS & DACS & OA & SA & CS & DACS & OA & SA & CS & DACS \\
\midrule
GPT-5.5 & 51.9 & 33.3 & 81.0 & 76.2 & 43.2 & 35.1 & 72.6 & 70.2 & 40.0 & 36.0 & 67.0 & 62.5 & 30.8 & 30.8 & 73.5 & 73.5 \\
GPT-5.4 & 33.3 & 29.6 & 65.2 & 59.5 & 35.1 & 24.3 & 73.0 & 67.1 & 28.0 & 20.0 & 68.4 & 62.9 & 15.4 & 7.7 & 75.2 & 66.0 \\
GPT-5.4 Mini & 11.1 & 7.4 & 25.7 & 23.7 & 18.9 & 5.4 & 28.5 & 23.9 & 8.0 & 0.0 & 19.8 & 16.0 & 23.1 & 7.7 & 23.6 & 16.7 \\
o4-mini & 40.7 & 29.6 & 53.7 & 48.7 & 29.7 & 16.2 & 45.8 & 42.5 & 44.0 & 24.0 & 51.2 & 42.8 & 23.1 & 7.7 & 51.2 & 47.7 \\
GPT-4.1 & 14.8 & 11.1 & 42.1 & 35.5 & 18.9 & 16.2 & 47.6 & 39.1 & 20.0 & 16.0 & 53.3 & 49.1 & 7.7 & 7.7 & 54.4 & 47.8 \\
GPT-4o (2024-11) & 25.9 & 14.8 & 46.1 & 36.4 & 18.9 & 13.5 & 36.0 & 33.3 & 16.0 & 16.0 & 46.9 & 43.7 & 7.7 & 0.0 & 46.0 & 38.1 \\
GPT-4o Mini (2024-07) & 14.8 & 7.4 & 23.1 & 20.3 & 10.8 & 2.7 & 14.6 & 9.6 & 4.0 & 0.0 & 17.3 & 11.1 & 7.7 & 7.7 & 26.5 & 20.7 \\
o3 & 44.4 & 18.5 & 39.6 & 34.7 & 24.3 & 10.8 & 33.5 & 27.6 & 40.0 & 24.0 & 47.6 & 40.8 & 30.8 & 7.7 & 39.9 & 38.0 \\
Claude Opus 4.7 & 37.0 & 33.3 & 71.4 & 69.6 & 32.4 & 32.4 & 71.6 & 66.7 & 20.0 & 20.0 & 62.2 & 61.1 & 23.1 & 15.4 & 70.4 & 64.0 \\
Claude Sonnet 4.6 & 22.2 & 14.8 & 68.0 & 61.8 & 21.6 & 21.6 & 59.8 & 59.4 & 12.0 & 12.0 & 61.8 & 59.4 & 7.7 & 7.7 & 75.5 & 70.6 \\
Grok 4.20 & 22.2 & 11.1 & 60.0 & 54.2 & 27.0 & 24.3 & 77.5 & 69.3 & 20.0 & 20.0 & 70.3 & 65.2 & 15.4 & 7.7 & 63.5 & 54.0 \\
Gemini 3.1 Pro & 33.3 & 25.9 & 72.1 & 67.0 & 35.1 & 32.4 & 74.4 & 71.0 & 36.0 & 36.0 & 70.4 & 69.3 & 30.8 & 30.8 & 77.7 & 75.2 \\
Gemini 3 Flash & 29.6 & 25.9 & 67.0 & 63.3 & 29.7 & 27.0 & 75.6 & 70.2 & 28.0 & 24.0 & 77.4 & 72.5 & 38.5 & 15.4 & 80.3 & 68.0 \\
Gemini 2.5 Pro & 25.9 & 25.9 & 71.1 & 68.8 & 27.0 & 24.3 & 73.6 & 70.8 & 36.0 & 28.0 & 72.4 & 67.3 & 23.1 & 7.7 & 76.5 & 72.7 \\
Gemini 2.5 Flash & 37.0 & 33.3 & 70.5 & 65.1 & 27.0 & 18.9 & 65.7 & 56.1 & 20.0 & 12.0 & 63.8 & 56.9 & 15.4 & 0.0 & 54.2 & 50.7 \\
Llama 4 Maverick & 11.1 & 7.4 & 39.9 & 29.8 & 10.8 & 10.8 & 47.8 & 42.4 & 24.0 & 12.0 & 48.3 & 38.6 & 23.1 & 7.7 & 54.6 & 39.7 \\
Qwen3-VL-32B & 7.4 & 3.7 & 30.0 & 23.8 & 5.4 & 0.0 & 33.0 & 28.7 & 20.0 & 8.0 & 47.7 & 38.8 & 15.4 & 0.0 & 31.8 & 22.0 \\
Qwen2.5-VL-72B & 11.1 & 7.4 & 39.3 & 29.2 & 5.4 & 2.7 & 34.8 & 32.2 & 16.0 & 8.0 & 47.5 & 40.2 & 0.0 & 0.0 & 35.7 & 32.3 \\
o3 Deep Research & 33.3 & 25.9 & 58.8 & 53.2 & 29.7 & 18.9 & 53.4 & 50.4 & 36.0 & 20.0 & 59.8 & 49.3 & 30.8 & 7.7 & 51.2 & 42.2 \\
o4-mini Deep Research & 18.5 & 7.4 & 29.4 & 27.2 & 32.4 & 13.5 & 39.4 & 32.3 & 28.0 & 16.0 & 45.5 & 38.8 & 30.8 & 7.7 & 37.6 & 31.7 \\
Tongyi DeepResearch 30B & 7.4 & 3.7 & 25.0 & 19.0 & 10.8 & 2.7 & 22.3 & 17.9 & 4.0 & 4.0 & 27.5 & 25.7 & 0.0 & 0.0 & 10.3 & 9.6 \\
Perplexity Sonar Deep Research & 18.5 & 14.8 & 40.3 & 30.8 & 29.7 & 29.7 & 46.1 & 43.8 & 20.0 & 20.0 & 43.7 & 39.5 & 15.4 & 15.4 & 34.7 & 27.3 \\
DeerFlow (kimi-k2.5) & 7.4 & 7.4 & 7.4 & 7.4 & 2.7 & 2.7 & 2.7 & 2.7 & 4.0 & 4.0 & 7.5 & 6.3 & 0.0 & 0.0 & 6.2 & 6.2 \\
DeerFlow (minimax-m2.7) & 14.8 & 11.1 & 39.0 & 26.7 & 21.6 & 13.5 & 37.3 & 35.0 & 16.0 & 16.0 & 35.2 & 30.4 & 7.7 & 7.7 & 35.9 & 31.8 \\
DeerFlow (qwen3-vl-235b) & 18.5 & 11.1 & 37.0 & 34.4 & 13.5 & 8.1 & 40.5 & 37.5 & 16.0 & 12.0 & 50.2 & 46.0 & 15.4 & 7.7 & 39.2 & 35.3 \\
\bottomrule
\end{tabular}
\end{adjustbox}
\end{table*}

\subsection{Failure-Mode Analysis}
\label{sec:appendix_failure_mode_analysis}

This section presents the protocol, coverage, taxonomy, and full group-level counts for the analysis summarized in Section~\ref{sec:failure_mode_analysis}.

\zh{
本节给出 failure-mode analysis 的 protocol、覆盖范围、taxonomy 和完整组级计数；该分析概述于第~\ref{sec:failure_mode_analysis} 节。
}

To characterize failures, we assign each unsupported checklist conclusion a diagnostic label from the MM-BrowseComp taxonomy~\citep{li2025mmbrowsecompcomprehensivebenchmarkmultimodal} (Table~\ref{tab:failure_taxonomy}). To keep the failure definition consistent, checklist support for all 25 systems is determined using the frozen verdicts of Qwen3.5-Flash (\texttt{qwen3.5-flash-2026-02-23}); this judge differs from the current judge used to report the performance metrics. Qwen3.6 Plus (\texttt{qwen3.6-plus-2026-04-02}) then assigns a taxonomy label to each failed checklist statement.

\zh{
为刻画失败，我们按照 MM-BrowseComp taxonomy~\citep{li2025mmbrowsecompcomprehensivebenchmarkmultimodal}，为每个未被支撑的 checklist 结论赋予诊断标签（表~\ref{tab:failure_taxonomy}）。为保持 failure definition 一致，全部 25 个系统的 checklist 支撑情况均由 Qwen3.5-Flash（\texttt{qwen3.5-flash-2026-02-23}）的冻结判定确定；这一 judge 与当前报告性能指标所用的 judge 不同。随后，由 Qwen3.6 Plus（\texttt{qwen3.6-plus-2026-04-02}）为每个失败的 checklist statement 分配 taxonomy label。
}

The analysis yields \BREWfailSteps{} checklist-step decisions from \BREWfailRuns{} model--item runs, covering all \BREWfailSystems{} systems in Table~\ref{tab:main_results}. Of these decisions, \BREWfailFailedSteps{} steps are unsupported and receive failure labels; thus, \BREWfailStepRate\% of the required intermediate conclusions are unsupported, and \BREWfailRunsWithFailure{} runs (\BREWfailRunRate\%) contain at least one unsupported conclusion. This rate is a step-weighted micro-average and is therefore not comparable with the per-system CS values in the main table.

\zh{
该分析从 \BREWfailRuns{} 个 model--item run 中得到 \BREWfailSteps{} 个 checklist-step decision，覆盖表~\ref{tab:main_results} 中全部 \BREWfailSystems{} 个系统。其中，\BREWfailFailedSteps{} 个 step 未得到支撑并被赋予 failure label；因此，\BREWfailStepRate\% 的必要中间结论未得到支撑，且 \BREWfailRunsWithFailure{} 个 run（\BREWfailRunRate\%）至少包含一个未支撑结论。此处比率是在全部系统上按 step 加权的 micro-average，因此不能与主表中的逐系统 CS 直接比较。
}

\begin{table*}[h]
    \centering
        \caption{Taxonomy of failure modes used in our error analysis.}
    \begin{tabularx}{\linewidth}{lX} 
        \toprule
        \textbf{Error Type} & \textbf{Definition} \\
        \midrule

        \texttt{VISUAL\_HALLUCINATION} & The model described something that was not in the image or grossly misidentified a key visual element. \\
        \midrule
        
        \texttt{TOOL\_EXECUTION\_FAILURE} & The model's tool (e.g., web browser) failed due to technical issues like website blocking, CAPTCHAs, or timeouts. \\
        \midrule
        
        \texttt{CONFIRMATION\_BIAS} & The model found an early, plausible-sounding answer and stopped searching for more correct alternatives. \\
        \midrule
        
        \texttt{KNOWLEDGE\_OVERRIDE} & The model ignored specific visual evidence and instead answered from its parameterized knowledge. \\
        \midrule
        
        \texttt{GUESSING\_OR\_FABRICATION} & The model's reasoning process failed, and it invented an answer or made an unsubstantiated guess. \\
        \midrule
        
        \texttt{INCORRECT\_REASONING} & The model had the correct facts but made a logical error in its reasoning chain to reach the final conclusion. \\
        \midrule
        
        \texttt{INSTRUCTION\_MISINTERPRETATION} & The agent got confused by the task prompt and failed to perform the intended action. \\

        \texttt{OTHERS} & Any failure mode that does not fit into the above categories. \\
        
        \bottomrule
    \end{tabularx}

    \label{tab:failure_taxonomy}
\end{table*}
\vspace{3mm}

Table~\ref{tab:failure_mode_analysis} reports the complete failure-mode counts and conditional distributions for each model group.

\zh{
Table~\ref{tab:failure_mode_analysis} 报告各模型组完整的 failure-mode 计数和条件分布。
}

\begin{table*}[t]
\centering
\scriptsize
\setlength{\tabcolsep}{3pt}
\caption{Distribution of failure modes assigned to failed checklist items, aggregated by model group. Percent columns are conditioned on failed checklist items within each group; passed checklist items are excluded. ``Failed steps'' counts the number of checklist entries with a non-null label in \texttt{evaluations.checklist\_failure\_modes}.\zh{ 按模型组聚合的 failed checklist item 失败类型分布。百分比列以各组 failed checklist item 为分母；通过的 checklist item 不计入。``Failed steps'' 是 \texttt{evaluations.checklist\_failure\_modes} 中非空标签的 checklist entry 数。}}
\label{tab:failure_mode_analysis}
\resizebox{\linewidth}{!}{%
\begin{tabular}{lrrrrrrrrrr}
\toprule
Group & Models & Failed steps & Know. override & Bad reasoning & Guess/fab. & Confirm. bias & Tool exec. & Visual halluc. & Instr. misread & Other \\
\midrule
Tool-augmented VLMs & 11 & 7,398 & 38.6 & 21.5 & 11.8 & 10.9 & 1.0 & 3.1 & 0.5 & 12.5 \\
Tool-free VLMs & 7 & 4,123 & 27.4 & 29.8 & 19.9 & 12.7 & 0.3 & 3.0 & 1.7 & 5.1 \\
Deep Research Systems & 4 & 3,152 & 43.5 & 16.9 & 13.7 & 10.2 & 2.4 & 5.3 & 0.5 & 7.5 \\
Framework-based agents & 3 & 2,736 & 19.4 & 8.6 & 4.6 & 3.9 & 39.9 & 1.5 & 1.5 & 20.6 \\
All systems & 25 & 17,409 & 33.8 & 20.6 & 12.9 & 10.1 & 7.2 & 3.2 & 0.9 & 11.1 \\
\bottomrule
\end{tabular}
}%
\end{table*}

\section{Case Studies}

The following audited case studies illustrate how final-answer accuracy and checklist-level process metrics differentiate among research behaviors. They cover a complete solution to an item constructed with the Multi-branch template, a late-stage error in the reasoning chain that propagates to downstream evidence, and an early visual grounding failure followed by partial recovery through a chance guess.

\zh{
以下经审计的案例研究展示了最终答案准确率与清单层面的过程指标如何区分不同的研究行为。这些案例分别包括：完整求解一个采用 Multi-branch 模板构造的条目；一个传播至下游证据的推理链后段错误；以及一次早期视觉实体识别失败后，通过偶然猜测实现的部分恢复。
}

\clearpage
\begin{longtable}{@{}p{0.16\linewidth}p{0.78\linewidth}@{}}
\caption{Case Study Part 1/3: A multi-branch success case. The model correctly completes every checklist item and combines conclusions from the two branches to obtain the correct final answer.}
\label{tab:case_study_part1}\\
\toprule
\textbf{Field} & \textbf{Original content} \\
\midrule
\endfirsthead
\caption[]{Case Study Part 1/3: A multi-branch success case. Continued.}\\
\toprule
\textbf{Field} & \textbf{Original content} \\
\midrule
\endhead
\bottomrule
\endfoot
Figure &
\par\smallskip
\includegraphics[height=2.4cm]{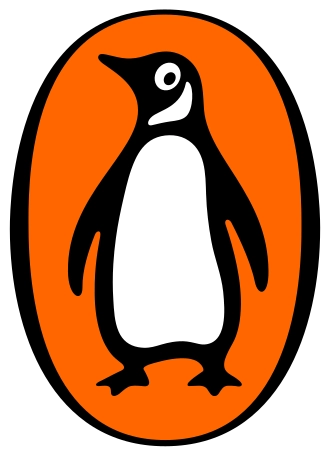} \\
\midrule
Question &
The classic cover design of this publisher, shown in the picture, is one of the most recognizable visual languages in British publishing history. The publisher was founded by someone with family ties to an established publishing institution. The name of this long-standing institution originates from a historical figure who laid the foundation for a renowned university library in Britain. This library features a famous tower displaying five classical orders of architecture. Meanwhile, the publisher’s Penguin logo has been revised, and one of the founding partners of the design company responsible for the revision studied at an art school in London that bears a phoenix as its emblem. How many years are there between the completion year of the tower and the graduation year of this designer? \\
\midrule
Model &
\nolinkurl{openai/gpt-5.5} \\
\midrule
Response &
\begin{minipage}[t]{\linewidth}
\raggedright
The answer is \textbf{337 years}.

\smallskip
Reasoning chain:

\begin{enumerate}
\setlength{\itemsep}{1pt}
\setlength{\parsep}{0pt}
\item The pictured publisher is \textbf{Penguin Books}; its 2003 Penguin logo refresh was done by \textbf{Pentagram Design}. (\url{https://www.penguin.co.uk/discover/articles/penguin-books-logo-history-edward-young-allen-lane})
\item Penguin's founder Allen Lane was tied to \textbf{The Bodley Head}, which was named after \textbf{Sir Thomas Bodley}, founder of the Bodleian Library; Allen Lane succeeded John Lane, his uncle, at The Bodley Head before leaving to set up Penguin. (\url{https://collections.reading.ac.uk/special-collections/collections/the-bodley-head-ltd-publisher/})
\item The Bodleian's \textbf{Tower of the Five Orders} reached its full five-storey form and main outline by \textbf{June 1619}, so I take the tower's completion year as \textbf{1619}. (\url{https://www.oxoniensia.org/volumes/1968/cole.pdf})
\item Pentagram was founded by five partners, including \textbf{Alan Fletcher}. (\url{https://www.pentagram.com/news/happy-50th-birthday-pentagram})
\item Fletcher studied at the \textbf{Royal College of Art} from \textbf{1953 to 1956}; the RCA's phoenix is a recurring official emblem/tradition. (\url{https://en.wikipedia.org/wiki/Alan_Fletcher_\%28graphic_designer\%29})
\end{enumerate}

\smallskip
Calculation:

\smallskip
\textbf{1956 - 1619 = 337}

\smallskip
So, there are \textbf{337 years} between the tower's completion year and Alan Fletcher's graduation/end year.
\end{minipage} \\
\midrule
Answer &
337 \\
\midrule
Checklist &
\begin{minipage}[t]{\linewidth}
\raggedright
\textcolor{green!45!black}{\checkmark} The cover in the picture is recognized as Penguin Books' classic orange three-part cover design. \par
\textcolor{green!45!black}{\checkmark} Penguin Books was founded by Allen Lane in 1935. \par
\textcolor{green!45!black}{\checkmark} Allen Lane was the nephew of John Lane, who founded the publishing house The Bodley Head. \par
\textcolor{green!45!black}{\checkmark} The name The Bodley Head derives from the logo featuring the head of Thomas Bodley. \par
\textcolor{green!45!black}{\checkmark} Thomas Bodley was the founder of the Bodleian Library. \par
\textcolor{green!45!black}{\checkmark} The Bodleian Library features a famous Tower of the Five Orders, displaying five classical architectural styles. \par
\textcolor{green!45!black}{\checkmark} The Tower of the Five Orders was completed around 1619. \par
\textcolor{green!45!black}{\checkmark} Penguin Books' penguin logo was redesigned in 2003 by Angus Hyland, a partner at the design firm Pentagram. \par
\textcolor{green!45!black}{\checkmark} One of Pentagram's founding partners was Alan Fletcher. \par
\textcolor{green!45!black}{\checkmark} Alan Fletcher studied at the Royal College of Art (RCA), whose emblem features a Phoenix. \par
\textcolor{green!45!black}{\checkmark} Alan Fletcher graduated from RCA in 1956. \par
\textcolor{green!45!black}{\checkmark} Calculating 1956 - 1619 = 337 years.
\end{minipage} \\
\midrule
Is\_Correct &
1 \\
\end{longtable}

\begin{longtable}{@{}p{0.16\linewidth}p{0.78\linewidth}@{}}
\caption{Case Study Part 2/3: A late-chain propagation failure. The model reasons correctly through checklist items 1--12, but an error at item 13 propagates through subsequent steps and leads to an incorrect final answer.}
\label{tab:case_study_part2}\\
\toprule
\textbf{Field} & \textbf{Original content} \\
\midrule
\endfirsthead
\caption[]{Case Study Part 2/3: A late-chain propagation failure. Continued.}\\
\toprule
\textbf{Field} & \textbf{Original content} \\
\midrule
\endhead
\bottomrule
\endfoot
Figure &
049365dc
\par\smallskip
\includegraphics[height=2.4cm]{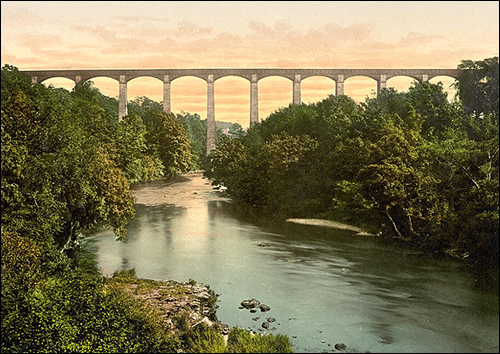} \\
\midrule
Question &
This elevated aqueduct bridge, spanning a mountain valley and closely associated with early British transportation engineering heritage, was designed with the involvement of an engineer; the same engineer is also credited with building a suspension bridge connecting the mainland and an island. The waterway beneath the bridge belongs to a famous strait, and the island is known for its unique administrative status and diverse coastal landscapes. On the island, there is a popular tourist spot with a well-known place name frequently photographed by visitors, and its railway station sign also displays this long place name. The coastal railway line at that station extends westward to a port town, where a ferry route crosses the Irish Sea to the capital port of Ireland. The port reached by this route is located at the edge of a bay, with a long seawall extending into the sea at the port entrance; at the end of this seawall stands a lighthouse, a landmark that appeared in a YouTube video by a travel vlogger with the real name Paul aka. The video features the full name of the club displayed on the wall of a house--what is that full name? \\
\midrule
Model &
\nolinkurl{openai/o4-mini-deep-research-2025-06-26} \\
\midrule
Response &
\begin{minipage}[t]{\linewidth}
\raggedright
The clues all point to Anglesey (linked by Telford's Menai Suspension Bridge over the Menai Strait (\url{https://www.britannica.com/topic/Menai-Bridge\#:~:text=Menai\%20Bridge\%2C\%20suspension\%20bridge\%20,from\%20masonry\%20towers\%20at\%20either})). The long-named village is Llanfairpwllgwyngyllgogerychwyrndrobwllllantysiliogogogoch (the railway station sign famously bears the full name) which lies on the line to Holyhead. Holyhead's huge breakwater (2.75 km long) with its lighthouse at the end is well known (\url{https://lighthouses.wales/the-lighthouses/holyhead-breakwater/\#:~:text=This\%20lighthouse\%20is\%20perched\%20at,by\%20the\%20Holyhead\%20Port\%20Authority}). In the travel video the wall of a house shows the letters ``RAOB''--this stands for the club's full name, the \textbf{Royal Antediluvian Order of Buffaloes} (\url{https://www.wikiwand.com/en/articles/Royal_Antediluvian_Order_of_Buffaloes\#:~:text=The\%20Royal\%20Antediluvian\%20Order\%20of,to\%20members}).

\smallskip
\textbf{Answer:} Royal Antediluvian Order of Buffaloes (\url{https://www.wikiwand.com/en/articles/Royal_Antediluvian_Order_of_Buffaloes\#:~:text=The\%20Royal\%20Antediluvian\%20Order\%20of,to\%20members}).
\end{minipage} \\
\midrule
Answer &
HALF MOON SWIMMING \& WATER POLO CLUB \\
\midrule
Checklist &
\begin{minipage}[t]{\linewidth}
\raggedright
\textcolor{green!45!black}{\checkmark} The elevated waterway bridge in the picture is the Pontcysyllte Aqueduct. \par
\textcolor{green!45!black}{\checkmark} One of the designers of the Pontcysyllte Aqueduct is Thomas Telford. \par
\textcolor{green!45!black}{\checkmark} The designer of the Menai Suspension Bridge is Thomas Telford. \par
\textcolor{green!45!black}{\checkmark} The Menai Suspension Bridge spans the Menai Strait. \par
\textcolor{green!45!black}{\checkmark} The island connected by the Menai Suspension Bridge is the Isle of Anglesey. \par
\textcolor{green!45!black}{\checkmark} There is a place called Llanfairpwllgwyngyll on the Isle of Anglesey. \par
\textcolor{green!45!black}{\checkmark} Llanfairpwllgwyngyll is popular for its extremely long Welsh place name. \par
\textcolor{green!45!black}{\checkmark} The station sign for Llanfairpwllgwyngyll displays this extremely long place name. \par
\textcolor{green!45!black}{\checkmark} The railway station with this long name is Llanfairpwll railway station. \par
\textcolor{green!45!black}{\checkmark} Llanfairpwll railway station is located on the North Wales Coast Line. \par
\textcolor{green!45!black}{\checkmark} The western end of the North Wales Coast Line is Holyhead. \par
\textcolor{green!45!black}{\checkmark} Holyhead railway station is located in Holyhead. \par
\textcolor{red!70!black}{$\times$} Holyhead has a ferry route to Dublin. \par
\textcolor{red!70!black}{$\times$} The Dublin endpoint of the Holyhead--Dublin ferry route is Dublin Port. \par
\textcolor{red!70!black}{$\times$} Dublin Port is located on the Dublin Bay side. \par
\textcolor{red!70!black}{$\times$} The bay entrance related to Dublin Port can be located via the Great South Wall. \par
\textcolor{red!70!black}{$\times$} The landmark at the end of the Great South Wall is Poolbeg Lighthouse. \par
\textcolor{red!70!black}{$\times$} The real name of the account The Hiking Hermit Travels is Paul aka. \par
\textcolor{red!70!black}{$\times$} Poolbeg Lighthouse appears in The Hiking Hermit Travels' video titled ``Looking for FREE things to do in DUBLIN? The GREAT SOUTH WALL and POOLBEG LIGHTHOUSE Walk | 4K video.'' \par
\textcolor{red!70!black}{$\times$} The full name of the club on the house wall in the video is HALF MOON SWIMMING \& WATER POLO CLUB.
\end{minipage} \\
\midrule
Is\_Correct &
0 \\
\end{longtable}

\begin{longtable}{@{}p{0.16\linewidth}p{0.78\linewidth}@{}}
\caption{Case Study Part 3/3: An early visual grounding failure. The model fails checklist items 1--7, guesses item 8 and then infers item 9, but misses the multimodal video step at item 10 and cannot recover item 11.}
\label{tab:case_study_part3}\\
\toprule
\textbf{Field} & \textbf{Original content} \\
\midrule
\endfirsthead
\caption[]{Case Study Part 3/3: An early visual grounding failure. Continued.}\\
\toprule
\textbf{Field} & \textbf{Original content} \\
\midrule
\endhead
\bottomrule
\endfoot
Figure &
a1839503
\par\smallskip
\includegraphics[height=2.4cm]{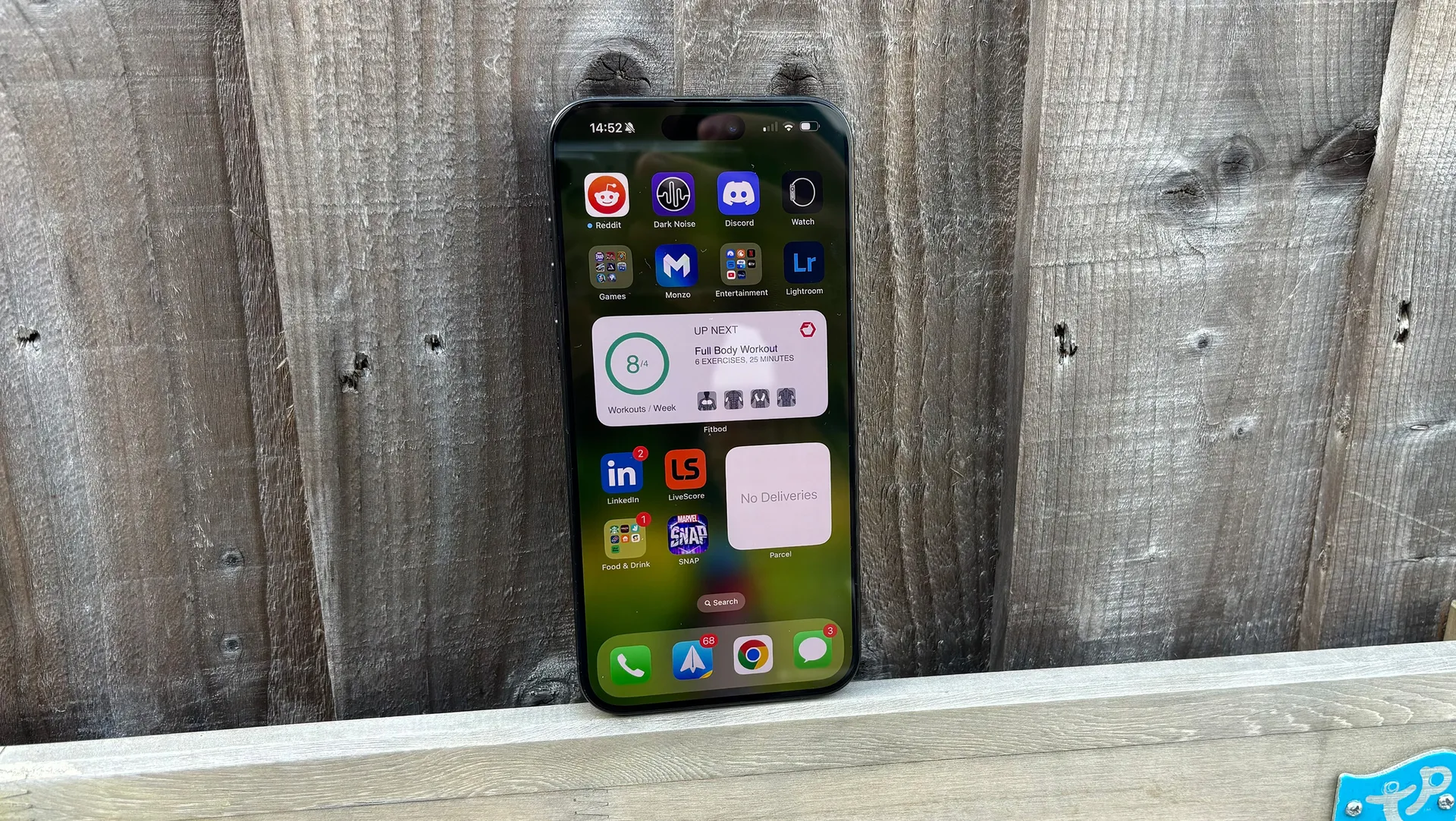} \\
\midrule
Question &
The image shows a flagship smartphone. The processor in this phone was developed by a well-known chip design team, which heavily relies on a leading provider of electronic design automation tools worldwide. Meanwhile, the metal frame material of this phone was first used in two aircraft. Of those, the one that was retired later was moved to a museum after completing a mission with the U.S. Air Force at Edwards AFB. This museum is affiliated with a program, which is a department of an institution. The official YouTube account of the U.S. Department of State has posted a video introducing this institution. In this video, the third person interviewed is wearing a shirt with a few words on it. What is the first word? \\
\midrule
Model &
\nolinkurl{openai/o4-mini-deep-research-2025-06-26} \\
\midrule
Response &
\begin{minipage}[t]{\linewidth}
\raggedright
The smartphone is the Samsung Galaxy S24 Ultra, which uses Samsung's Exynos chip (designed using Synopsys EDA tools) and a titanium frame. Its frame metal -- aerospace-grade titanium -- was first used in two notable aircraft: the Space Shuttle Enterprise (which flew test landings at Edwards Air Force Base) and an F-117 Nighthawk stealth fighter. The F-117 (tail \#783) was later moved to the Air Force Flight Test Museum at Edwards AFB. That museum participates in the Smithsonian Affiliations program (a program of the Smithsonian Institution). The U.S. State Department's YouTube channel has a video introducing the Smithsonian Institution, and in that video the third person interviewed is wearing a shirt with text. The first word on her shirt is \textbf{``Smithsonian.''}
\end{minipage} \\
\midrule
Answer &
climate \\
\midrule
Checklist &
\begin{minipage}[t]{\linewidth}
\raggedright
\textcolor{red!70!black}{$\times$} The product displayed in the picture is the Apple iPhone 15 Pro Max \par
\textcolor{red!70!black}{$\times$} The processor chip installed in the iPhone 15 Pro Max is the A17 Pro \par
\textcolor{red!70!black}{$\times$} The A17 Pro chip is designed by the Apple Silicon team \par
\textcolor{red!70!black}{$\times$} The Apple Silicon team collaborates deeply with EDA tool provider Synopsys \par
\textcolor{red!70!black}{$\times$} The frame material of the iPhone 15 Pro Max is titanium \par
\textcolor{red!70!black}{$\times$} The Lockheed A-12 and the SR-71 ``Blackbird'' were two of the first aircraft frames to use titanium \par
\textcolor{red!70!black}{$\times$} After all USAF and NASA SR-71 operations at Edwards AFB were completed, the SR-71 Flight Simulator was moved in July 2006 to the Frontiers of Flight Museum at Love Field Airport in Dallas, Texas \par
\textcolor{green!45!black}{\checkmark} The Frontiers of Flight Museum is an affiliate within the Smithsonian Affiliations program \par
\textcolor{green!45!black}{\checkmark} Smithsonian Affiliations is a division of the Smithsonian Institution \par
\textcolor{red!70!black}{$\times$} The official YouTube account of the U.S. Department of State has posted a video titled ``The Legacy of the Smithsonian Institution'' \par
\textcolor{red!70!black}{$\times$} In the video ``The Legacy of the Smithsonian Institution,'' the first word on the shirt of the third interviewee is ``climate''
\end{minipage} \\
\midrule
Is\_Correct &
0 \\
\end{longtable}

\clearpage

\section{Release, Reproducibility, and Ethics}
\label{sec:code-release}

\subsection{Released Artifacts and Reproducibility}

We will release the benchmark data together with the evaluator, the exact judge prompt, verified direct-dependency graphs, and regression tests. The benchmark data include the questions, short answers, category/type/subtask labels, irreducible checklists, source URLs, and associated image assets. These artifacts specify both the evaluation inputs and the scoring procedure, allowing researchers to recompute OA, SA, CS, and DACS from model responses or to re-judge cached responses under the same protocol.

\zh{
我们将公开 benchmark 数据、evaluator、评测所用的 judge prompt 原文、经核定的直接依赖图以及 regression tests。Benchmark 数据包括 questions、short answers、category/type/subtask labels、不可约 checklists、source URLs 和关联图像资产。这些材料同时明确了评测输入和评分流程，使研究者能够基于模型回复重新计算 OA、SA、CS 和 DACS，或按照相同协议对缓存回复重新进行 judge 判定。
}

For each evaluated model, we record the model identifier, inference configuration, browsing setting, and evaluation outputs used in the analysis. The regression tests cover score aggregation and dependency-aware scoring, helping detect implementation changes that would alter reported results. Together, the data schema, judge prompt, evaluator, and recorded configurations provide a consistent basis for reproducing the evaluation and extending it to new models.

\zh{
对于每个参评模型，我们记录分析所使用的 model identifier、inference configuration、browsing setting 和 evaluation outputs。Regression tests 覆盖分数汇总与依赖感知评分，以帮助发现可能改变报告结果的实现变动。数据结构、judge prompt、evaluator 和所记录的配置共同构成一致的复现基础，也便于将评测扩展到新的模型。
}

\subsection{Ethical Safeguards}

The benchmark is constructed from publicly accessible web content and is intended to evaluate information seeking, evidence integration, and multimodal reasoning. During item construction and audit, annotators avoid questions that depend on private or sensitive personal information unless that information is both publicly documented and essential to a legitimate informational task. Source URLs are retained to support evidence inspection and attribution, and each item is reviewed for answerability, evidential support, and the substantive use of non-text evidence.

\zh{
本 benchmark 基于公开可访问的网络内容构建，旨在评估信息检索、证据整合和多模态推理能力。在题目构建与审核过程中，标注者避免设计依赖私人或敏感个人信息的问题；仅当相关信息已有公开记录且对于合理的信息任务确有必要时，才会予以使用。我们保留 source URLs 以支持证据核查与归因，并逐题审核其可回答性、证据支撑以及非文本证据是否得到实质性使用。
}

\end{document}